\documentclass[10pt,twocolumn,letterpaper]{article}
\usepackage[T1]{fontenc}
\usepackage{lmodern}

\usepackage{iccv}              
\usepackage[accsupp]{axessibility}

\usepackage{graphicx}
\usepackage{booktabs} 
\usepackage{array}    
\usepackage{rotating} 
\usepackage{array}
\definecolor{iccvblue}{rgb}{0.21,0.49,0.74}
\usepackage[pagebackref,breaklinks,colorlinks,allcolors=iccvblue]{hyperref}

\def\paperID{CVAM-12} 
\def\confName{ICCV}
\def\confYear{2025}

\title{Training‑Free Diffusion Framework for Stylized Image Generation with Identity Preservation}

\author{
Mohammad Ali Rezaei$^{1}$ \quad
Helia Hajikazem$^{1}$ \quad
Saeed Khanehgir$^{1}$ \quad
Mahdi Javanmardi$^{2}$ \\
$^{1}$Computer Vision Group, Part AI Research Center, Tehran, Iran \\
$^{2}$Department of Computer Engineering, Amirkabir University of Technology, Tehran, Iran \\
{\tt\small \{mohammad.rezaei, helia.hajikazem, saeed.khanehgir\}@partdp.ai, mjavan@aut.ac.ir}
}

\begin{document}
\maketitle
\begin{abstract}

Although diffusion models have demonstrated remarkable generative capabilities, existing style transfer techniques often struggle to maintain identity while achieving high-quality stylization. This limitation becomes particularly critical in practical applications such as advertising and marketing, where preserving the identity of featured individuals is essential for a campaign's effectiveness. This limitation is particularly severe when subjects are distant from the camera or appear within a group, frequently leading to a significant loss of identity. To address this issue, we introduce a novel, training-free framework for identity-preserved stylized image synthesis. Key contributions include: the "Mosaic Restored Content Image" technique, which significantly enhances identity retention in complex scenes, and a training-free content consistency loss that improves the preservation of fine-grained details by directing more attention to the original image during stylization. Our experiments reveal that the proposed approach substantially exceeds the baseline model in concurrently maintaining high stylistic fidelity and robust identity integrity, all without necessitating model retraining or fine-tuning.
\end{abstract}
\begin{figure}[!t]
    \centering
    \includegraphics[width=1\linewidth]{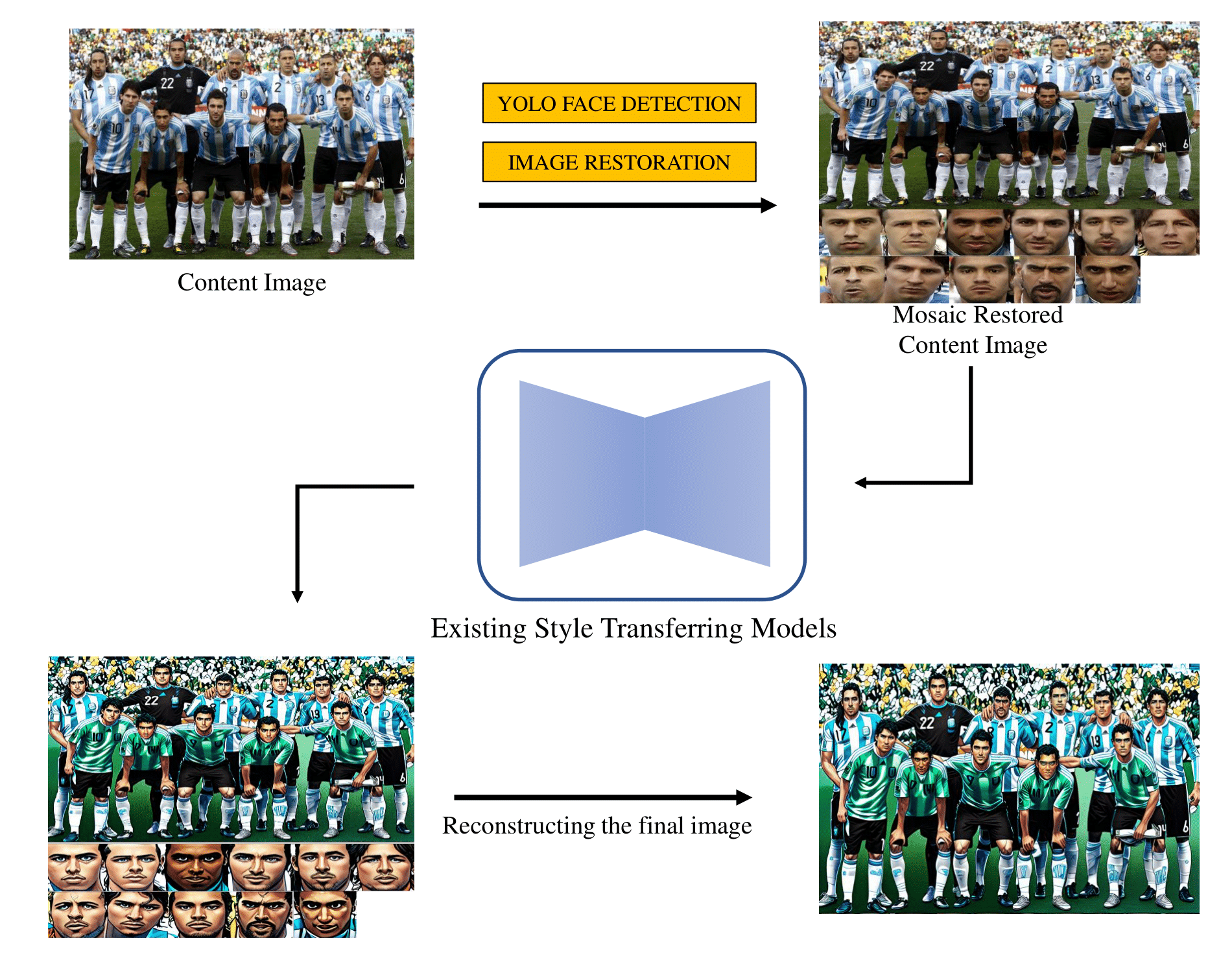}
    \caption{Overview of our identity-preserving pipeline. First, YOLOv8 \cite{varghese2024yolov8} detects faces in the content image, and Real-ESRGAN \cite{wang2021real} enhances them. The enhanced faces are then combined with the original background to form a mosaic image. This mosaic image undergoes style transfer, producing stylized faces and a stylized background within the mosaic. Finally, the stylized faces are extracted from this processed mosaic and reinserted into the final output to maintain identity details.}
    \label{fig:fig1}
\end{figure}
    
\section{Introduction}

In recent years, image style transfer has become a transformative technique in computer vision, leveraging deep learning and generative models to modify an image's visual style while preserving its intrinsic content structure \cite{jing2019neural}. The core of style transfer lies in disentangling an image's \textit{content} (the structural elements and identities of objects) from its \textit{style} (aesthetic attributes like color and texture) \cite{frenkel2024implicit, men2022dct}. While this technique has found applications across various creative domains, its adoption in high-stakes industrial settings, particularly marketing and advertising, faces significant practical hurdles \cite{al2024artistic, banar2021transfer}.

For these commercial applications, preserving the original image's core content is not merely a desirable feature but a critical requirement. A prominent example is in advertising campaigns featuring celebrity endorsers, where brand identity and trust are tied to the recognizable face of an individual. In this context, the failure to preserve identity during stylization can neutralize a campaign's effectiveness and undermine significant financial investment. This issue becomes particularly challenging in real-world scenarios, such as on billboards or in digital banners, where faces may be distant from the camera or part of a group, resulting in small facial regions that are highly susceptible to distortion.

The practical deployment of style transfer at an industrial scale is further hindered by two primary technical challenges inherent in diffusion models. The first is the number of inference steps. The iterative nature of these models requires many steps to generate high-fidelity images \cite{song2020denoising}, leading to high computational costs. This is a major bottleneck for industrial systems that require fast, on-demand generation for a large user base. The second and more critical challenge for the aforementioned applications is \textbf{Identity Preservation}. While recent methods have aimed to improve content fidelity \cite{mokady2023null, garibi2024renoise} or inject identity information \cite{ye2023ip, wang2024instantstyleplus}, they often struggle to maintain the distinct facial characteristics of individuals, especially under the challenging conditions of small or distant faces.

This paper introduces a novel, training-free framework that directly tackles these barriers to the industrial adoption of style transfer. We propose a solution designed to maintain robust identity integrity, even in challenging scenarios, without sacrificing computational efficiency. Our key contributions are:
\begin{itemize}
    \item A \textbf{Mosaic Restored Content Image} technique that enhances small or distant faces before stylization to ensure their identity is robustly preserved.
    \item A training-free \textbf{Content Consistency Loss} that refines the diffusion process to protect fine-grained content details during the stylistic transformation.
\end{itemize}

Our approach provides an effective and efficient solution for generating high-quality stylized images that faithfully preserve subject identity, paving the way for reliable use in demanding, real-world applications. The remainder of this paper is organized as follows: Section 2 reviews related work, Section 3 details our proposed methodology, and Section 4 presents our experimental results and analysis.


\section{Related Work}

\textbf The field of artistic style transfer has undergone significant transformations due to the rapid advancement of generative AI techniques.  A notable example is the transfer of image styles \cite{gatys2016image,lai2017deep,li2017universal,li2018closed,park2019arbitrary,wang2020diversified,zhang2022domain} in recent years, where the appearance of an image is altered by incorporating the artistic features of one or more reference images.  The aim is to merge the style elements of the source images with the content of the target image to create stylized output. 
Based on advances in generative AI models, style transfer has evolved into three primary domains: Neural Style Transfer (NST) \cite{gatys2016image}, Generative Adversarial Networks (GANs) \cite{goodfellow2014generative}, and Diffusion Models \cite{ho2020denoising}. In the following, we will explore the evolution of these approaches, highlighting their strengths and limitations, and ultimately focusing on the domain most relevant to our work: diffusion-based style transfer.

\subsection{Neural Style Transfer (NST)} 

The traditional Neural Style Transfer (NST) approach to image style transfer relies on non-parametric texture transformation methods \cite{hertzmann2001image,efros1999texture} but often struggles with complex images due to its reliance on fundamental, rather than abstract, feature extraction. Gatys et al.\cite{gatys2016image}represent the style and content of an image as two distinct parts by combining convolutional neural networks (CNN) with texture generation methods. In this method, the high-level convolutional layers of the CNN capture the image's content, while the low-level layers represent the style. However, this approach faces challenges such as slow convergence, and insufficient stylistic output. Johnson et al.\cite{johnson2016perceptual} improved the efficiency of style transfer by employing feedforward networks with perceptual loss functions and a pretrained VGG model. However, similar to \cite{gatys2016image}, their approach faced difficulties in achieving realistic results due to the limitations of the Gram matrix in style extraction.

\subsection{GANs}

GANs have become fundamental to artistic stylization due to their ability to capture high-level data patterns. They are widely applied in various tasks, including painting \cite{gao2020rpd}, cartooning \cite{gao2022learning}, and font stylization \cite{xie2021dg}.  More GAN-based approaches \cite{goodfellow2014generative,isola2017image,katzir2020crossdomain,park2020contrastive,zhu2017unpaired}integrate paired \cite{isola2017image} and unpaired \cite{katzir2020crossdomain,park2020contrastive,zhu2017unpaired} datasets to improve model flexibility and generalization. The early methods relied on conditional GAN architectures, such as Pix2Pix \cite{isola2017image}. These methods are supervised, meaning they need paired examples of input and output images. In contrast, CycleGAN \cite{zhu2017unpaired} proposes an unsupervised image-to-image translation model that learns bidirectional mappings between two domains using adversarial loss and cycle consistency. This approach enables effective style transfer without the need for paired training data.
DualStyleGAN \cite{yang2022Pastiche} proposed a dual-path model that allows control of both structural and artistic styles. Furthermore, DCT-Net \cite{men2022dct} introduced a few-shot portrait stylization model that enhances style transfer by first calibrating the content distribution and then applying localized translation.


\subsection{Diffusion Models}

Recent advancements in diffusion models have revolutionized image stylization, enabling sophisticated and adaptable style transfer techniques \cite{chen2025comprehensive}.  Some methods leverage textual inversion, with InST  \cite{zhang2023inversion} using textual inversion to encode styles as text embeddings to offer more flexible style representation. Stylediffusion \cite{wang2023stylediffusion} introduced a CLIP-based loss for better style-content disentanglement during diffusion model training. Furthermore, several approaches utilize text as a conditioning mechanism for style control or content generation  \cite{yang2023zero,everaert2023diffusion}. Training-free methods, such as   DiffStyle \cite{yang2022Pastiche} modify skip connections within h-space \cite{kwon2023diffusion} to balance style and content preservation. However, when used with Stable Diffusion \cite{vonplaten2023diffusers}, it can sometimes change textures and the overall layout of the image in ways that aren't intended. A major challenge within style transfer maintaining the integrity of the original content; for instance, a portrait intended to resemble a painting must still depict the same person.
StyleID \cite{Chung_2024_CVPR} tries to solve this by adjusting how the image is processed and using some clever tricks to preserve the original content with query preservation and initial latent AdaIN.

inversion-based models like StyleAlign \cite{hertz2023style}which uses shared self-attention and Adaptive Instance Normalization (AdaIN)\cite{huang2017arbitrary}   to get an accurate style transfer but these methods may struggle with style-content leakage.
 More recent works, like IP-Adapter \cite{ye2023ip}  and Style-Adapter \cite{wang2023styleadapter} offer greater control and flexibility in style transfer by using a specialized cross-attention mechanism that processes text and image features in separate attention layers.
 Despite their advantages, encoder-based methods like IP-Adapter face challenges in differentiating style from content.  Therefore, ControlNet \cite{zhang2023adding} is employed to effectively regulate and maintain the content of the generated image, ensuring that the content remains intact while applying the desired stylistic transformation. InstantStyle \cite{wang2024instantstyle} enhances IP-Adapter for image stylization by incorporating the CLIP embedding of the style images into targeted blocks within SDXL. This method effectively separates content and style, enabling a more precise and adaptable style transfer process. InstantStyle-Plus \cite{wang2024instantstyleplus} introduces a novel diffusion-based style transfer method, employing a tile-based ControlNet, semantic adapter, and style extractor to improve content preservation and style fidelity. This study demonstrates that by utilizing specific components, the performance of diffusion models in style transfer can be significantly enhanced.

\section{Methodology}

\begin{figure*}[ht]
    \centering
    \includegraphics[width=1\linewidth]{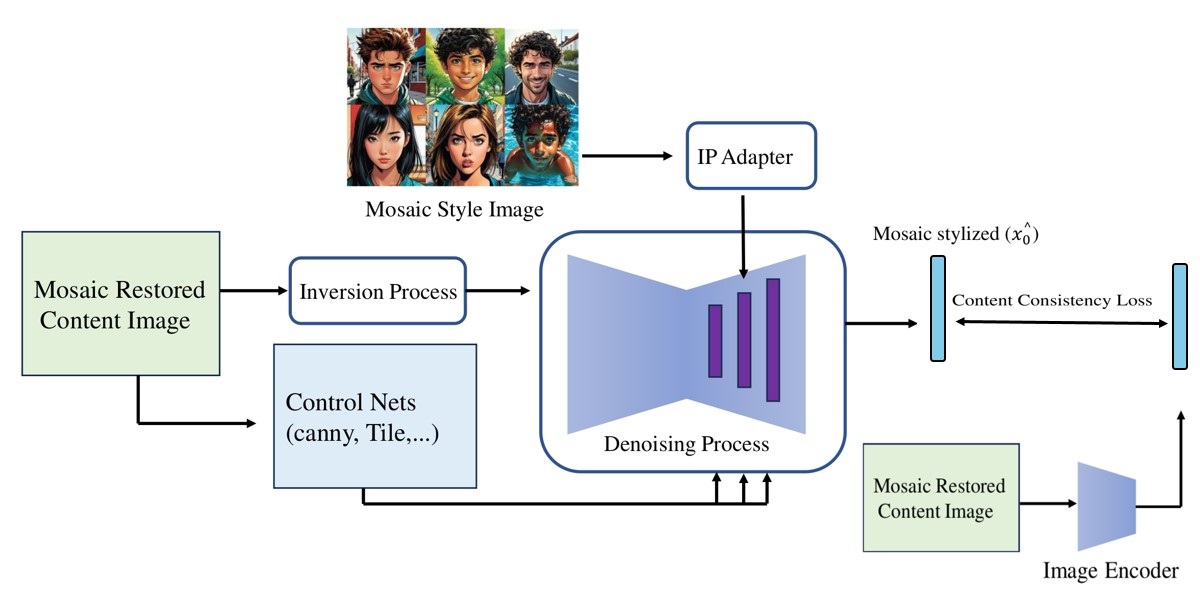}
    \caption{Overview of our proposed style transfer pipeline. Our method builds upon the InstantStyle-Plus pipeline \cite{wang2024instantstyleplus} by incorporating a Mosaic Restored Content Image to preserve identity in images. To further enhance identity preservation and detail fidelity, we introduce Content Consistency Loss, which refines the denoising process.}
    \label{fig:1}
\end{figure*}
In the Introduction section, we identified two key challenges in the style transfer process. The first challenge is the need for a high number of inference steps to achieve high-quality image synthesis with diffusion models. This results in increased computational demands and longer processing times, which may limit their practicality in industrial settings.
While this issue can be partially addressed by employing the pipeline introduced in \cite{wang2024instantstyleplus} along with inversion algorithms \cite{garibi2024renoise}, it leads to a secondary problem: the failure to preserve the identity and details of individuals in the scene when the number of inference steps is reduced.
In our research, we investigated the impact of facial size and the number of subjects on identity preservation, particularly focusing on scenarios where faces are small or images contain multiple individuals.
Our experiments reveal that the identity preservation performance of the base pipeline is  dependent on the size and quantity of faces within the image. Specifically, when faces are sufficiently large, individual identity details are well maintained; however, when the facial regions are small or the image contains multiple faces, a significant loss of distinctive identity features is observed.
To address this issue, we propose a solution that incorporates face detection and image enhancement techniques. Specifically, we used yolo \cite{varghese2024yolov8} face detection model to identify the faces of individuals in the scene. Subsequently, we employ the Real-ESRGAN  model for facial enhancement \cite{wang2021real} to improve the clarity and quality of the detected faces. In \cite{rezaei2025you}, it was observed that using image restoration techniques can significantly enhance image segmentation by making objects more clearly defined. Therefore, to increase the overall image quality, we incorporate this technique into our pipeline.
The next step involves creating a mosaic image that contains the enhanced face images and the original background.
This mosaic image is used as an input to the style transfer pipeline, ensuring that facial identities are preserved.
In the final output, the stylized face images are extracted from the stylized mosaic image and placed back into their original positions in the scene, keeping the integrity of the content image  (see Figure~\ref{fig:fig1}).

This approach, which utilizes straightforward computational methods, offers a robust solution to the identity preservation issue; specifically, it significantly improves the retention of facial identities in stylized output.

\subsection{Enhancing the Denoising Process to Improve Content Preservation}

In image stylization using Stable Diffusion, a key challenge is balancing the transfer of style with the preservation of content details, such as identity and structure.
In text-to-image models like SDXL, which are trained on billions of images across diverse styles, the text prompt plays a pivotal role in determining the output's stylistic domain. Consequently, when employing targeted style keywords to generate images in a specific style, it becomes necessary to increase the model’s attention to the text prompt. However, this heightened focus on style can inadvertently compromise content fidelity, leading to the loss of critical details, particularly the identity of objects or individuals. While adding control networks like Canny, tile, or depth maps can improve detail retention, they also increase the model's size, slow down processing, and consume more memory. Despite this, even with additional control networks, the model still struggles to preserve details effectively.

To address this problem, we propose a refined denoising process for latents in VAE space that preserves the identity and structure of the content while still allowing effective style transfer. The key idea is to refine the predicted noise at each timestep and ensure that the latents remain close to their original content representation. This technique involves denoising the initial latent by refining the predicted noise at each step, allowing us to maintain a balance between content and style. Our method focuses on calculating the loss between the stylized image (created with the predicted noise) and the original content image, both of which exist in VAE space. We then compute the gradient of this loss with respect to the predicted noise and use this gradient to iteratively refine the noise, bringing the latents closer to their initial latent content.

The process begins with the generation of an initial stylized image in the VAE space, where we compute the L2 distance between the stylized image and the original content image. This loss guides the model to preserve the details of the content while applying the style. The gradient of this loss is used to adjust the predicted noise, which is defined in multiple steps.

\subsection{Mathematical Formulation}

Given the latent representation \( z_{t}\) in the timestep \( t \), initialized from a content image via an inversion algorithm \cite{garibi2024renoise}, we estimate the denoised content representation using the following.

\begin{equation}
\hat{x}_0 = \frac{z_{t} - \sqrt{1 - \alpha_t} \cdot \hat{\epsilon}_t}{\sqrt{\alpha_t}}
\end{equation}
where \( \hat{x}_0 \) represents the estimated denoised image in VAE space, \( \hat{\epsilon}_t \) is the predicted noise at timestep \( t \), and \( \alpha_t \) is the cumulative product of noise schedule parameters.
To ensure that the stylized output remains close to the original content, we define the content consistency loss as:

\begin{equation}
\mathcal{L}_{\text{content}} = \| \hat{x}_0 - x_c \|_2^2
\end{equation}
where \( x_c \) is the original content latent representation in VAE space.
To refine the noise prediction, we compute the gradient of the content consistency loss with respect to the predicted noise:

\begin{equation}
\frac{\partial \mathcal{L}_{\text{content}}}{\partial \hat{\epsilon}_t} = 2 (\hat{x}_0 - x_c) \cdot \left( - \frac{\sqrt{1 - \alpha_t}}{\sqrt{\alpha_t}} \right)
\end{equation}
Using a hyperparameter \( \lambda_c \) to control the refinement strength, the refined predicted noise is computed as:

\begin{equation}
\hat{\epsilon}_t^{\text{refined}} = \hat{\epsilon}_t - \lambda_c \cdot \frac{\partial \mathcal{L}_{\text{content}}}{\partial \hat{\epsilon}_t}
\label{eq:refined_noise}
\end{equation}
where \( \lambda_c \) is a tunable hyperparameter that determines the strength of the content preservation constraint.
Finally, the latents are updated using a deterministic DDIM scheduler step \cite{song2020denoising}:

\begin{equation}
z_{t-1} = \sqrt{\alpha_{t-1}} (\frac{z_{t} - \sqrt{1 - \alpha_t} \cdot \hat{\epsilon}_t^{\text{refined}}}{\sqrt{\alpha_t}}) + \sqrt{1 - \alpha_{t-1}} \hat{\epsilon}_t^{\text{refined}}
\end{equation}
By iteratively refining the noise prediction at each timestep, this approach preserves the integrity of the content while applying the desired style transformation. This novel refinement strategy controls the incorporation of style information in the stylized image while emphasizing detailed information in the output. An overview of our proposed pipeline, integrating this process, is shown in Figure~\ref{fig:1}.

\section{Experiments}

In this chapter, we present an experimental evaluation of our proposed method.
To implement our ideas, we utilized the InstantStyle-Plus pipeline \cite{wang2024instantstyleplus} as our base, built upon the Stable Diffusion XL (SDXL 1.0) model, and incorporated several modifications.
For all stylization experiments, the diffusion process (inference) was performed for \textbf{10 denoising steps}.
The initial latent representation of the content image (or mosaic restored content image) was obtained using the inversion algorithm from \cite{garibi2024renoise} with \textbf{6 inversion steps}.
The choice of these relatively low step counts was made to evaluate the efficiency and identity preservation capabilities of our method under computationally constrained conditions, a key challenge highlighted earlier.
The details of our other modifications and the experimental setup will be discussed in the following sections.

\subsection{Effect of Image Captioning on Identity Preservation}

\begin{figure}[ht]
    \centering
    \includegraphics[width=0.85\linewidth, height=0.35\textheight]{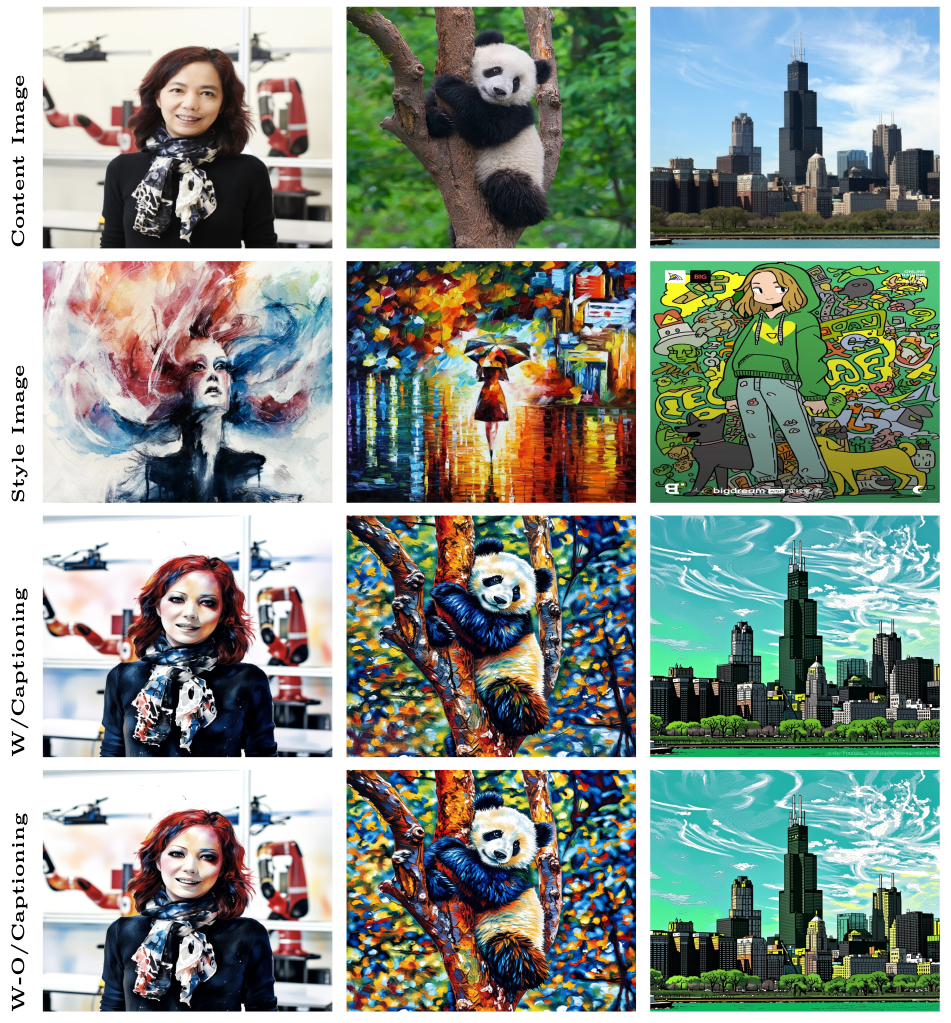}
    \caption{\textbf{Comparison with and without image captioning.} 
      Rows: (1) Content Image, (2) Style Image, (3) Stylized Output with Captioning, (4) Without Captioning. 
      Columns show different samples. Our method yields similar results under both settings.
      }\label{fig:fig5}
\end{figure}

\begin{figure}
    \centering
    \includegraphics[width=0.80\linewidth, height=0.20\textheight]{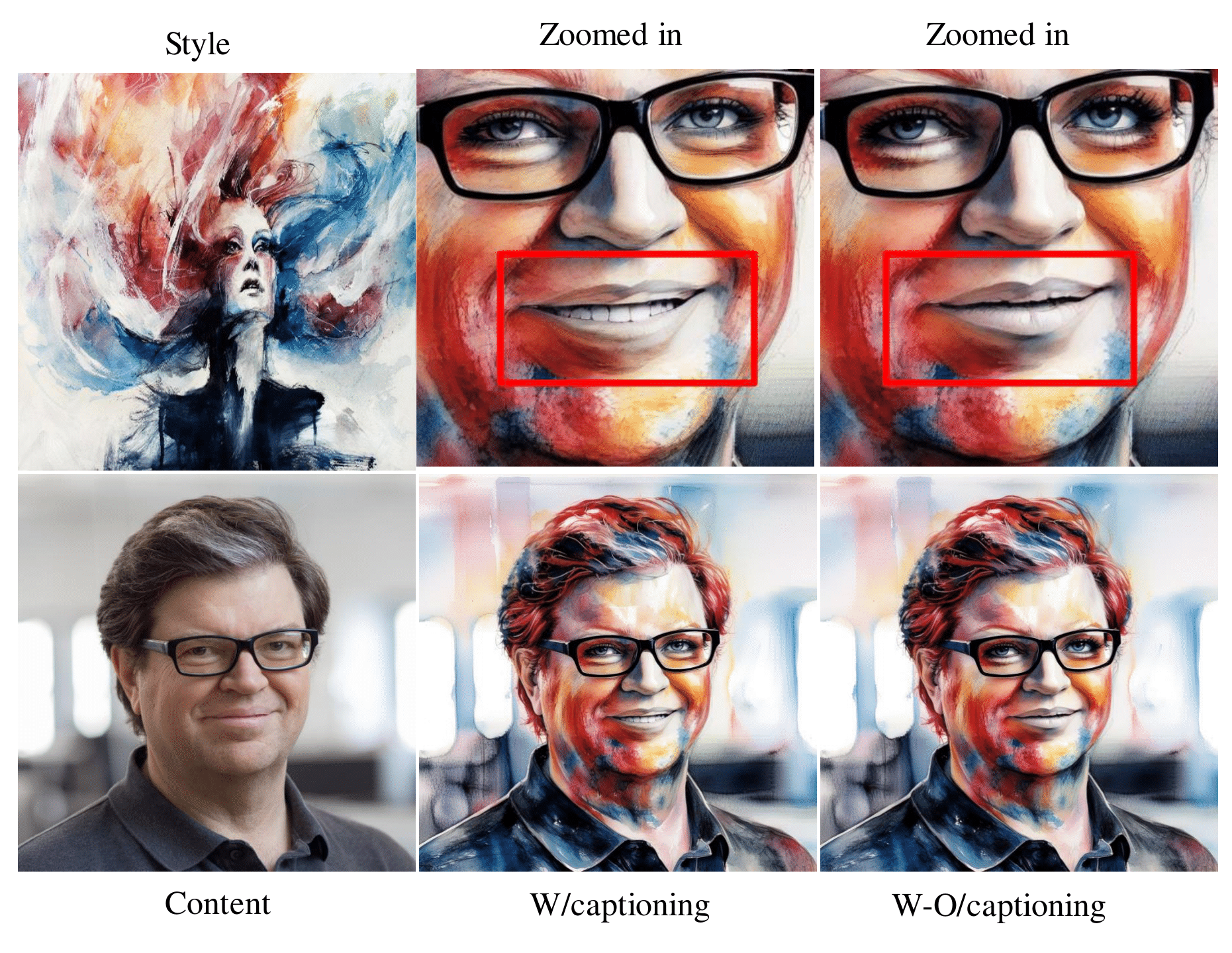}
    \caption{  \textbf{Effect of image captioning.} Top row: Style image and zoomed-in stylized results with and without captioning.  
  Bottom row: Content image and full stylized outputs.  
  The BLIP-generated caption “a man wearing glasses is smiling in an office” misleads the model and causes incorrect teeth generation.
  }\label{fig:fig6}
\end{figure}

During our evaluation of the InstantStyle-Plus pipeline, we observed that the base pipeline employs BLIP image captioning \cite{blip} to generate text prompts to reconstruct the structural details of the content image. Our experiments revealed that ControlNet modules, such as Canny and Tile ControlNet, can effectively preserve the content structure without relying on these descriptive prompts Figure~\ref{fig:fig5}.
Although generated prompts are intended to guide structural reconstruction, our findings show that using image captioning can sometimes lead to the generation of misleading or incorrect visual elements, due to generalized or inaccurate textual descriptions. For example, in Figure~\ref{fig:fig6}, the caption generated by BLIP 'a man wearing glasses is smiling in an office' incorrectly suggests visible teeth, although the subject's teeth are not shown in the original image. This misrepresentation causes the model to hallucinate details, such as artificially adding teeth, leading to a loss of identity fidelity. Furthermore, using image captioning models increases both memory usage and inference time, despite producing results that are visually similar to those obtained without captioning.

\begin{figure}[ht]
    \centering
    \includegraphics[width=1\linewidth, height=0.35\textheight]{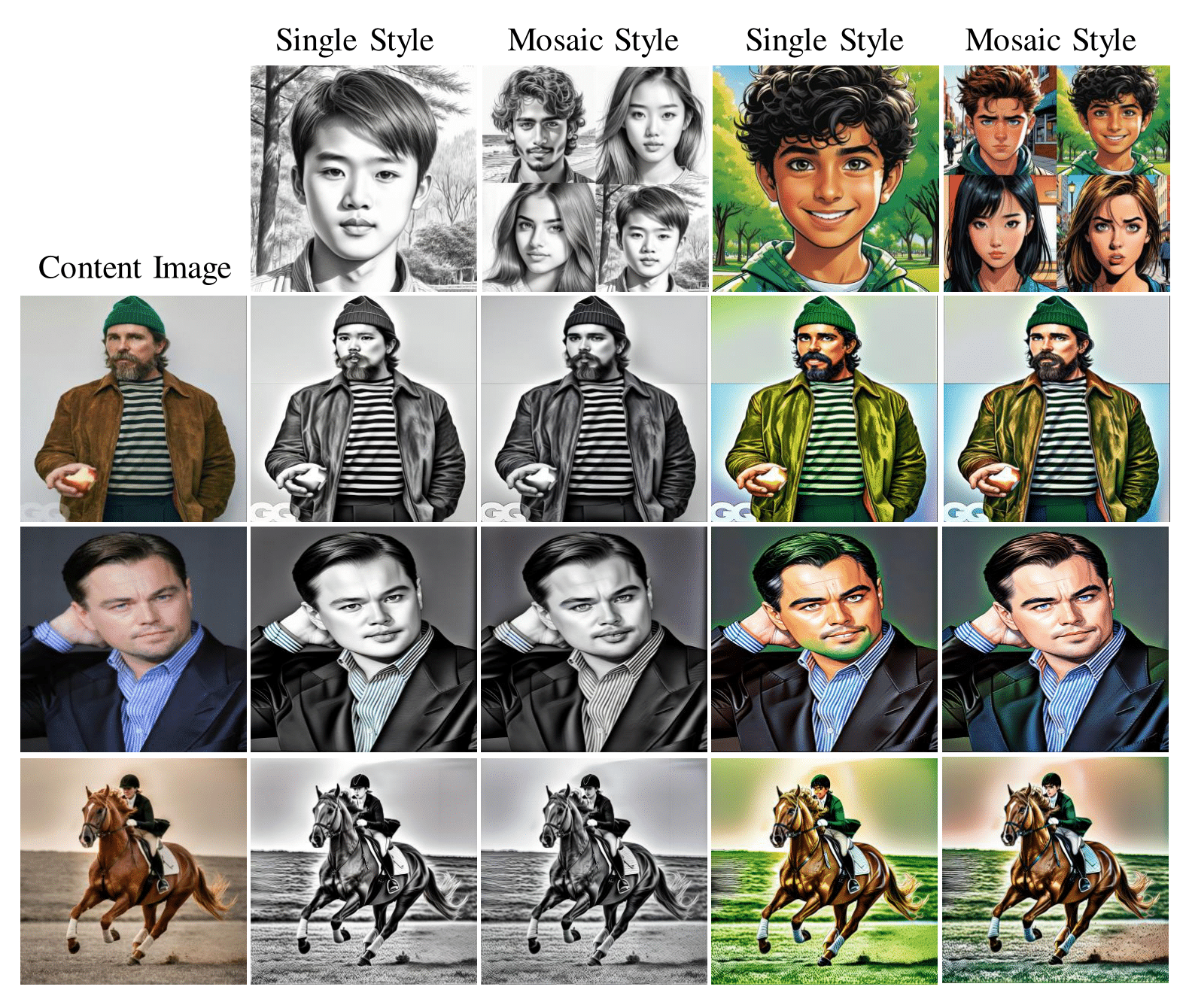}
    \caption{Comparison of Style Transfer Results Using Single vs. Mosaic-Style References. Using a mosaic-style reference composed of multiple images from the same stylistic domain reduces identity and background leakage compared to a single style reference, thereby enhancing facial identity preservation and achieving a more consistent color distribution.}
    \label{fig:2}
\end{figure}


\newcommand{\myimg}[1]{\includegraphics[width=1.6cm, height=1.6cm]{#1}}

\begin{table*}[ht]
\centering
\renewcommand{\arraystretch}{0.9}

\begin{tabular}{
  @{} >{\centering\arraybackslash}m{1.5cm}
  @{\hspace{0.7em}} 
  *{5}{>{\centering\arraybackslash}m{1.5cm}@{}}
  @{\hspace{0.7em}} 
  *{5}{@{}>{\centering\arraybackslash}m{1.3cm}@{}}
  @{}
}

\toprule

& \multicolumn{5}{c}{\textbf{Our Method}} & \multicolumn{5}{c}{\textbf{ISP}} \\

\cmidrule(lr){2-6} \cmidrule(lr){7-11}

{\textbf{Style   Content }} &

\myimg{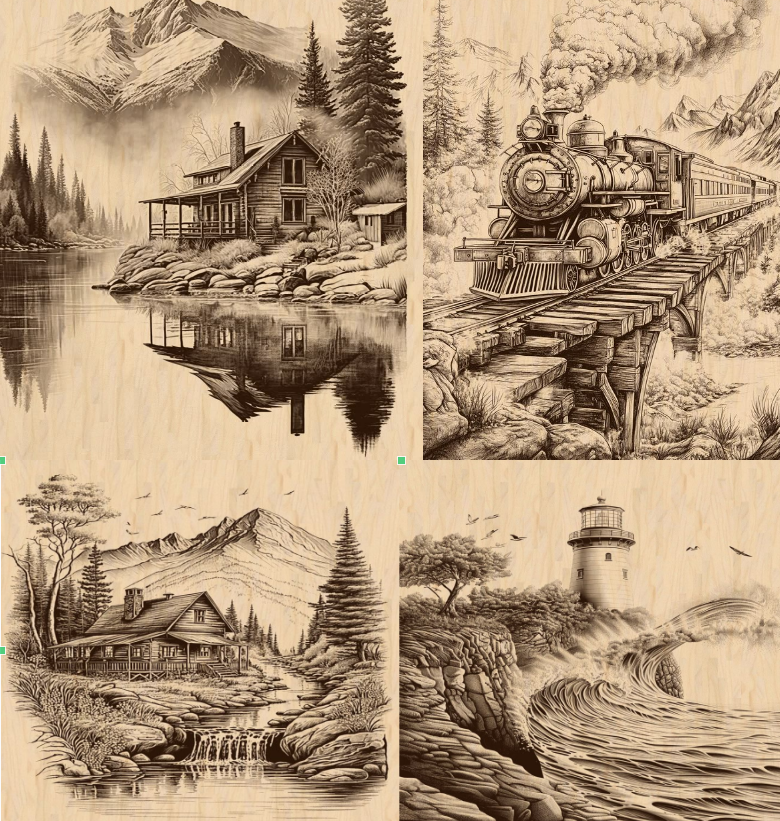} &
\myimg{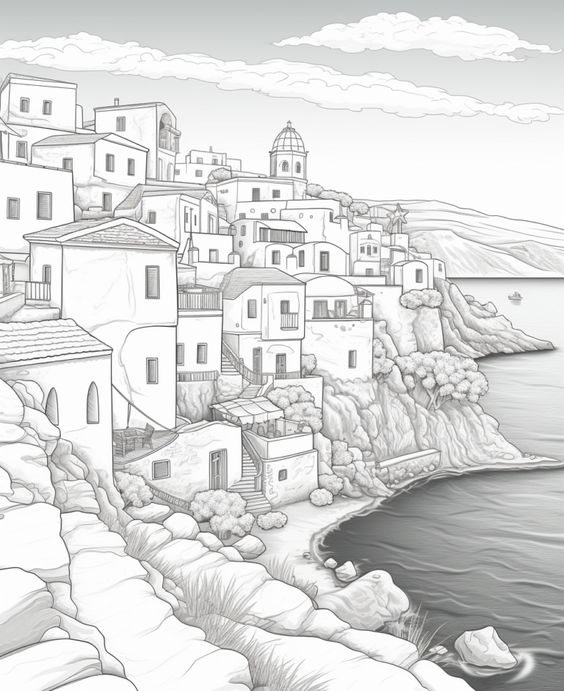} &
\myimg{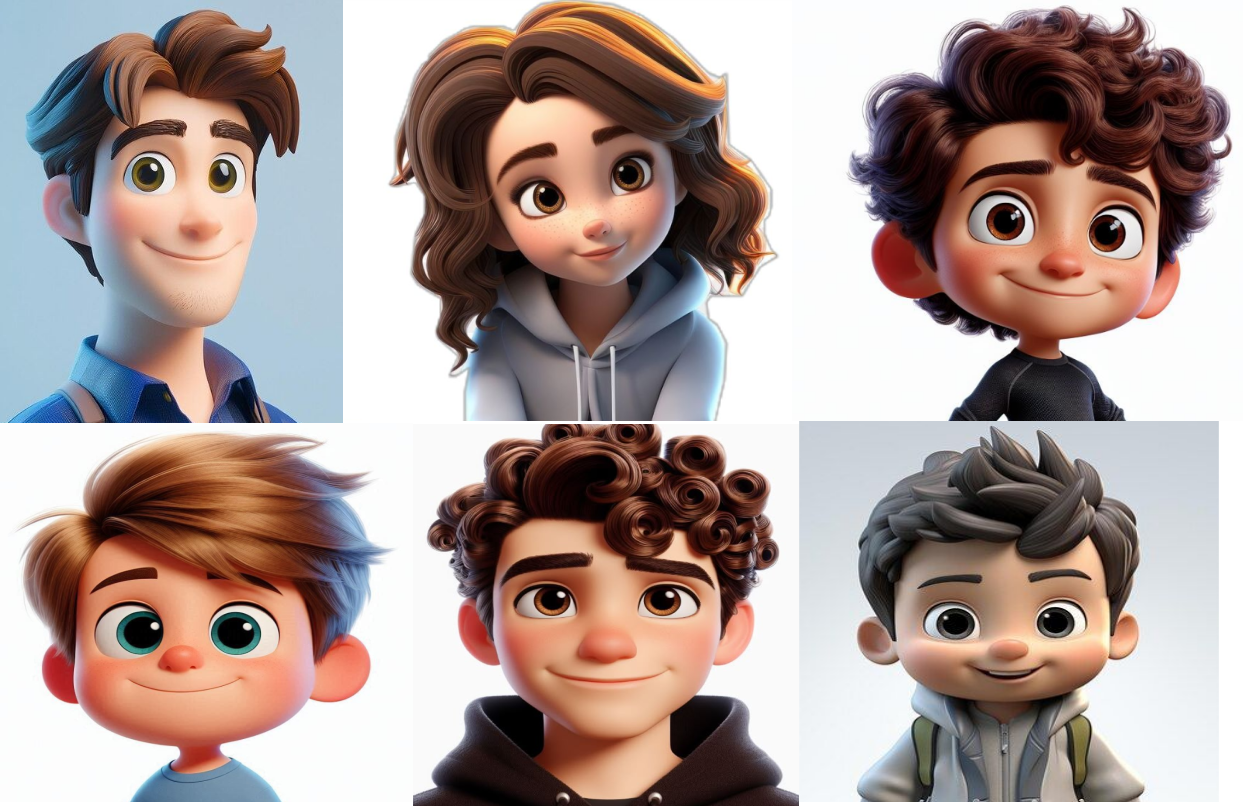} &
\myimg{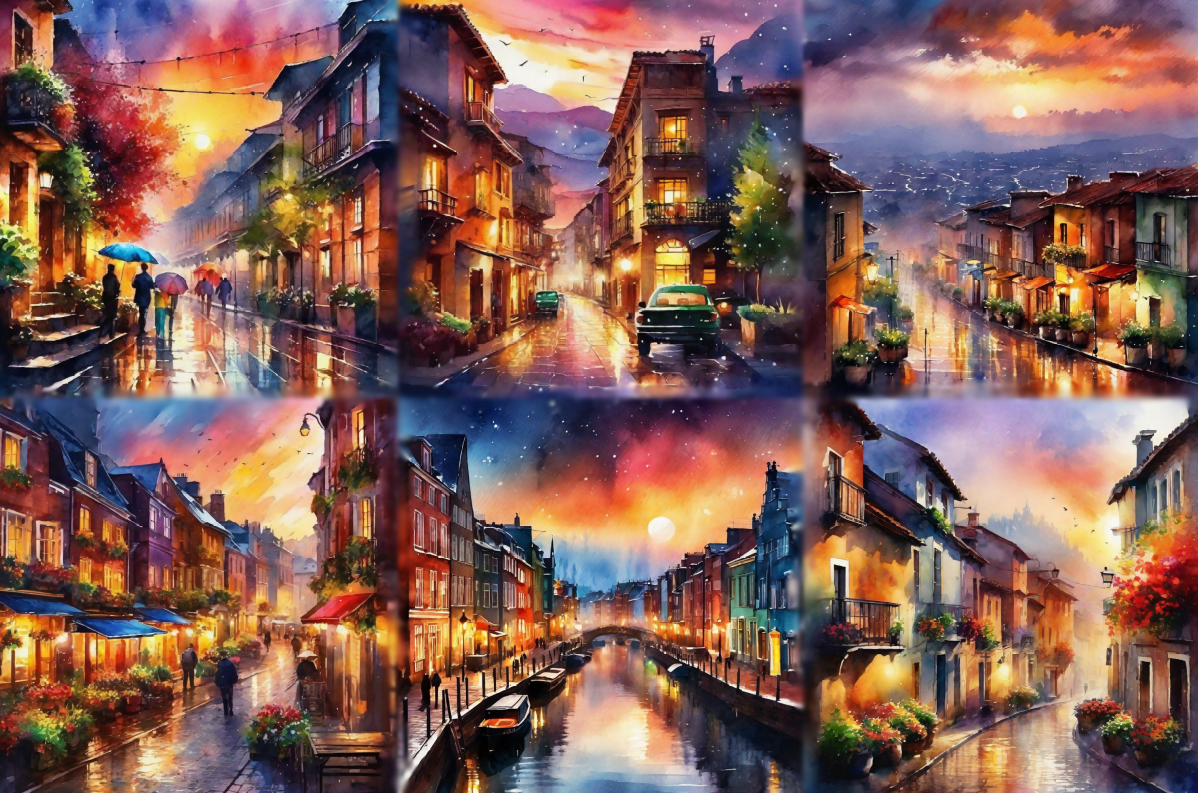} &
\myimg{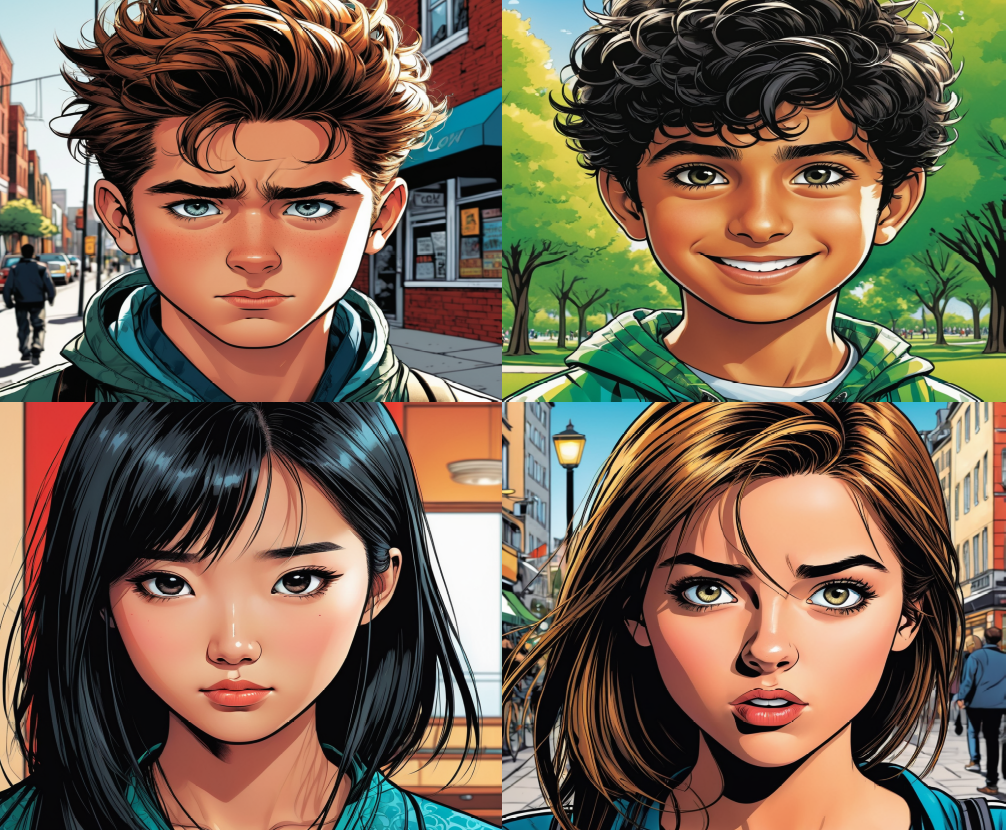} &
\myimg{base_img/style/sketch_gold.png} &
\myimg{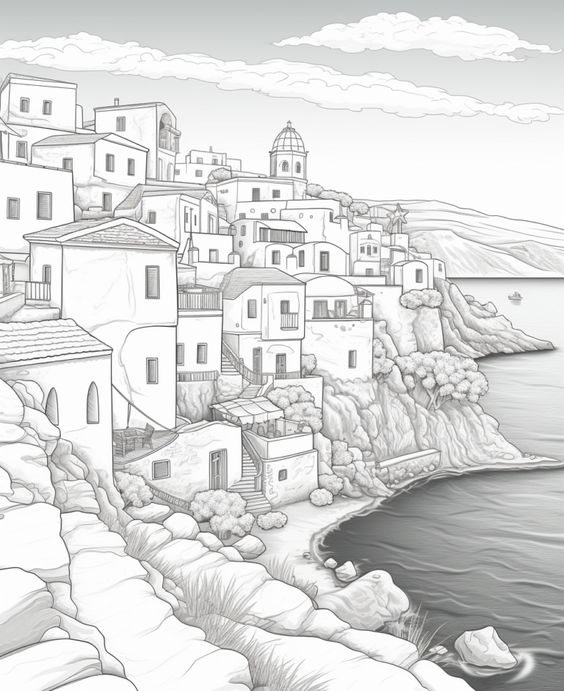} &
\myimg{base_img/style/anime.png} &
\myimg{base_img/style/watercolor.png} &
\myimg{base_img/style/comic.png} \\

\midrule 

\myimg{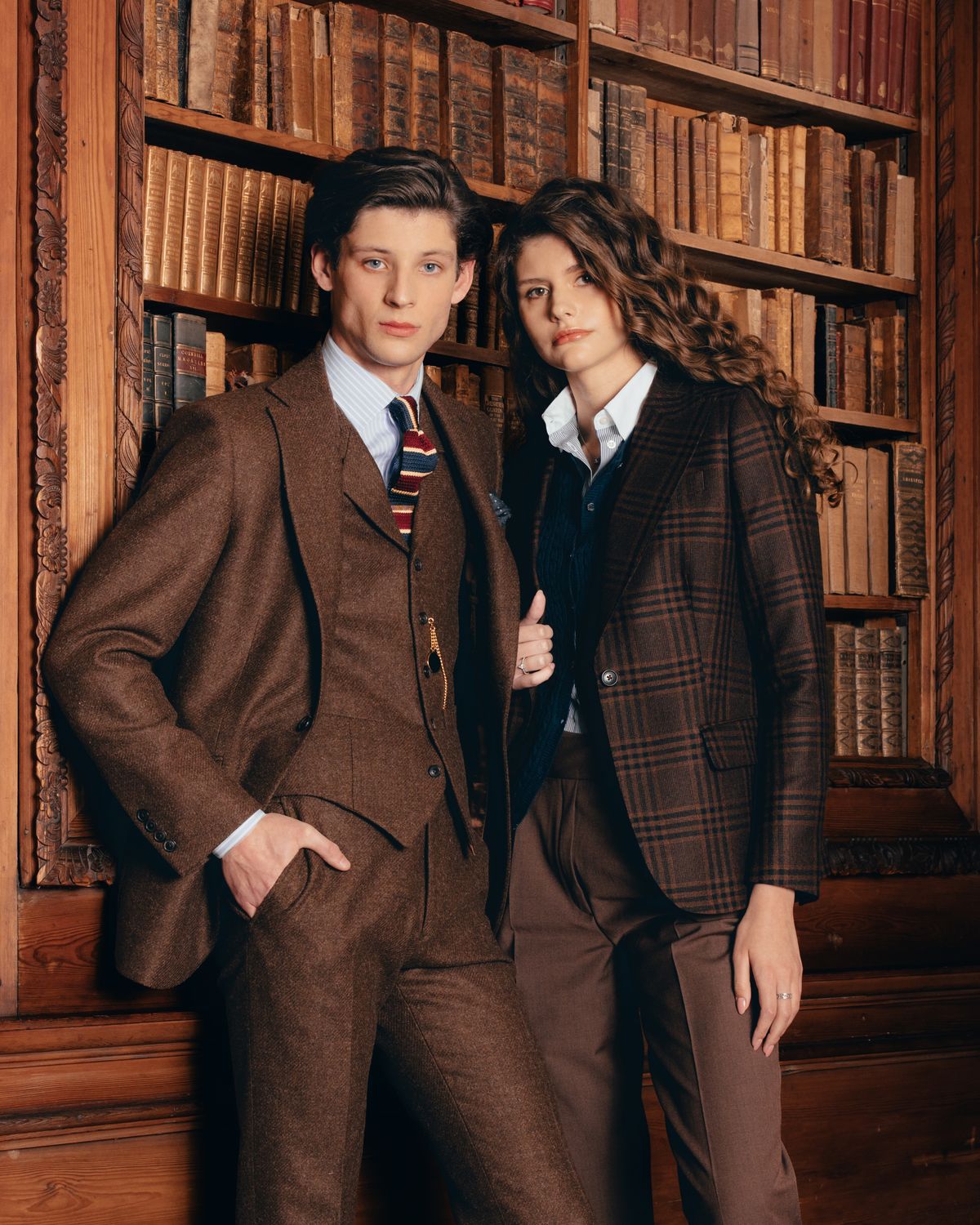} &
\myimg{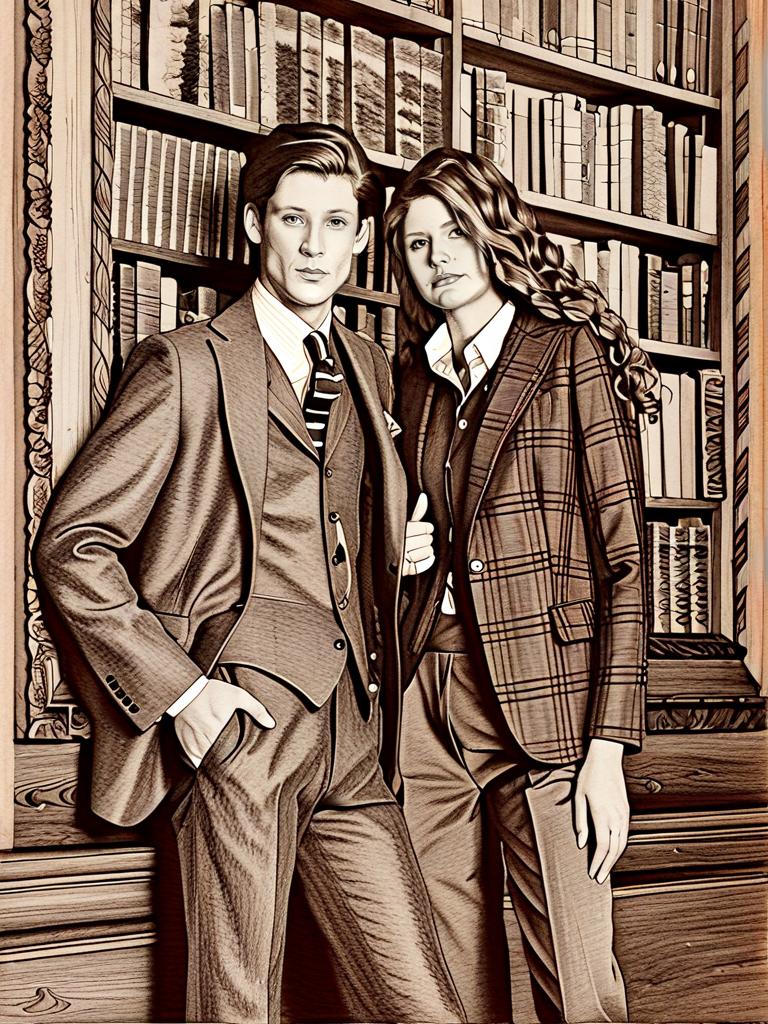} &
\myimg{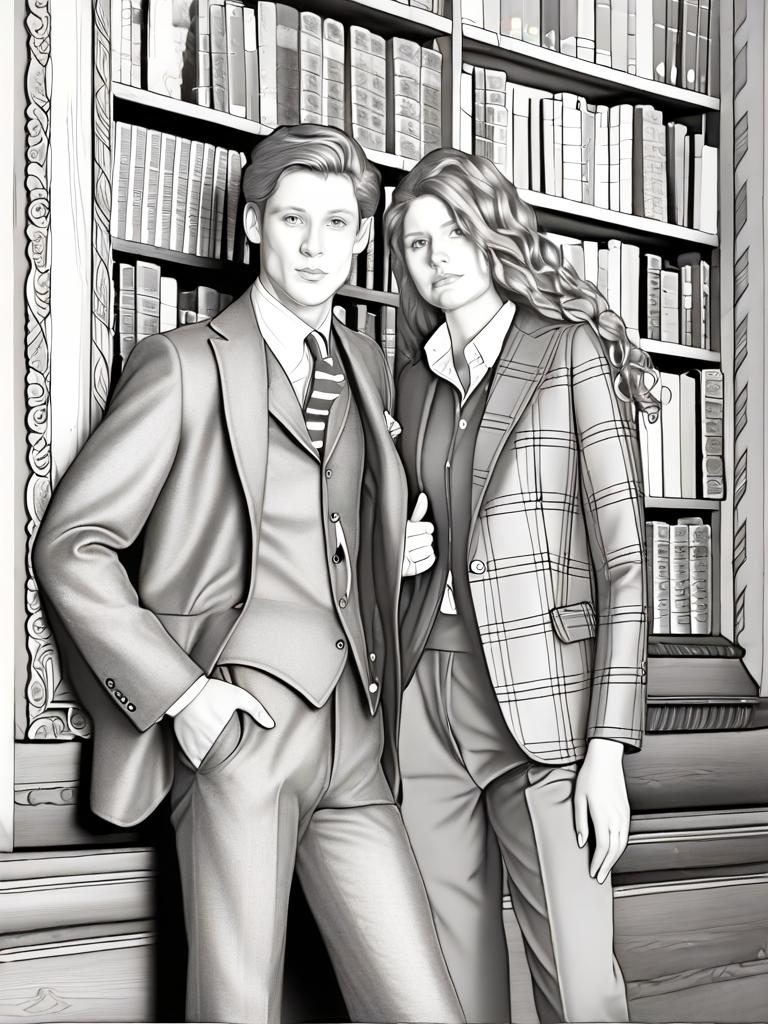} &
\myimg{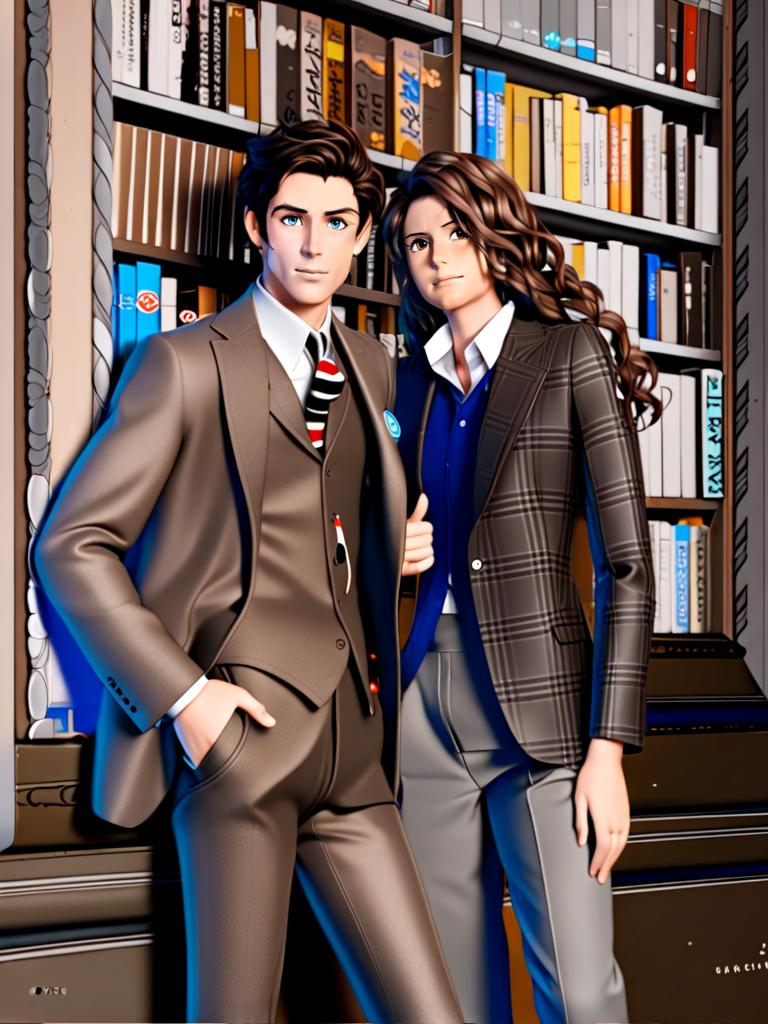} &
\myimg{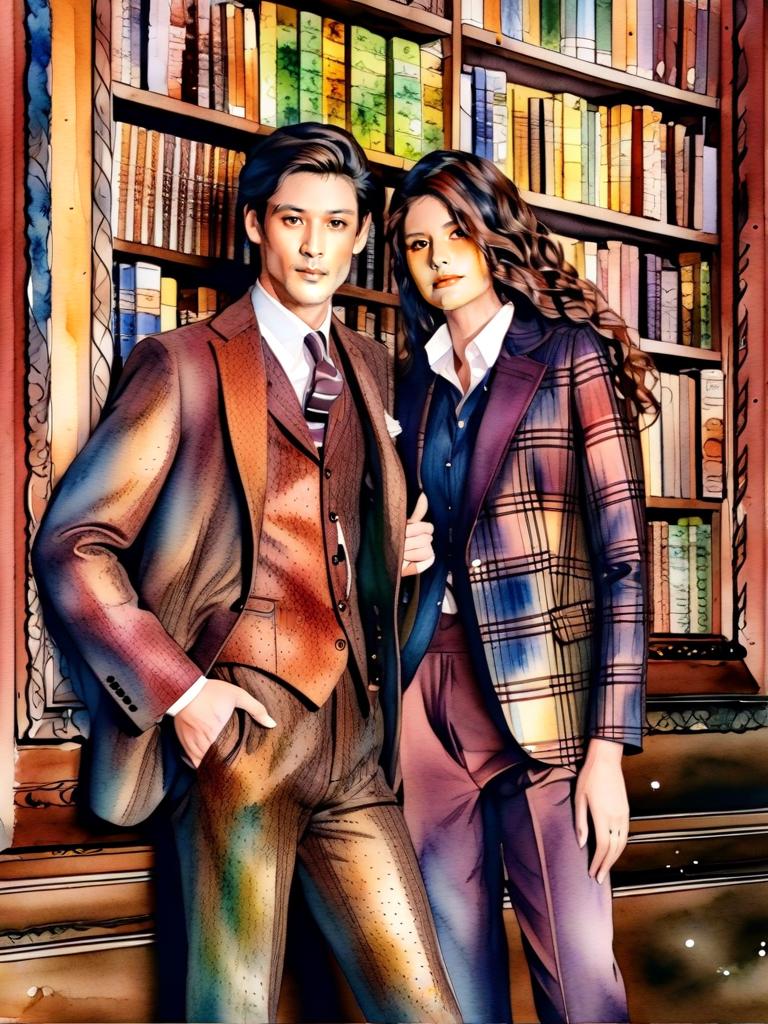} &
\myimg{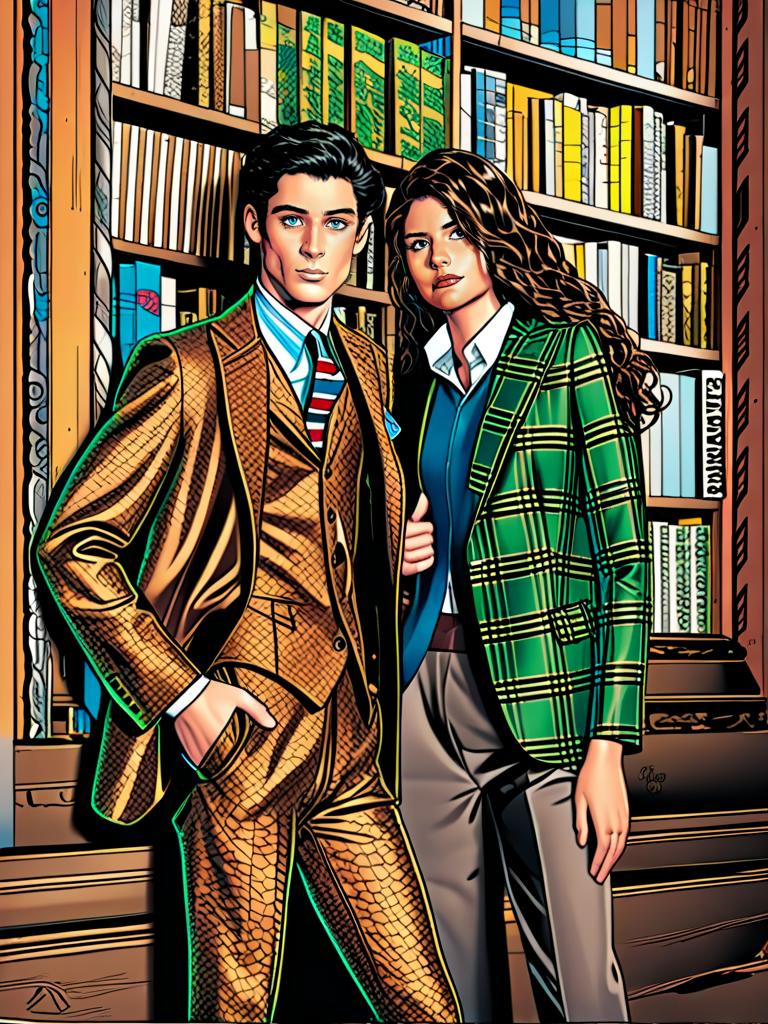} &
\myimg{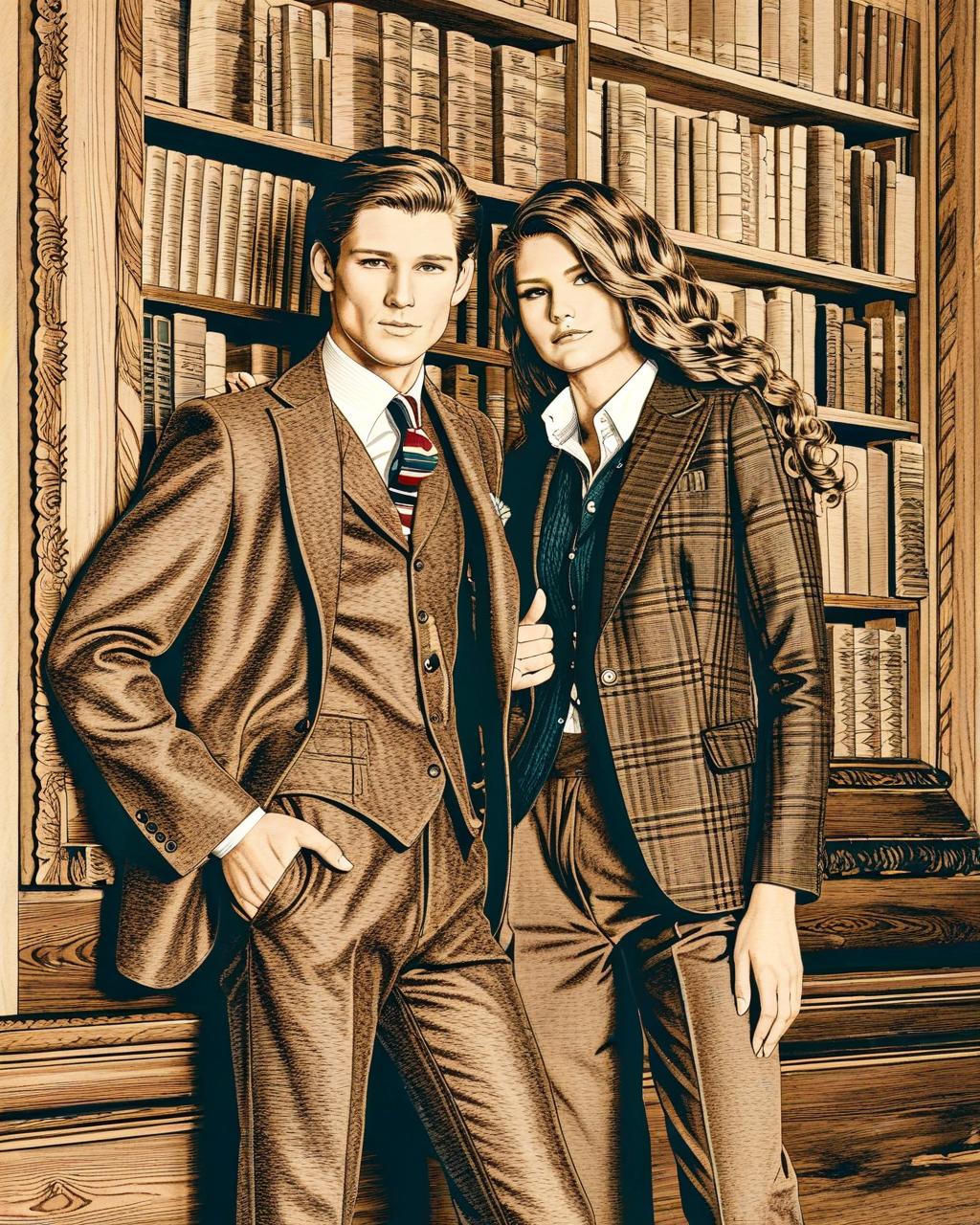} &
\myimg{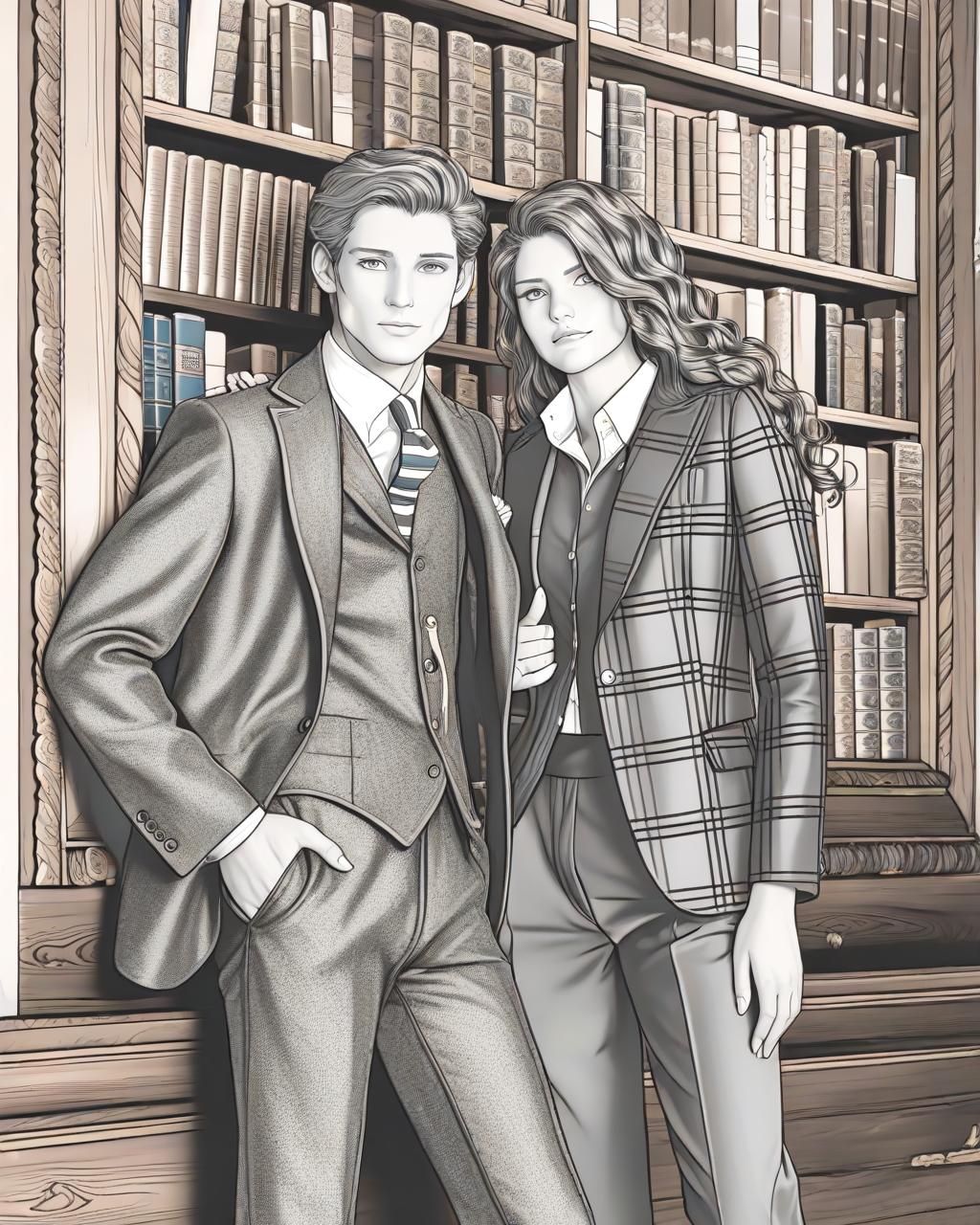} &
\myimg{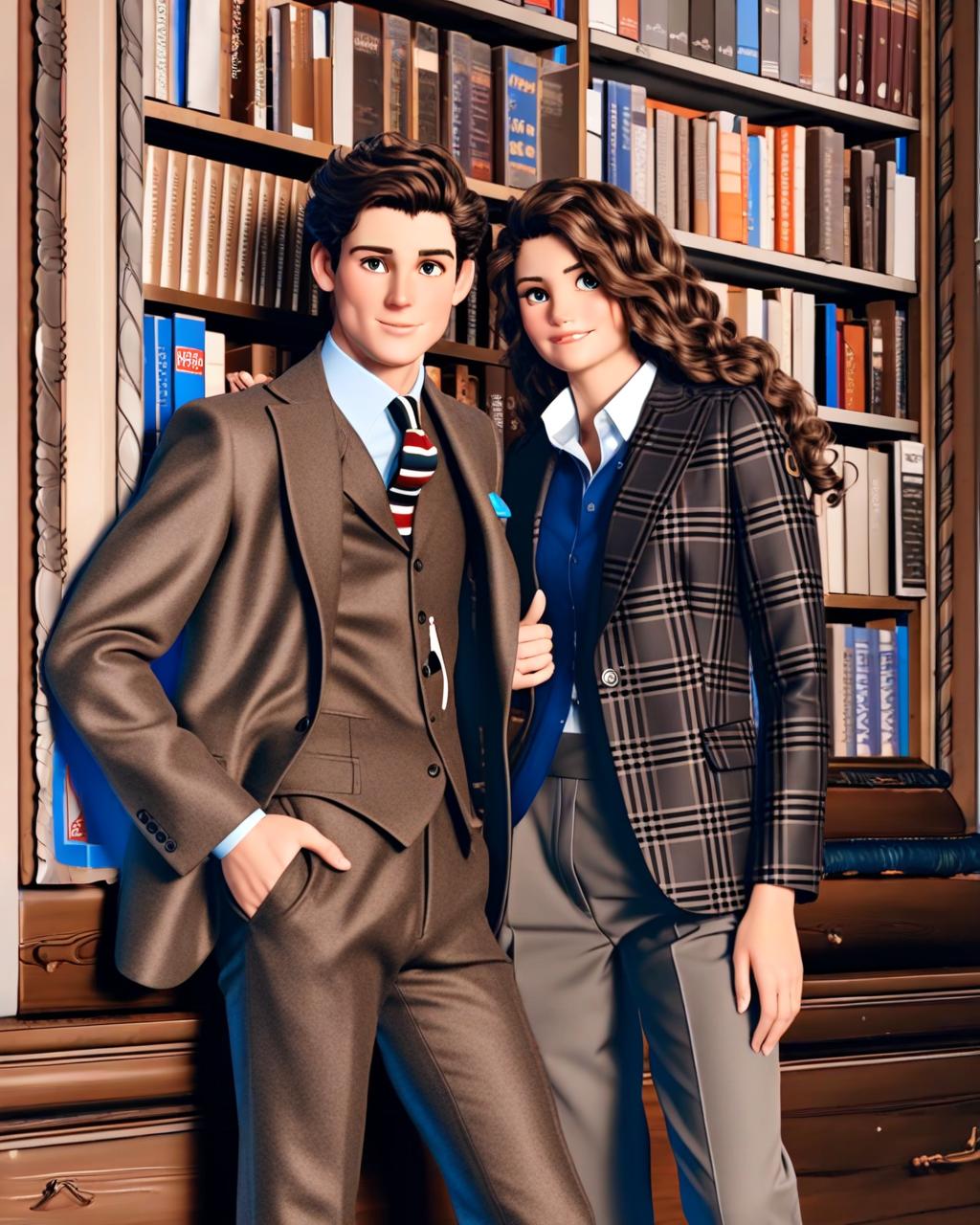} &
\myimg{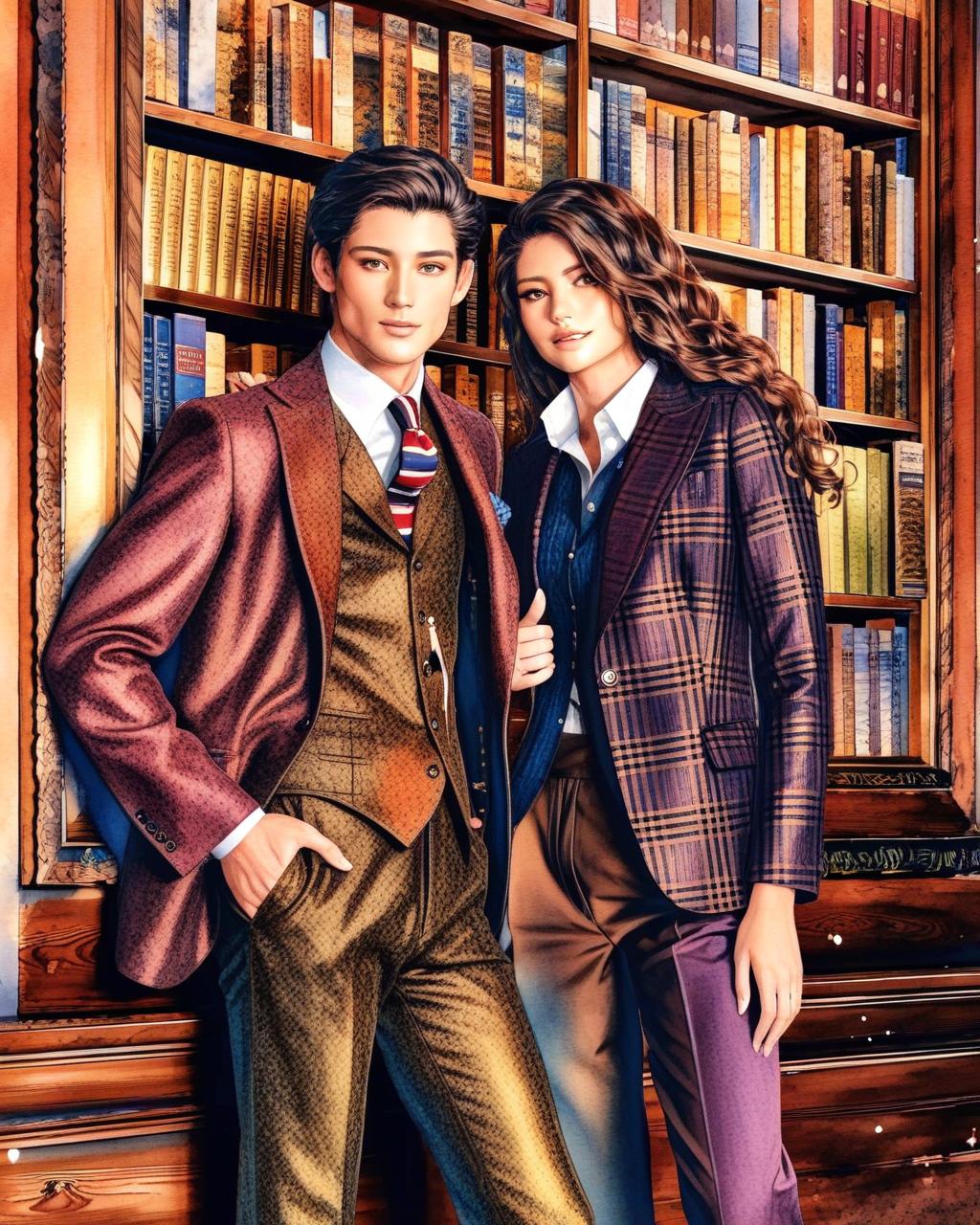} &
\myimg{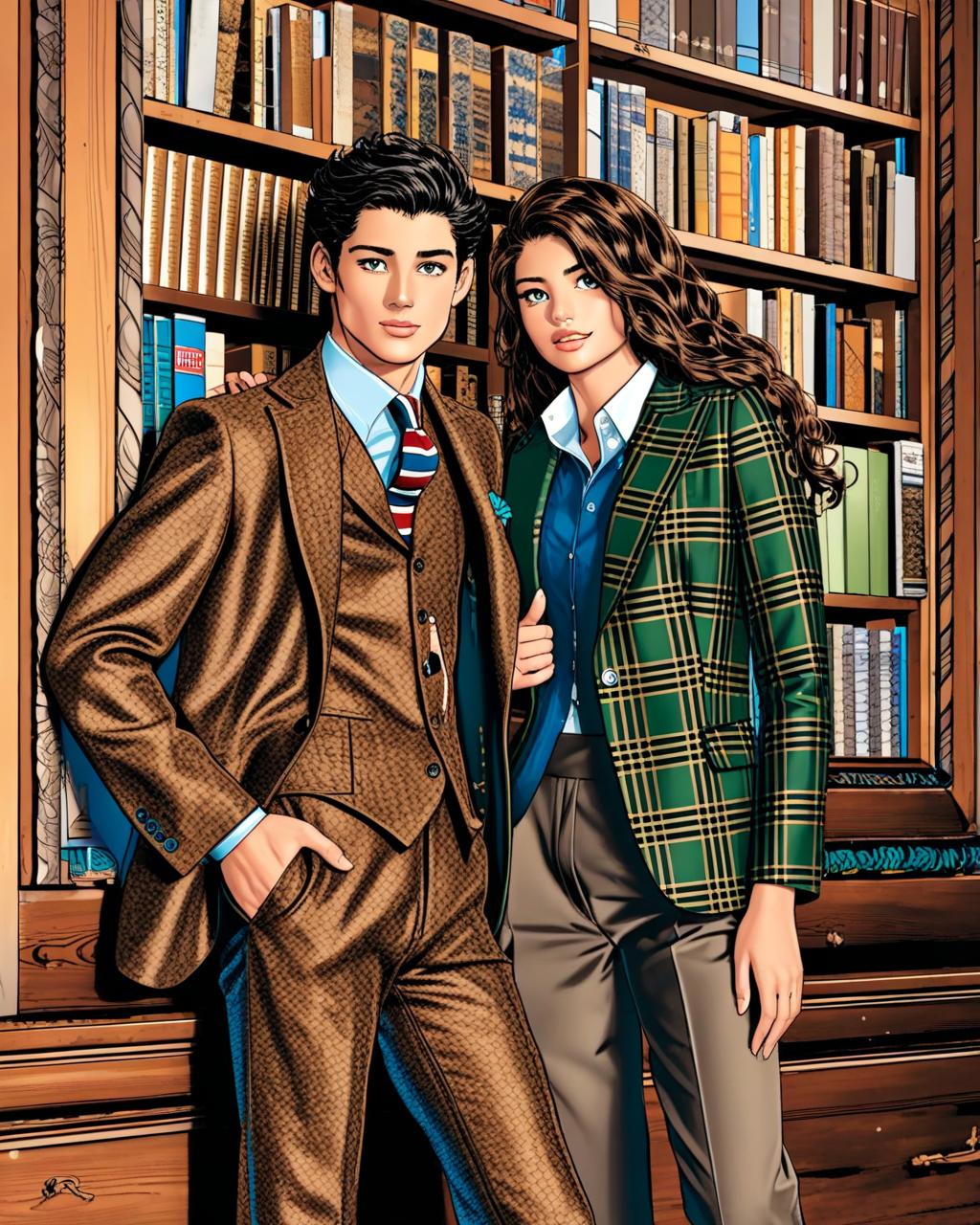} \\


\myimg{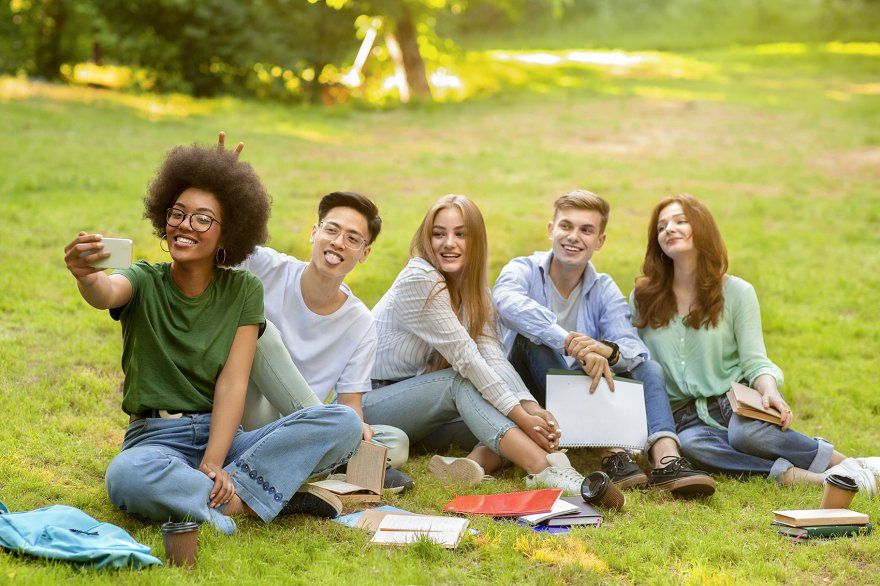} &
\myimg{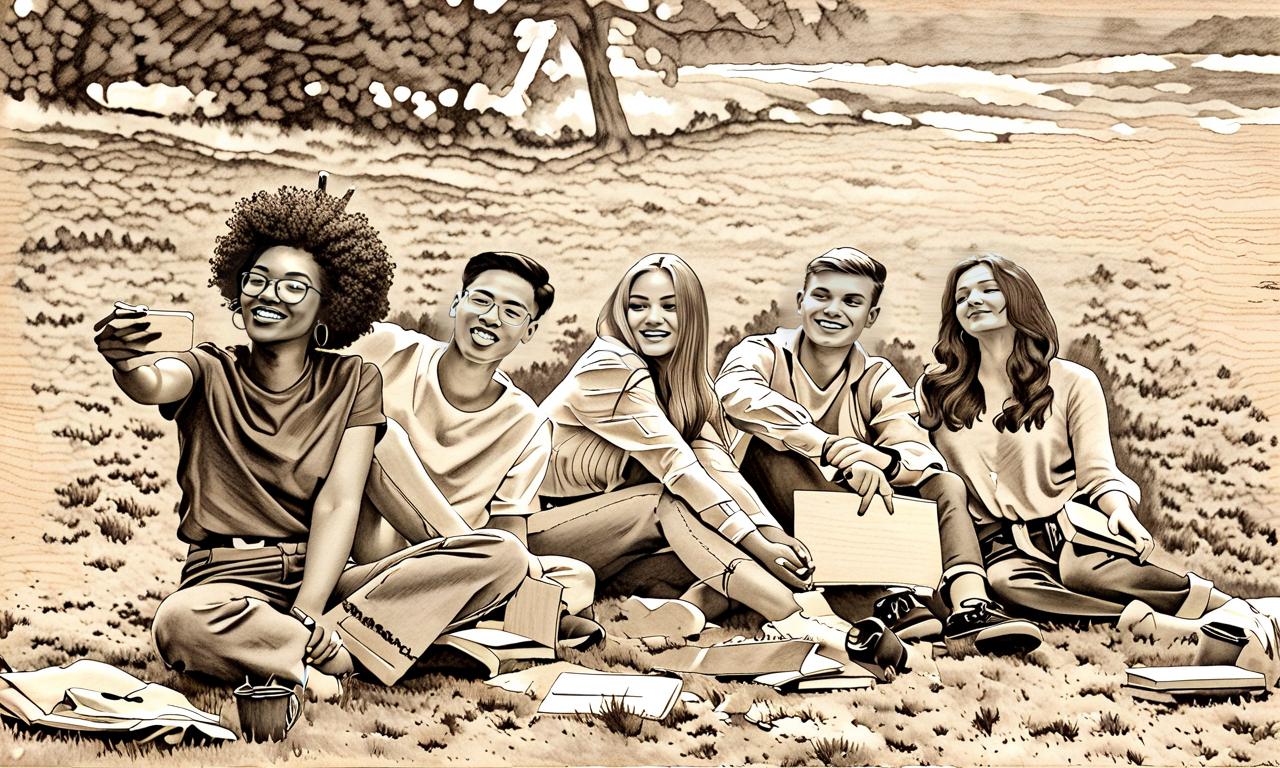} &
\myimg{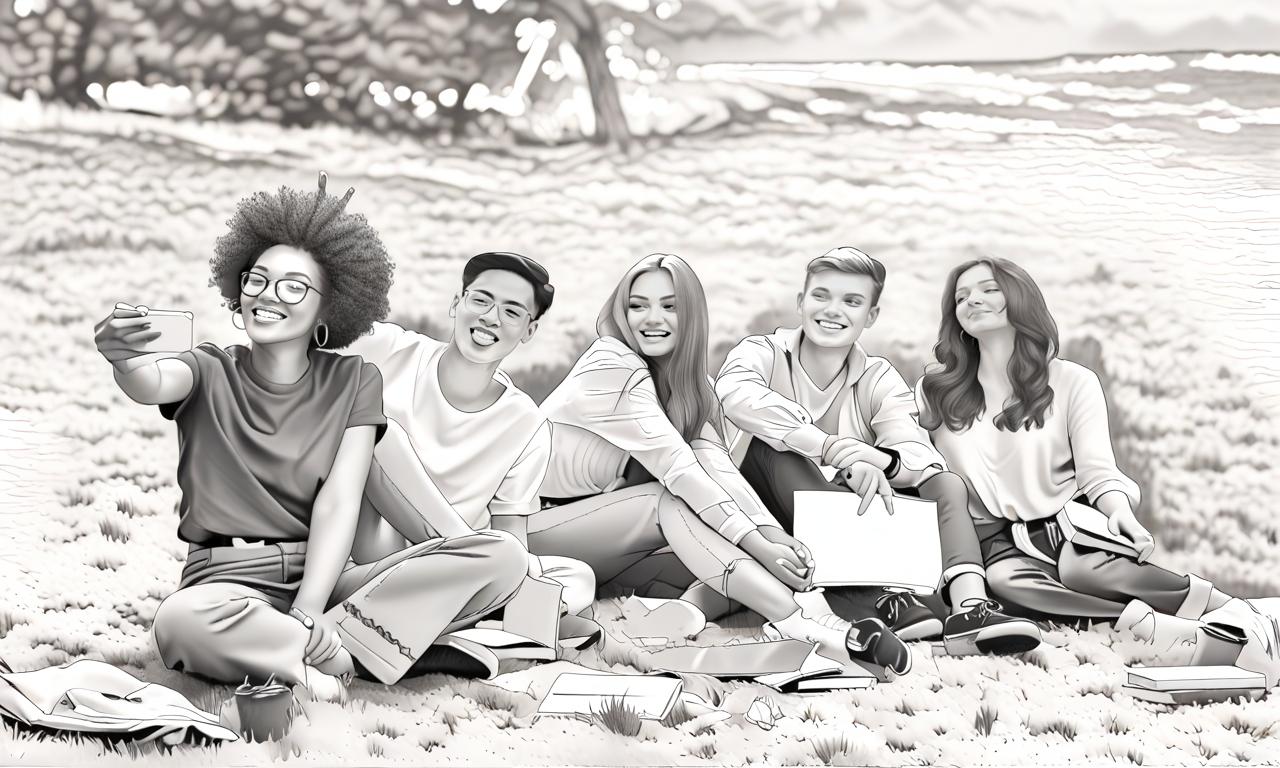} &
\myimg{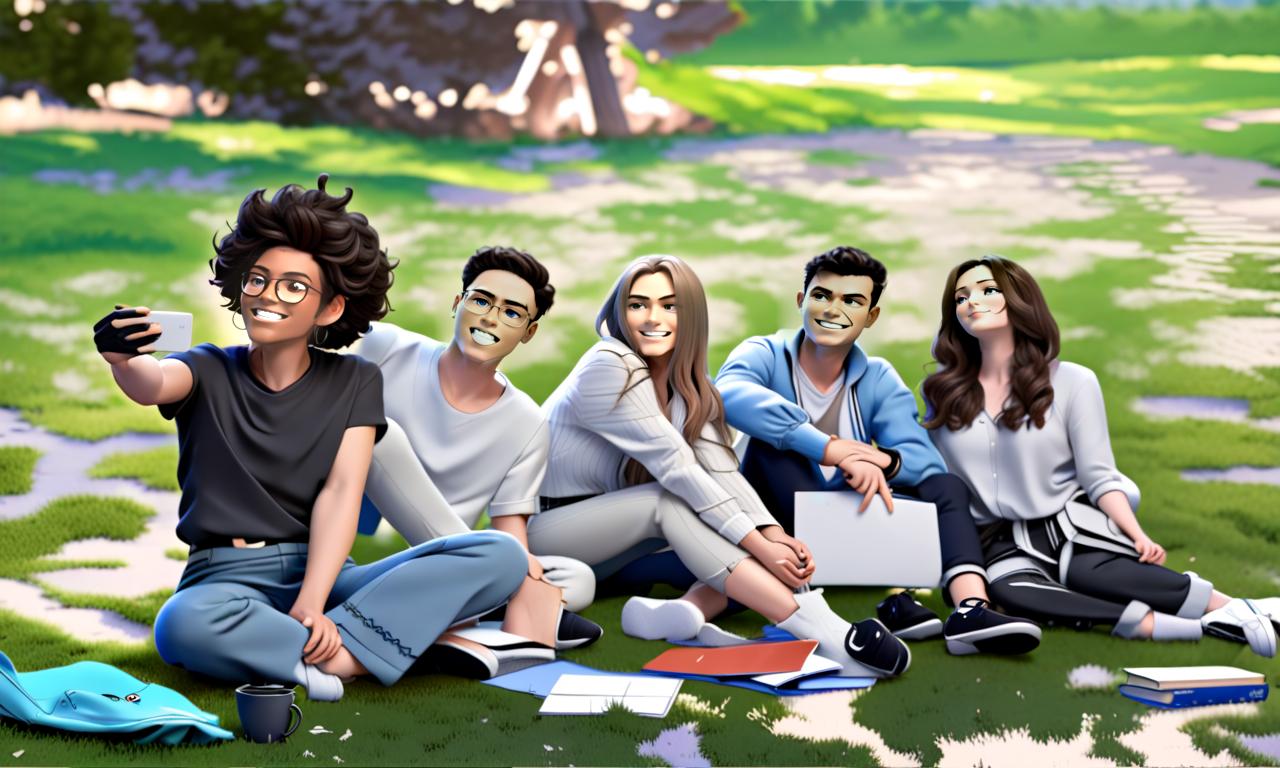} &
\myimg{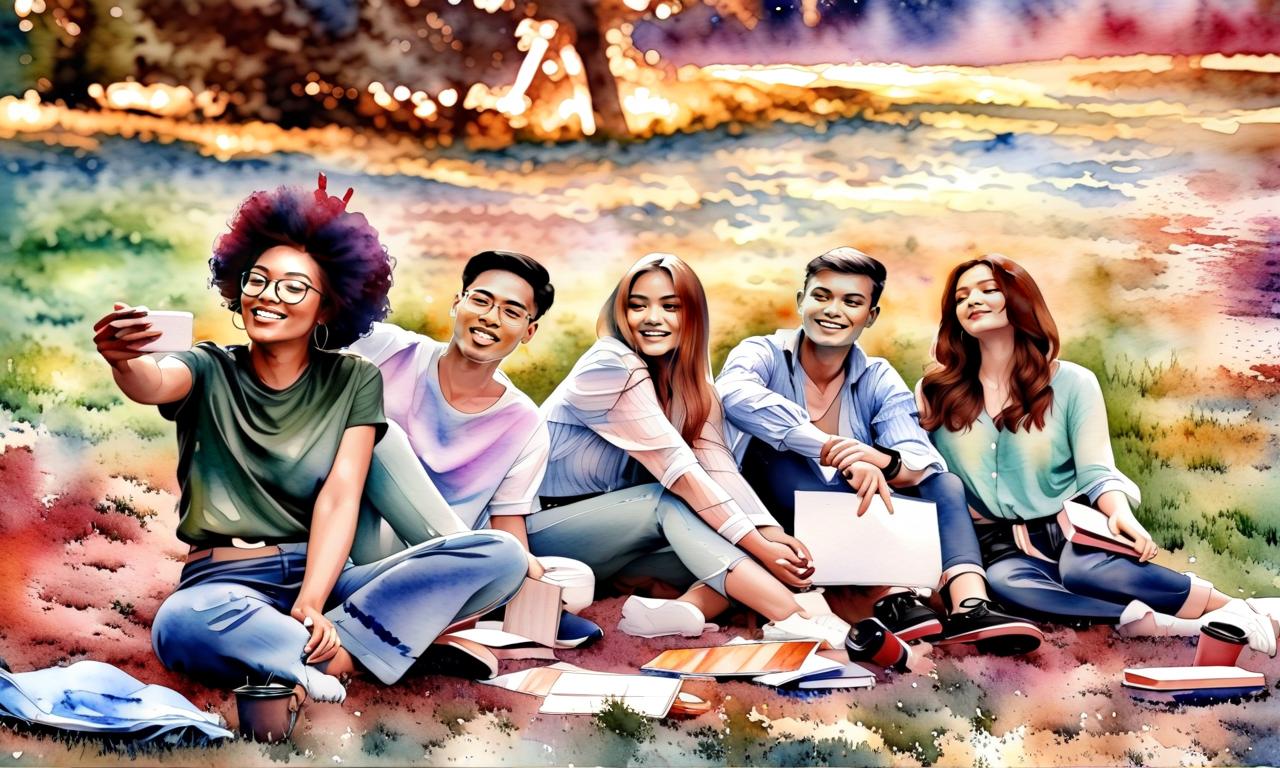} &
\myimg{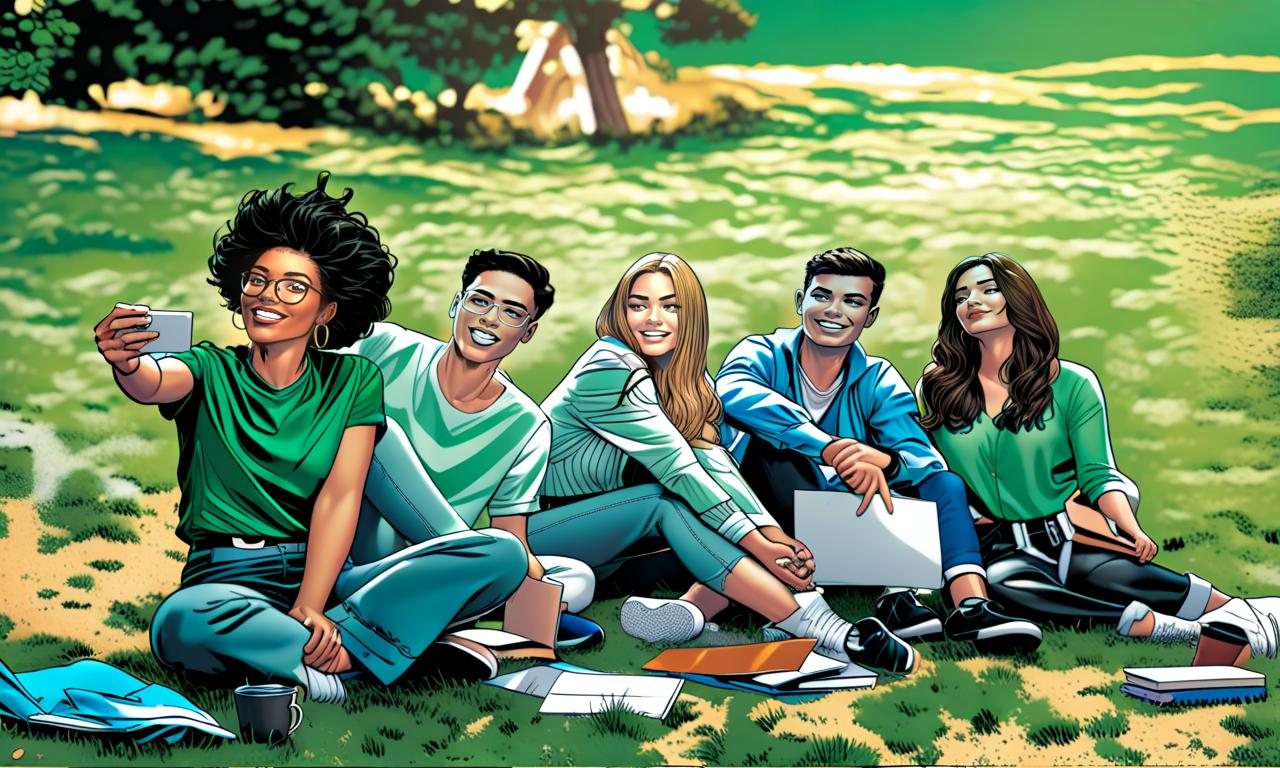} &
\myimg{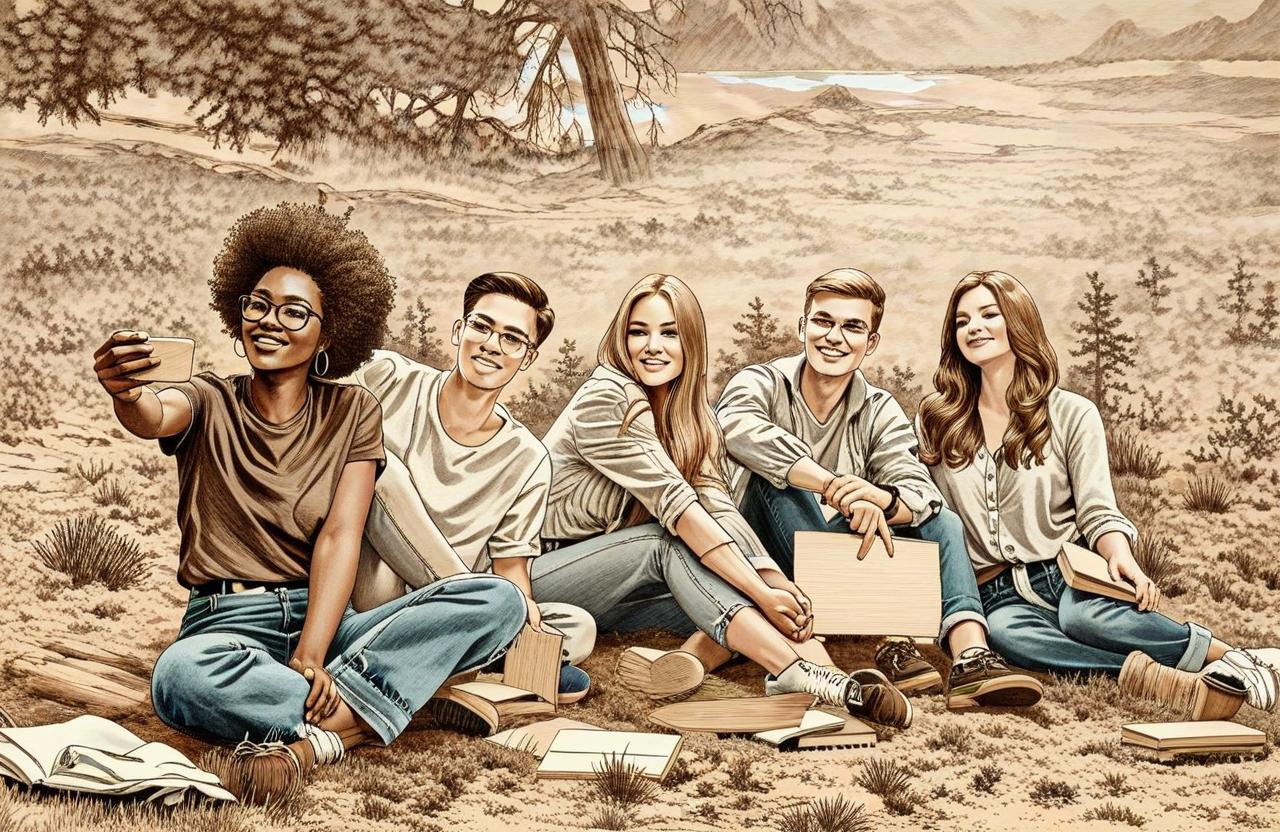} &
\myimg{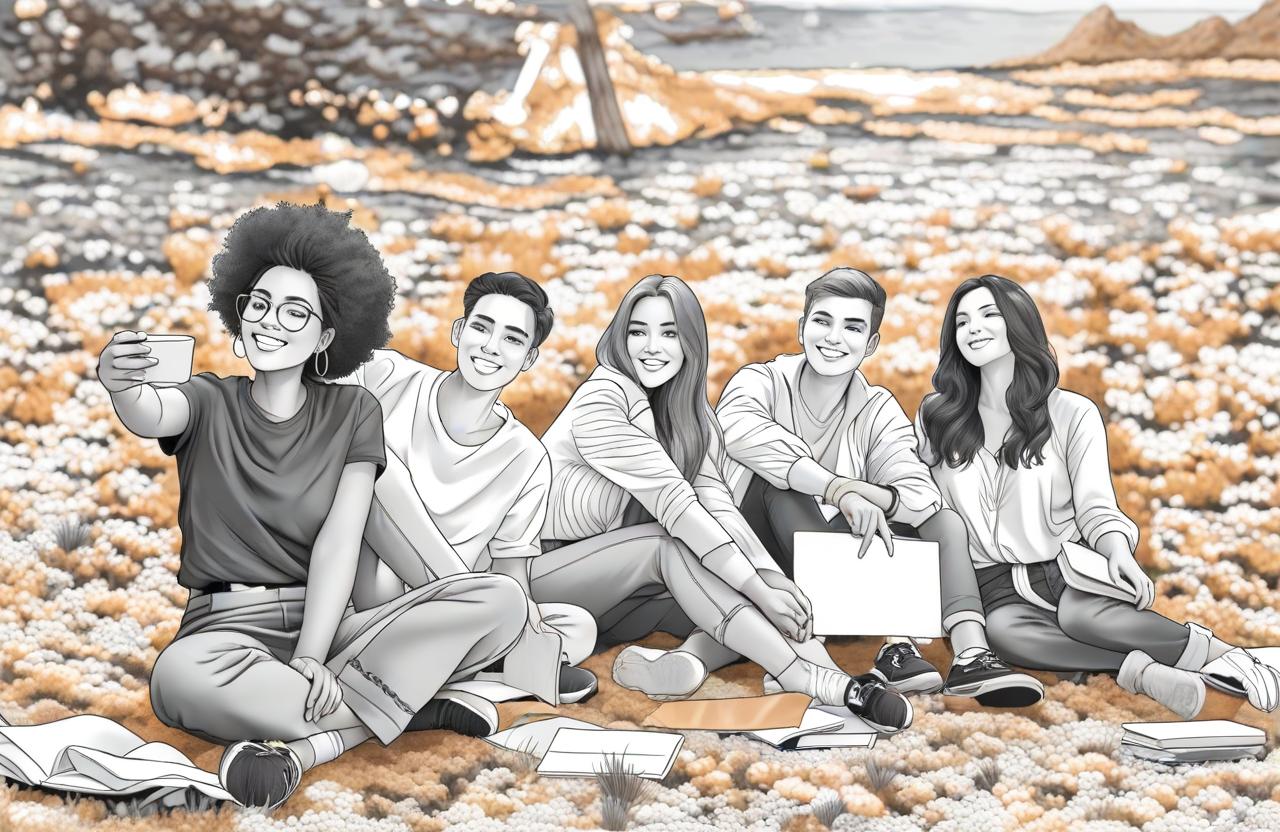} &
\myimg{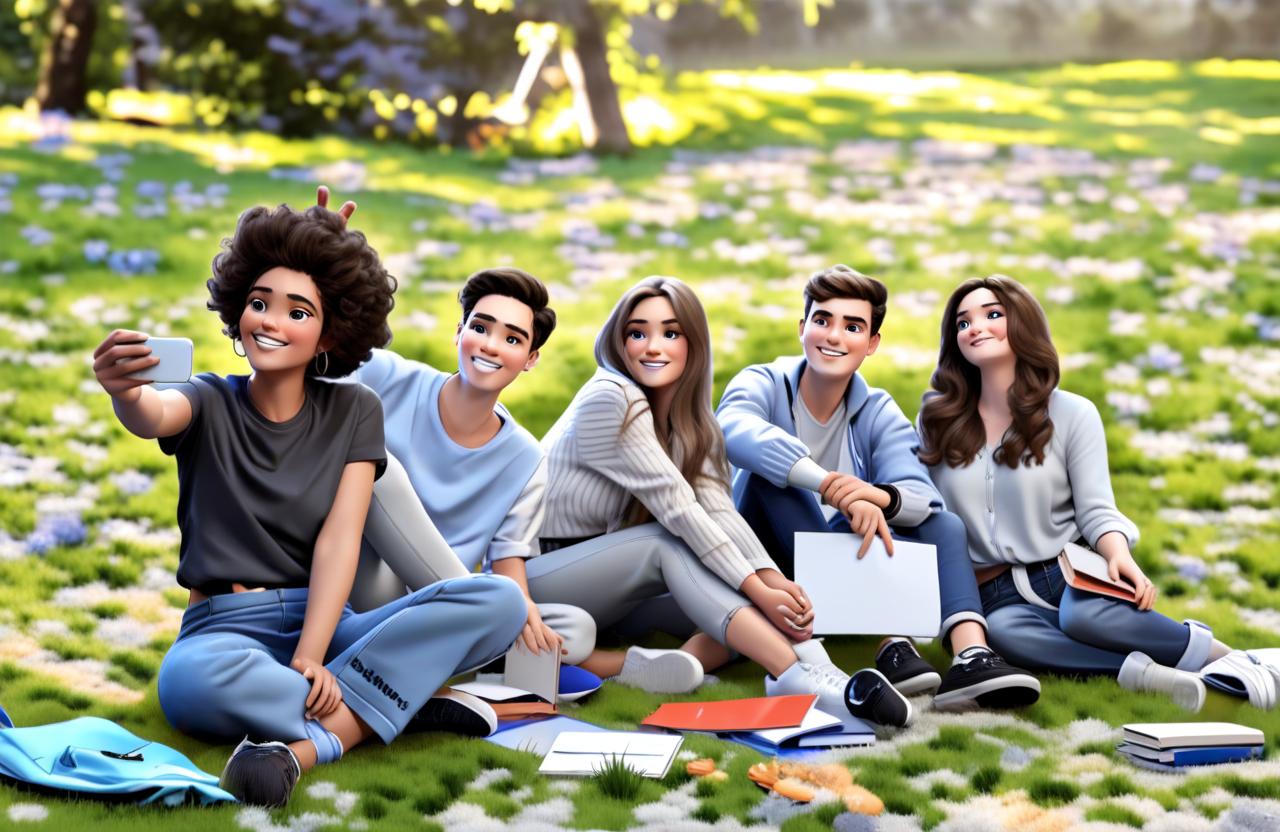} &
\myimg{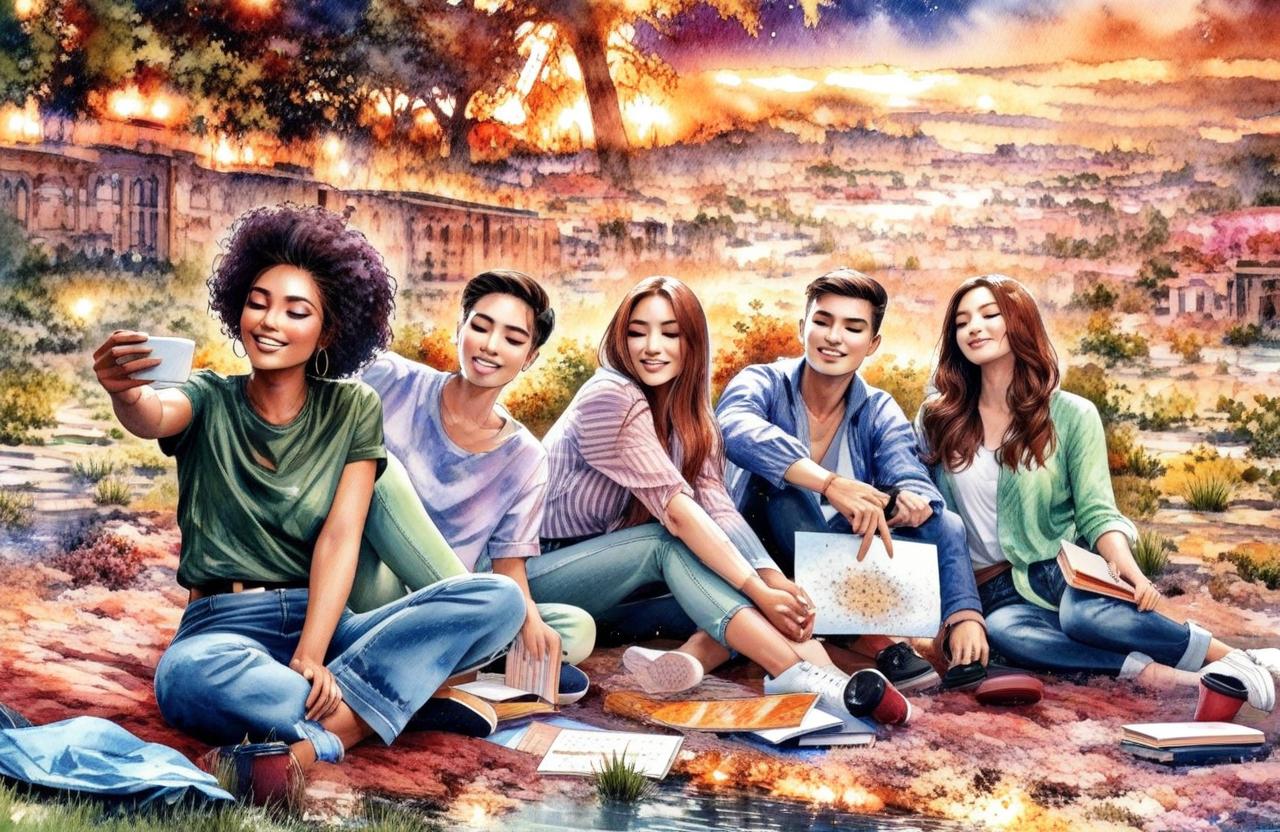} &
\myimg{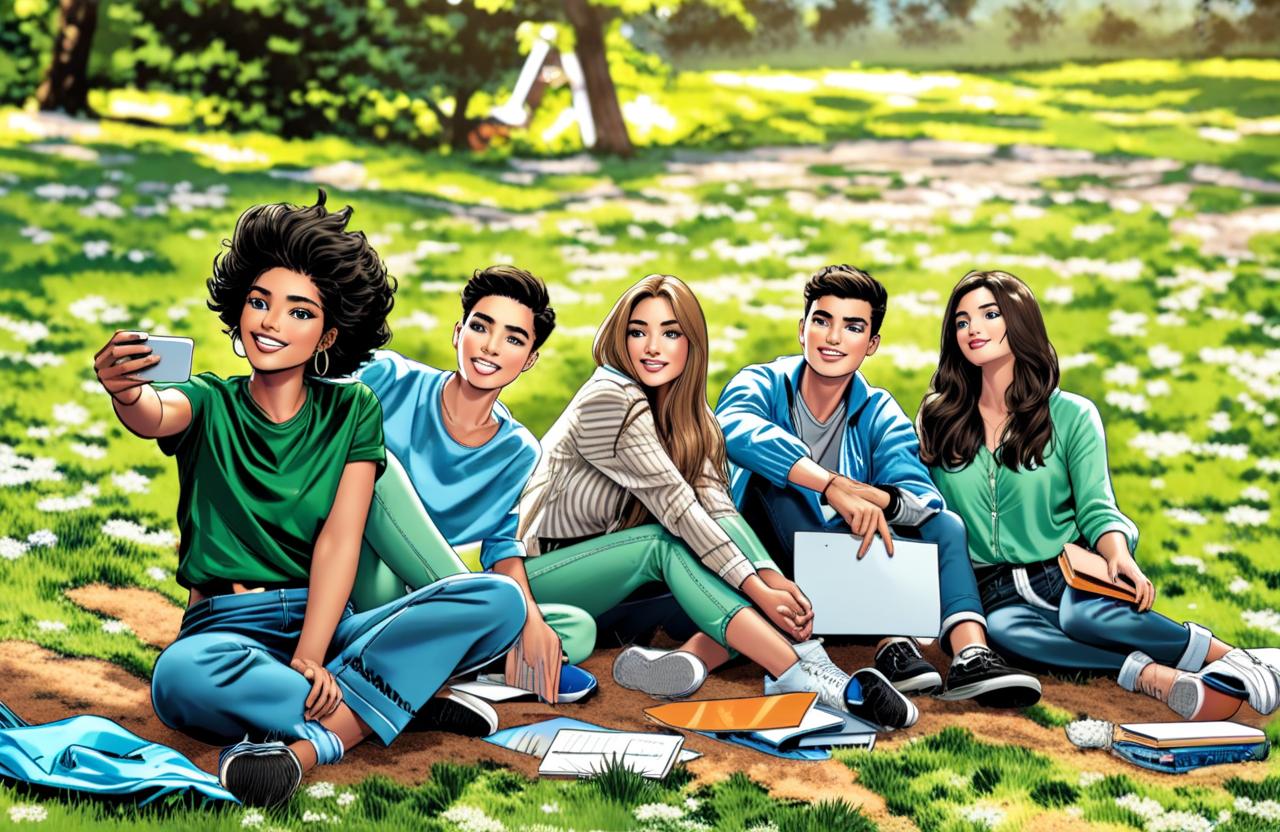} \\

\myimg{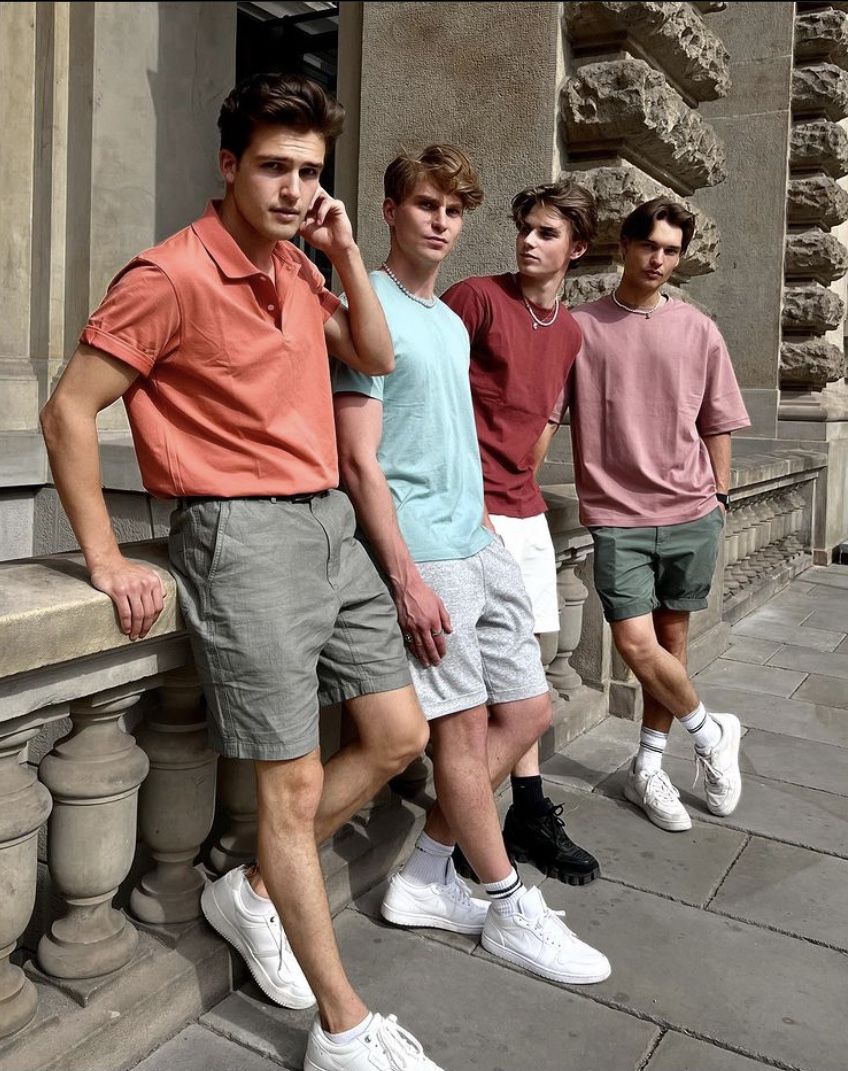} &
\myimg{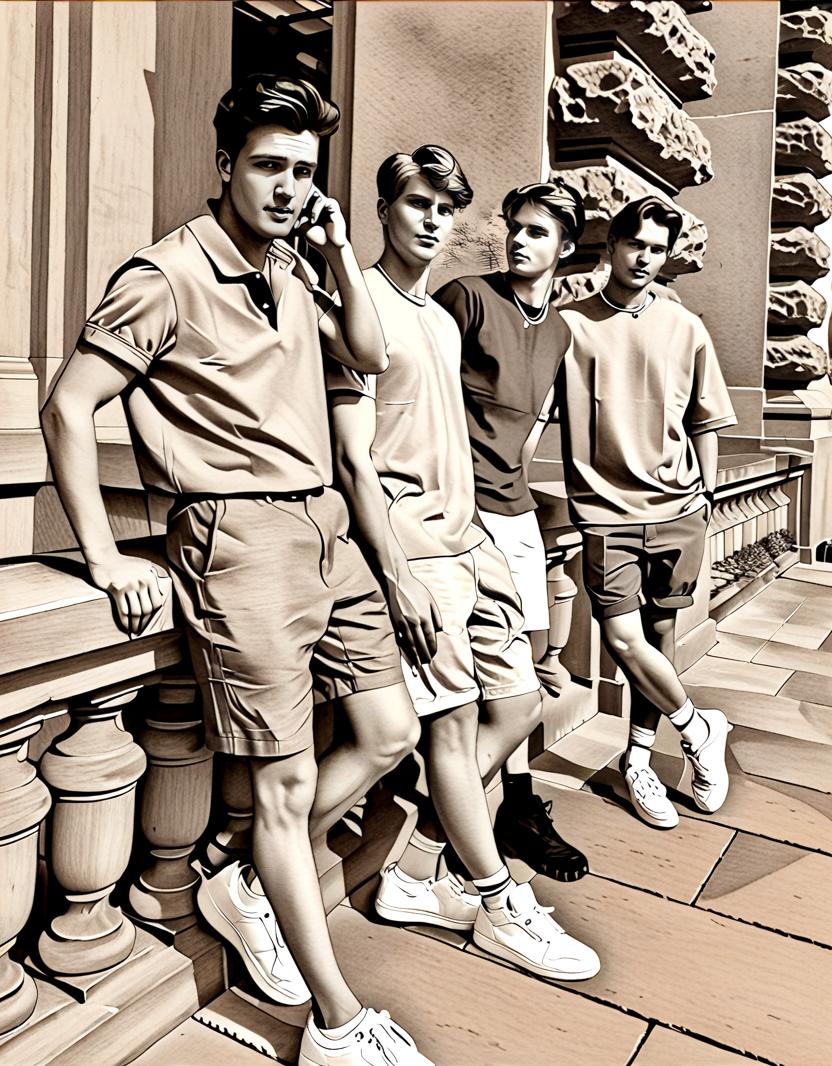} &
\myimg{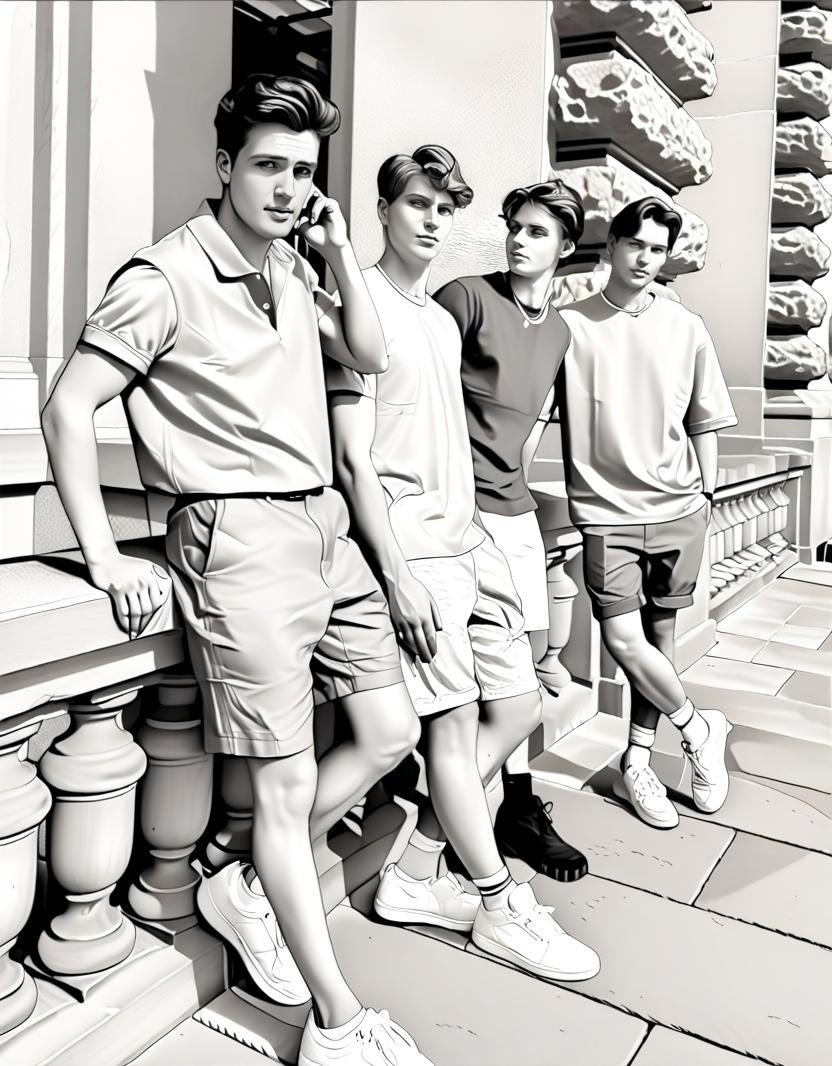} &
\myimg{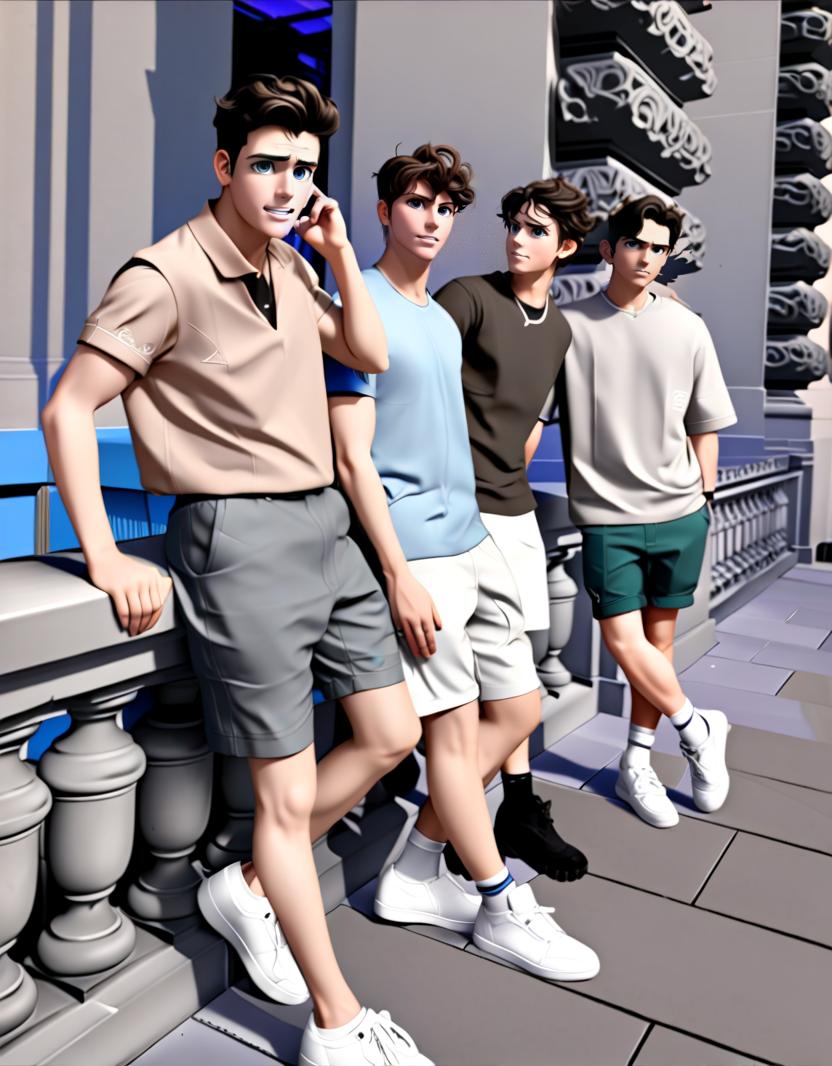} &
\myimg{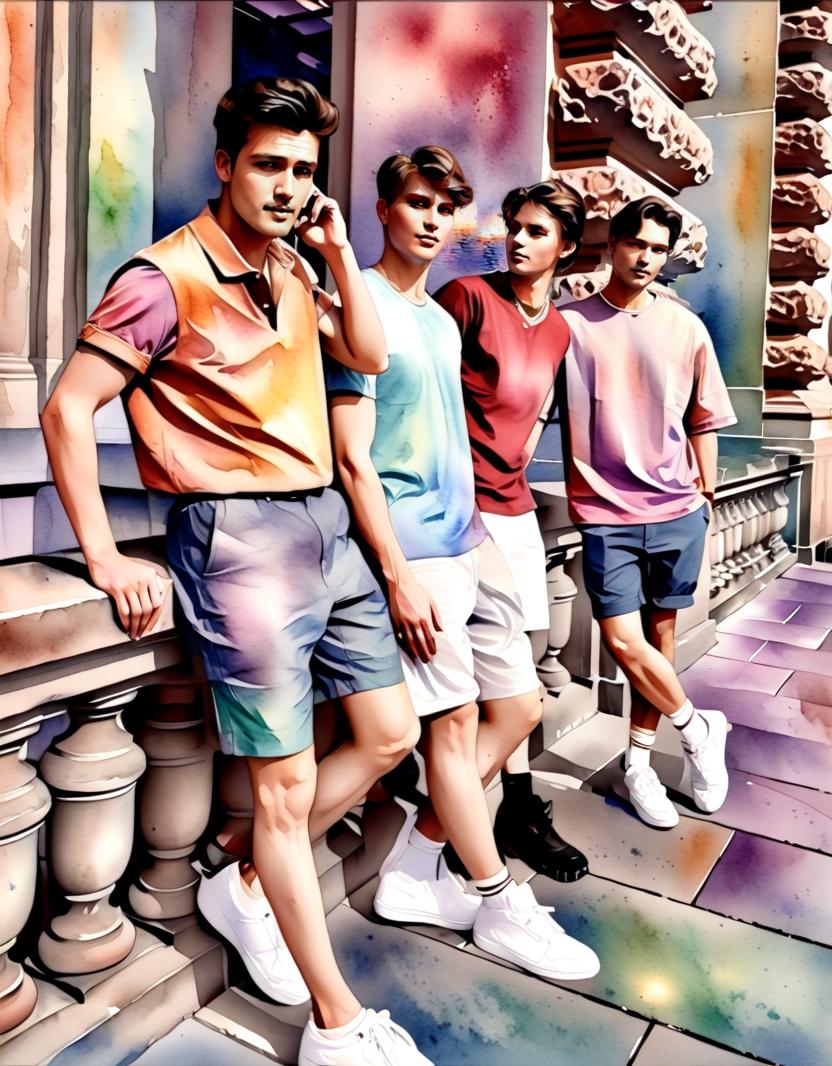} &
\myimg{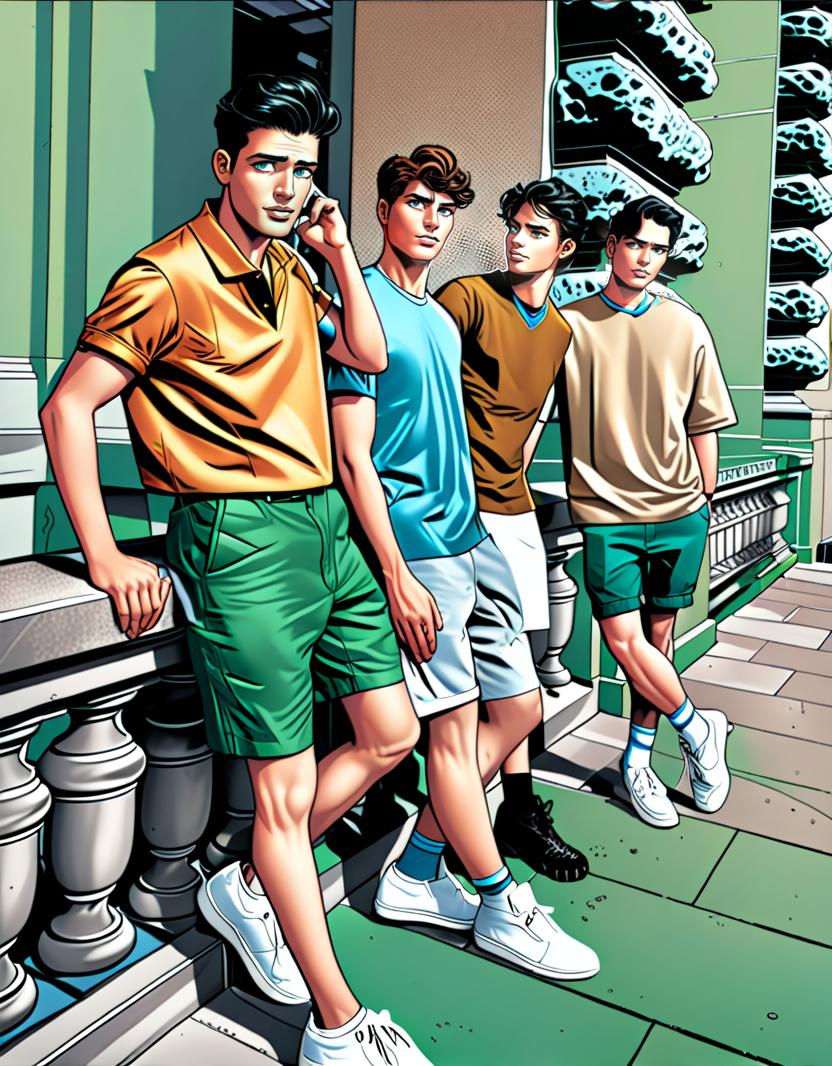} &
\myimg{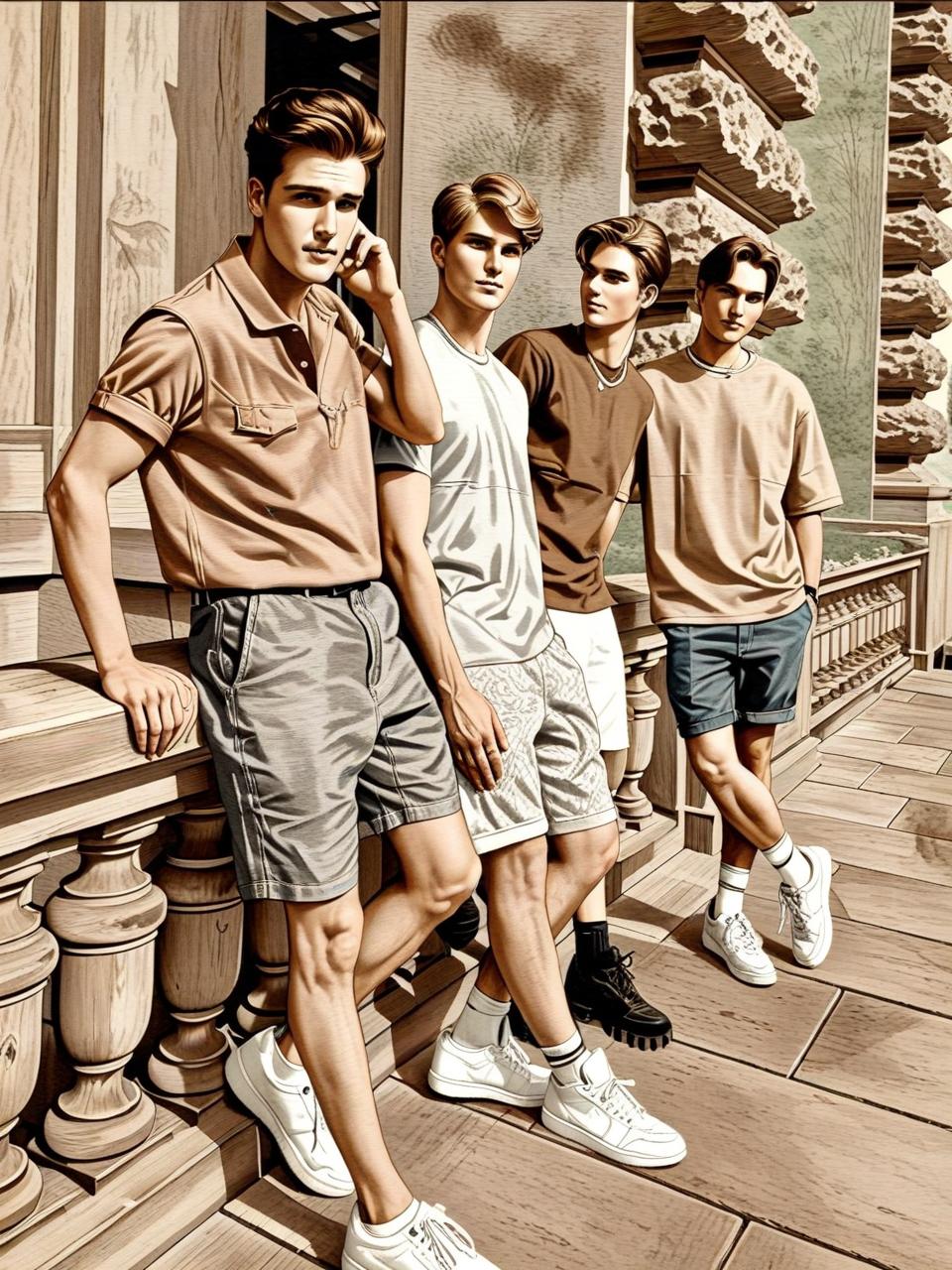} &
\myimg{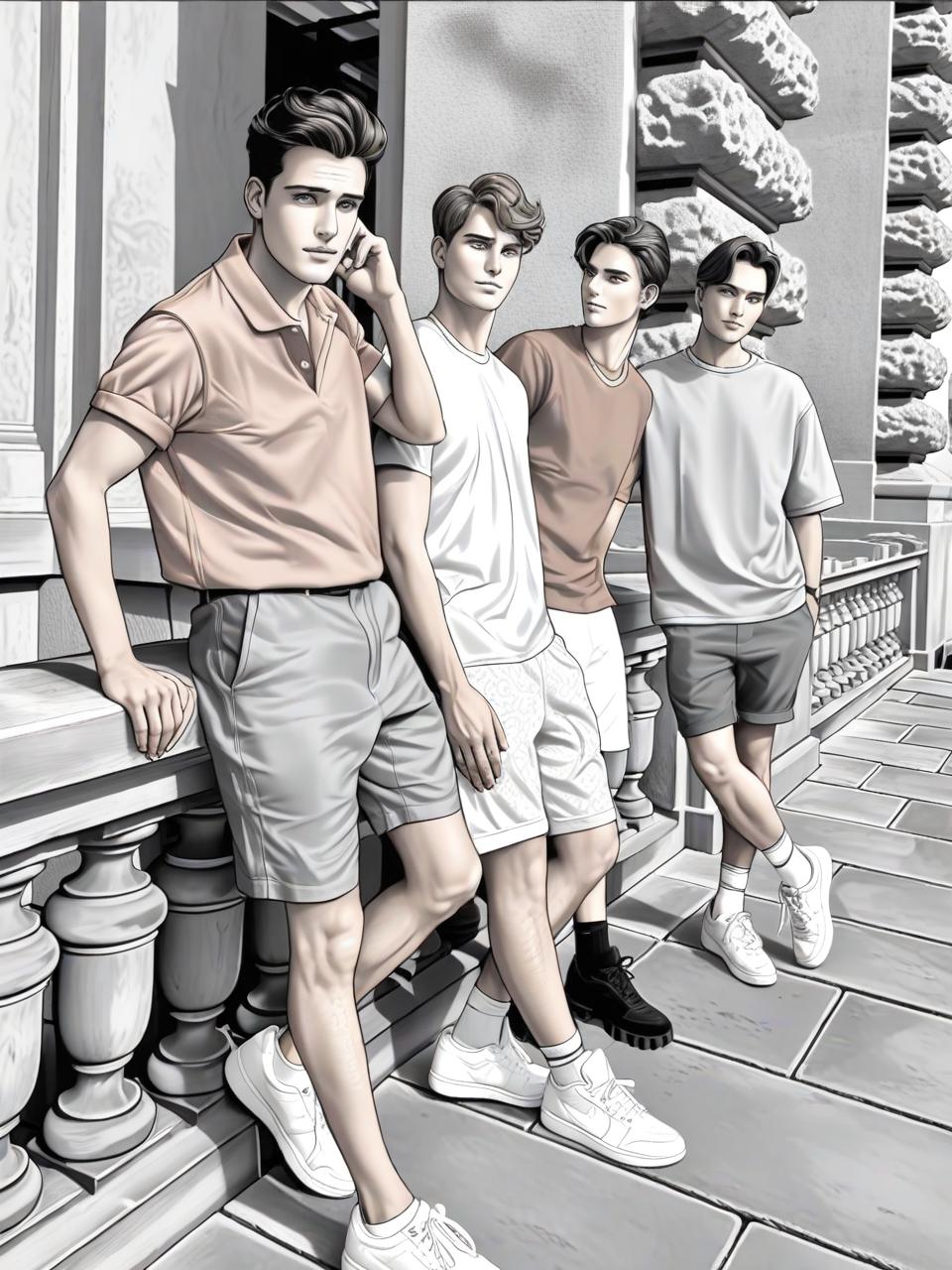} &
\myimg{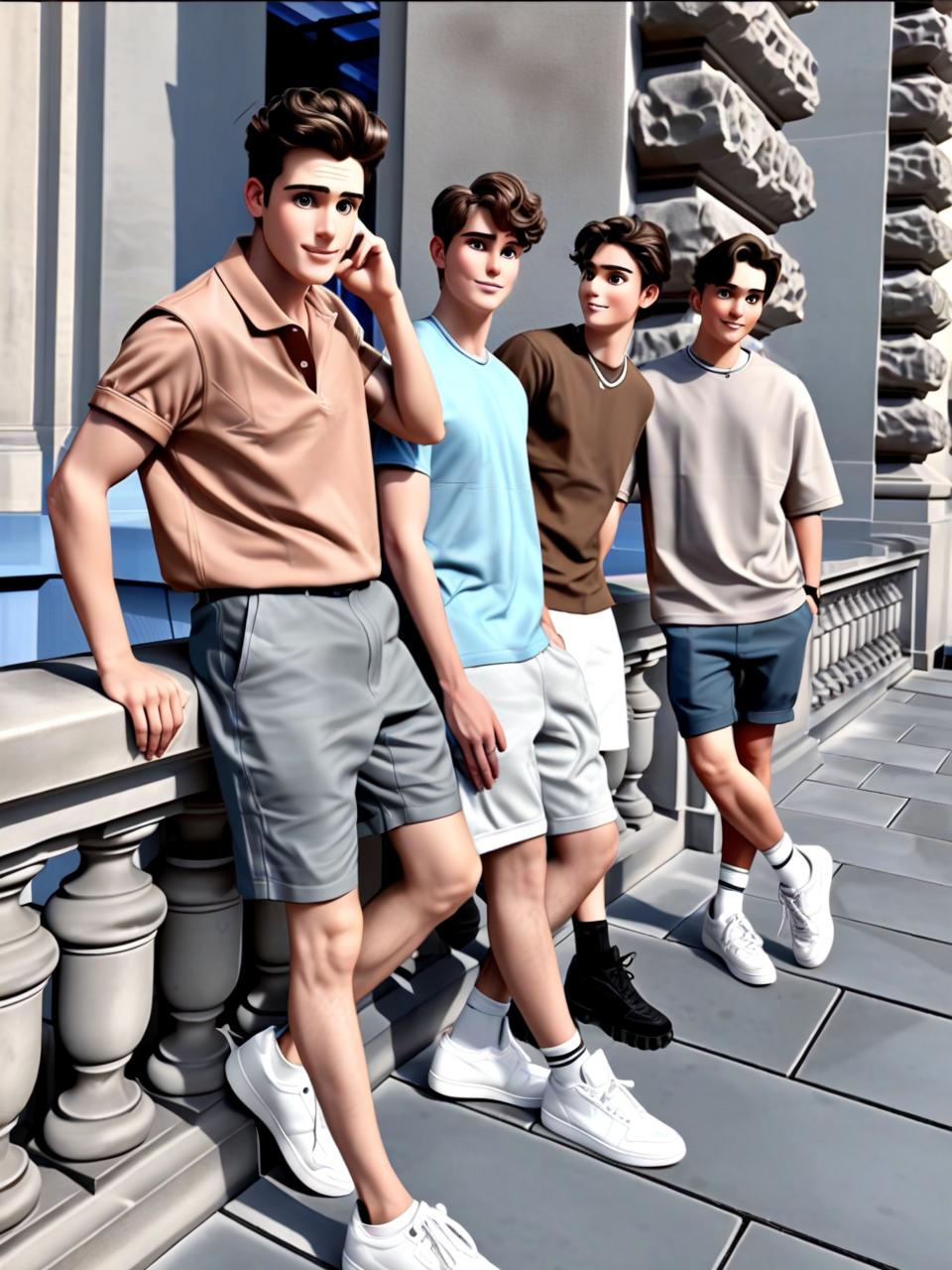} &
\myimg{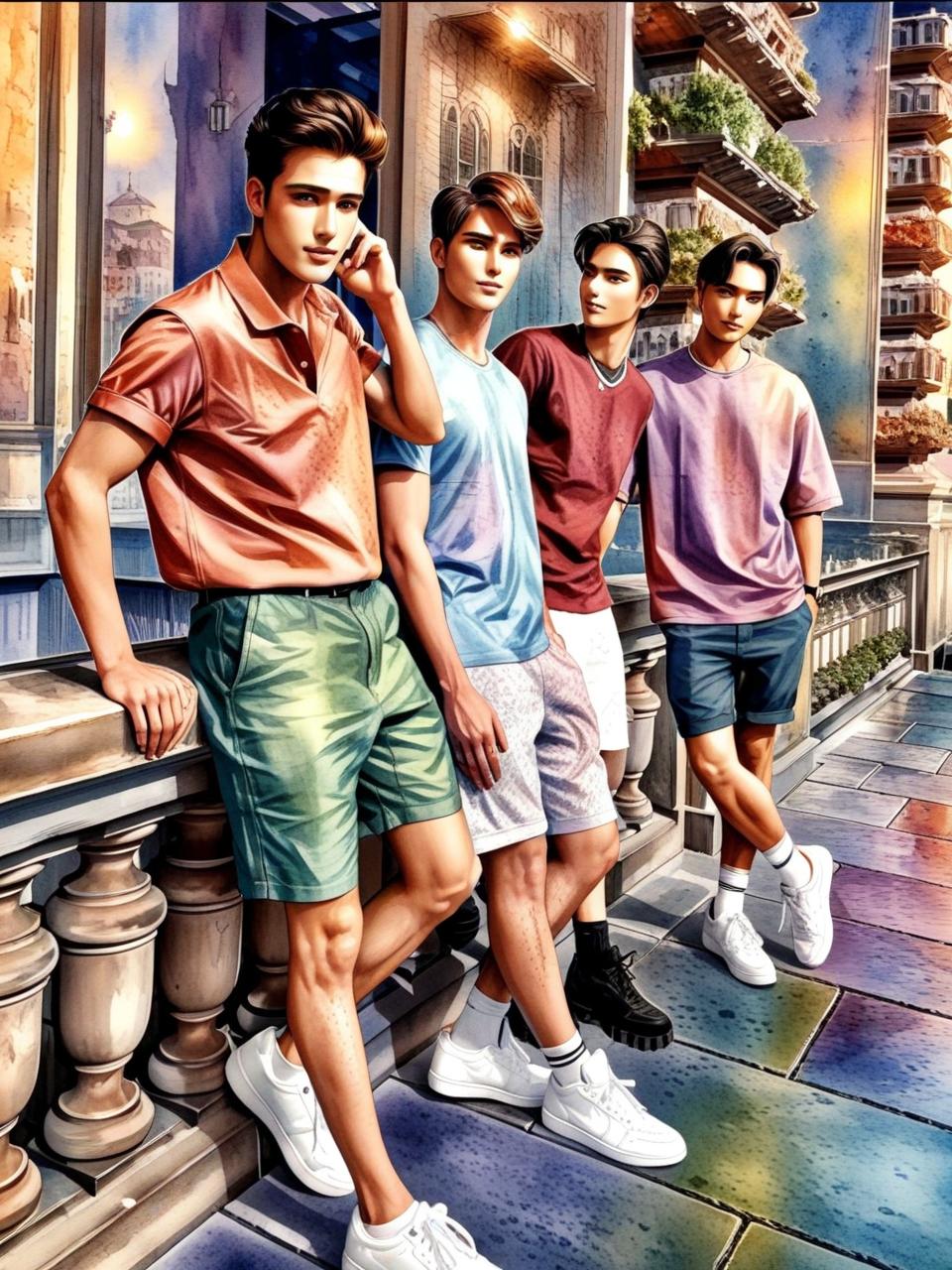} &
\myimg{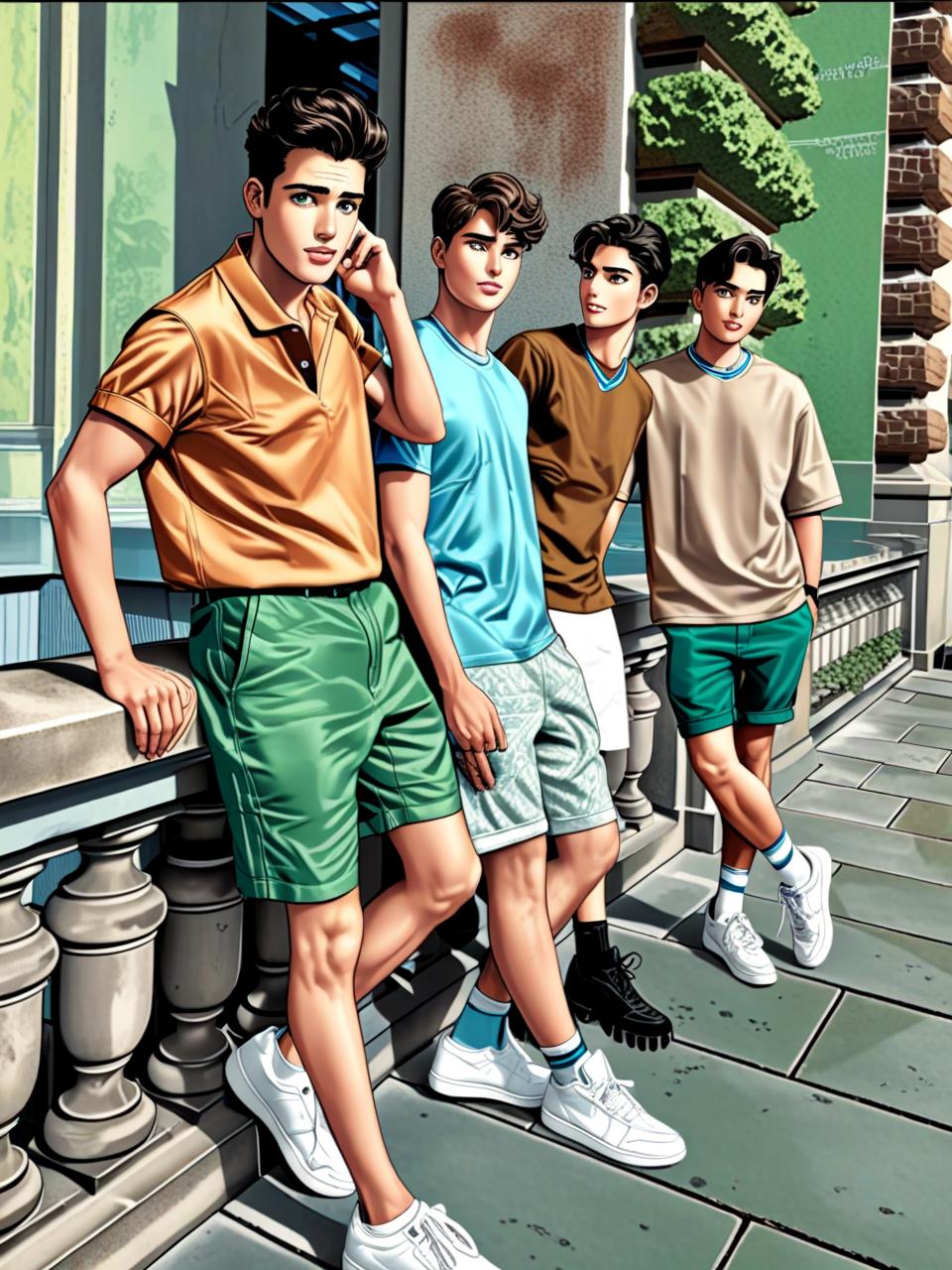} \\

\myimg{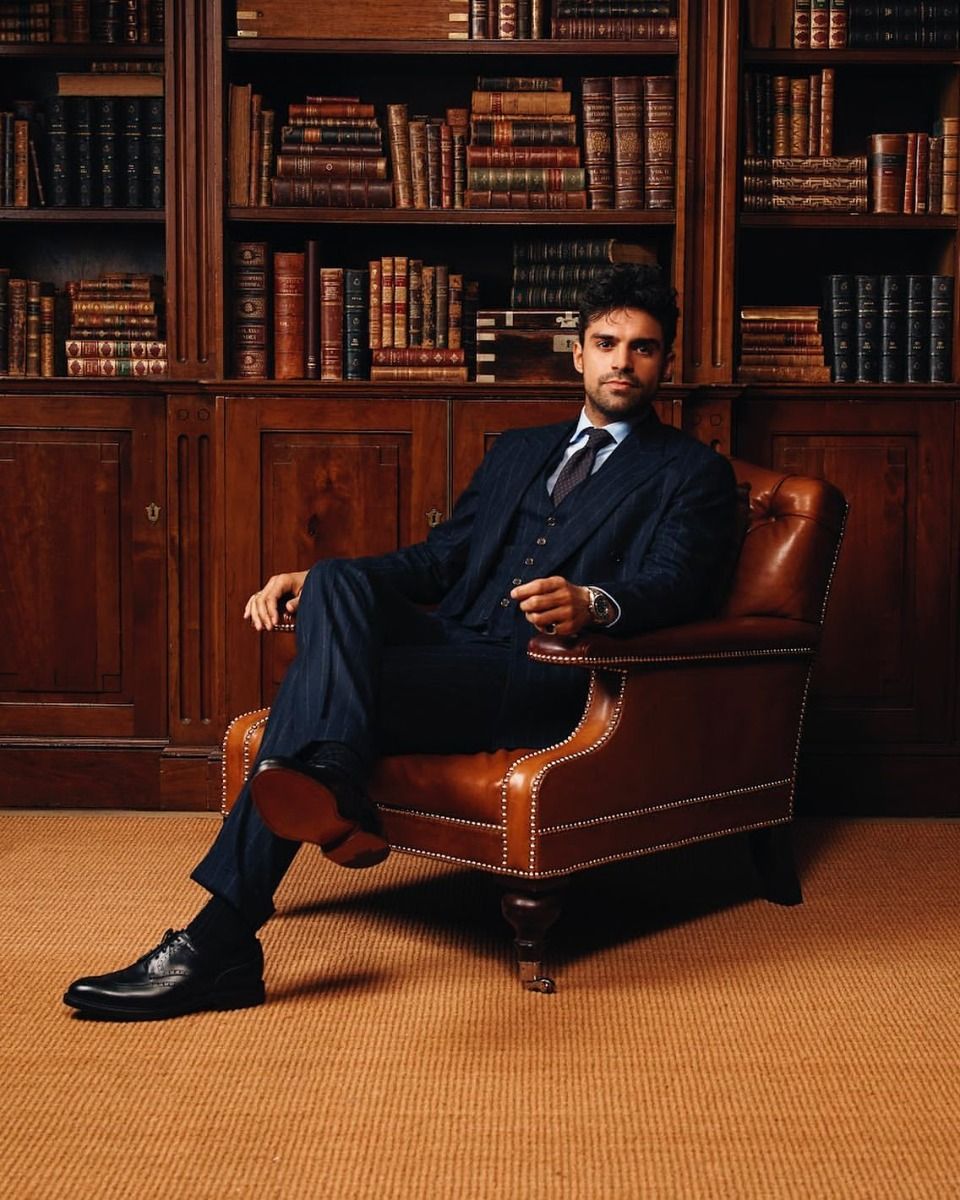} &
\myimg{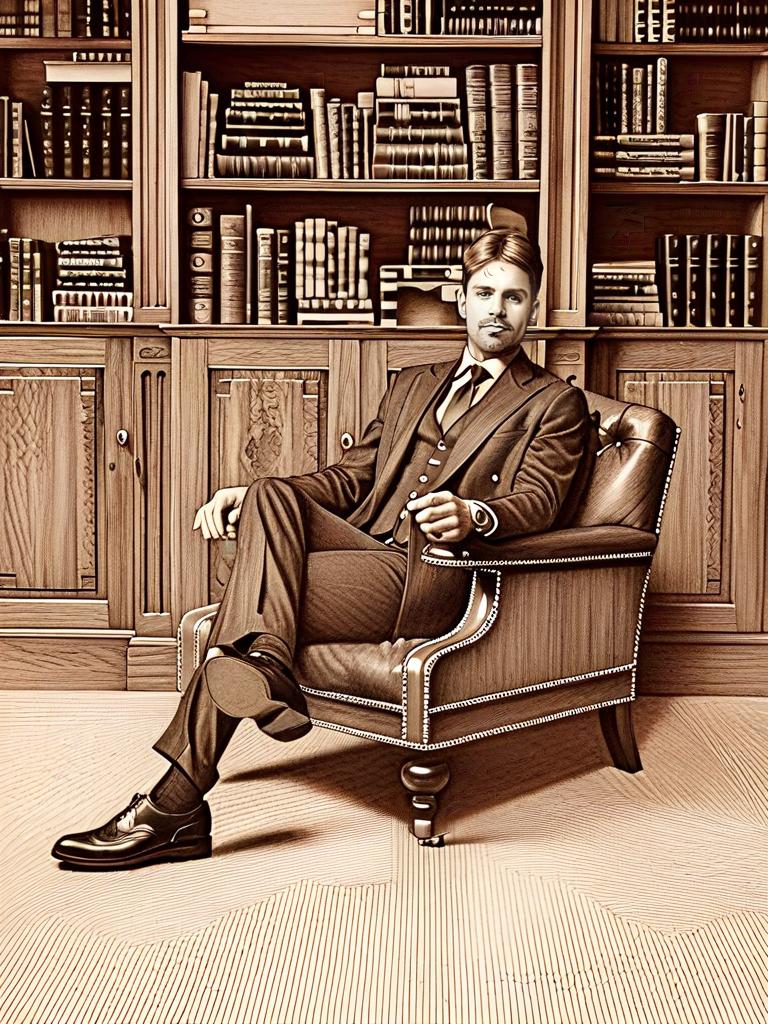} &
\myimg{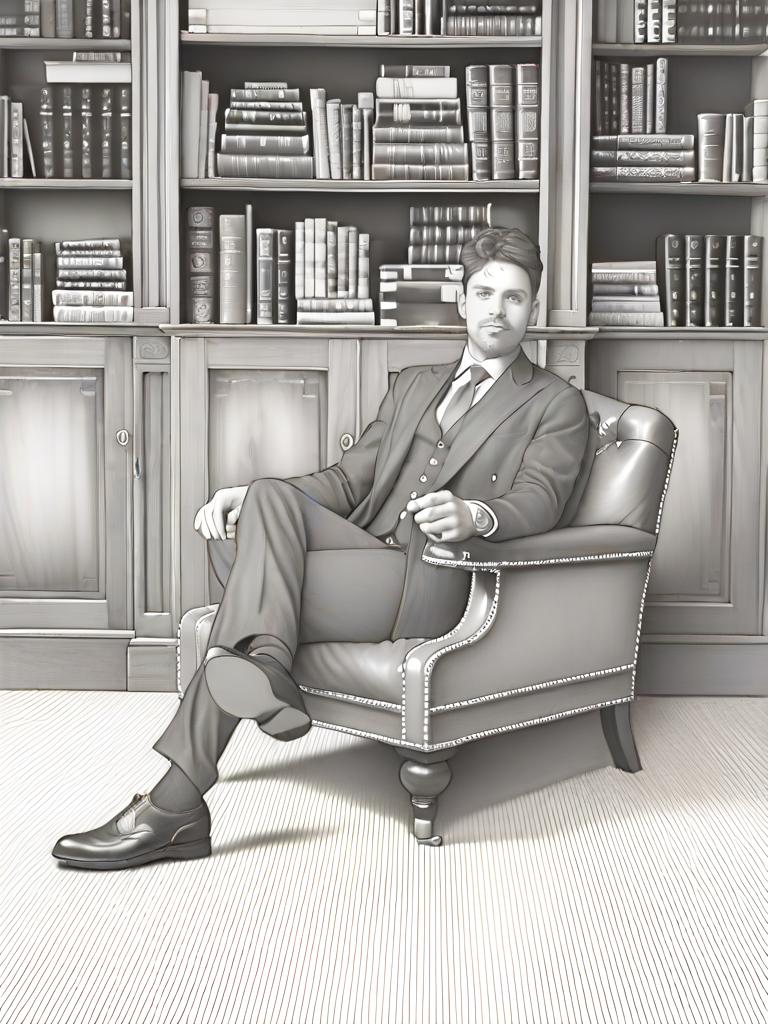} &
\myimg{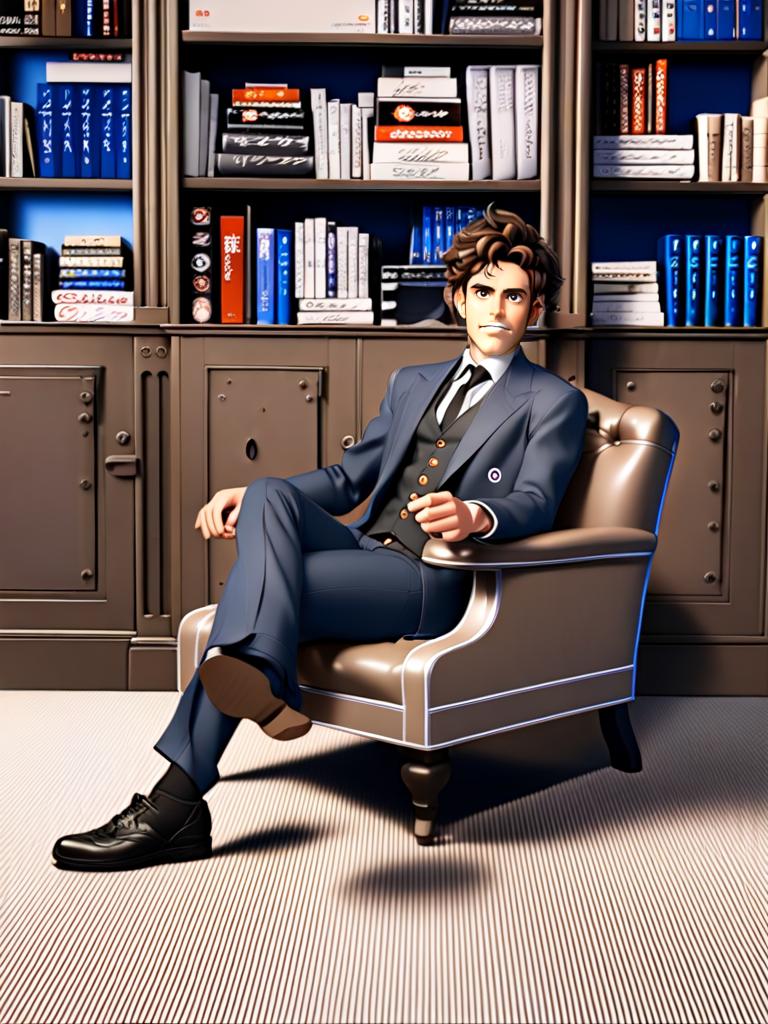} &
\myimg{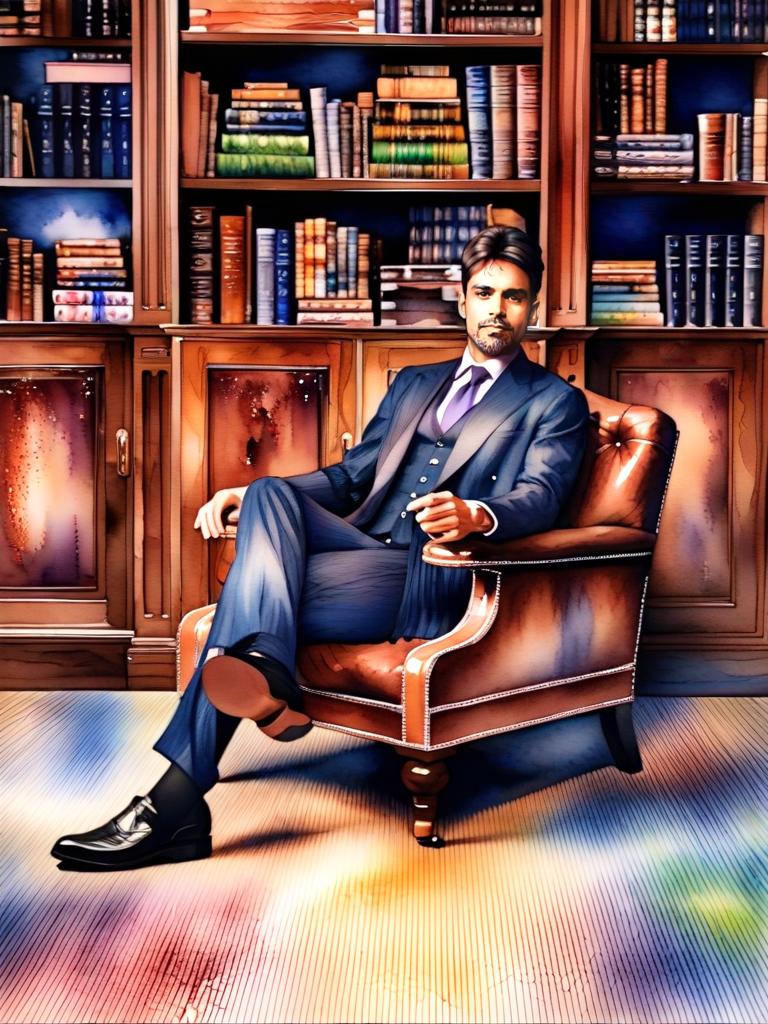} &
\myimg{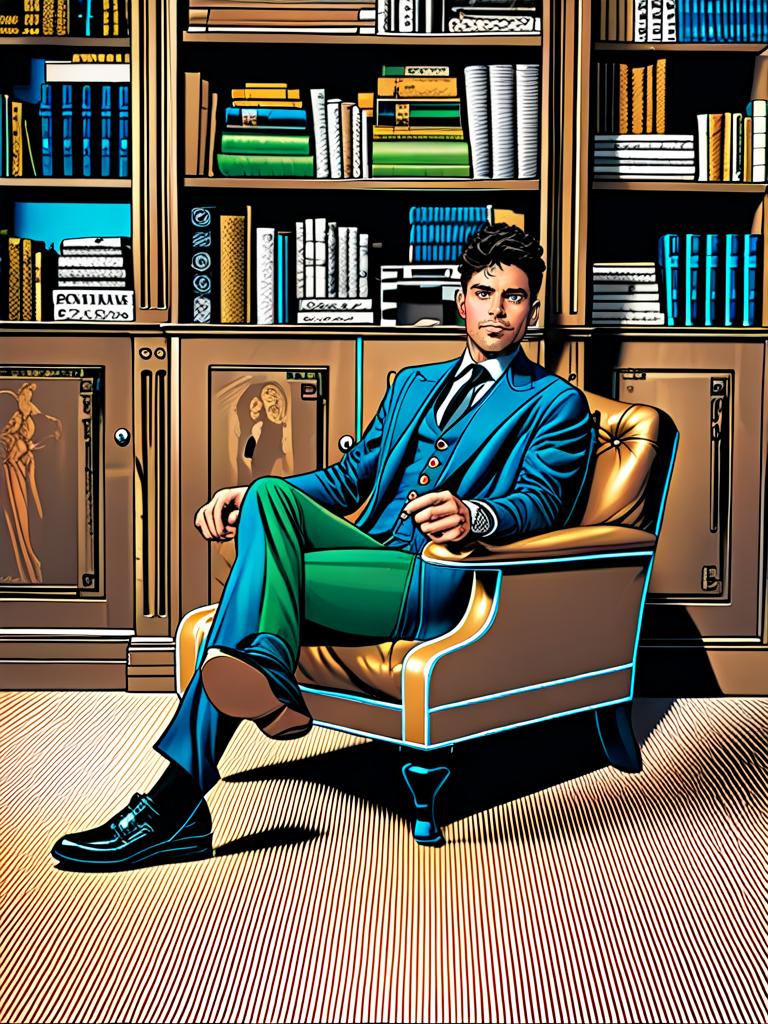} &
\myimg{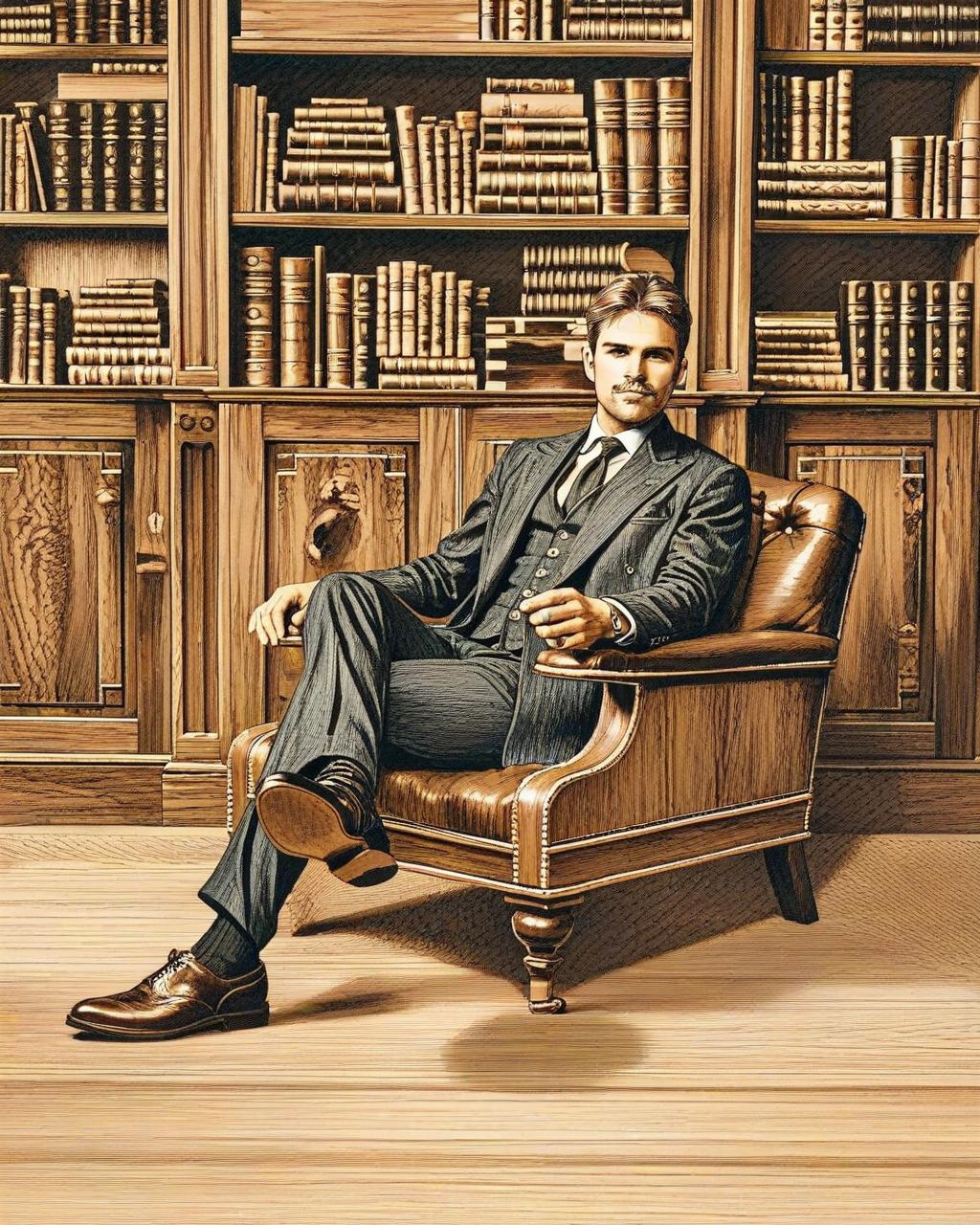} &
\myimg{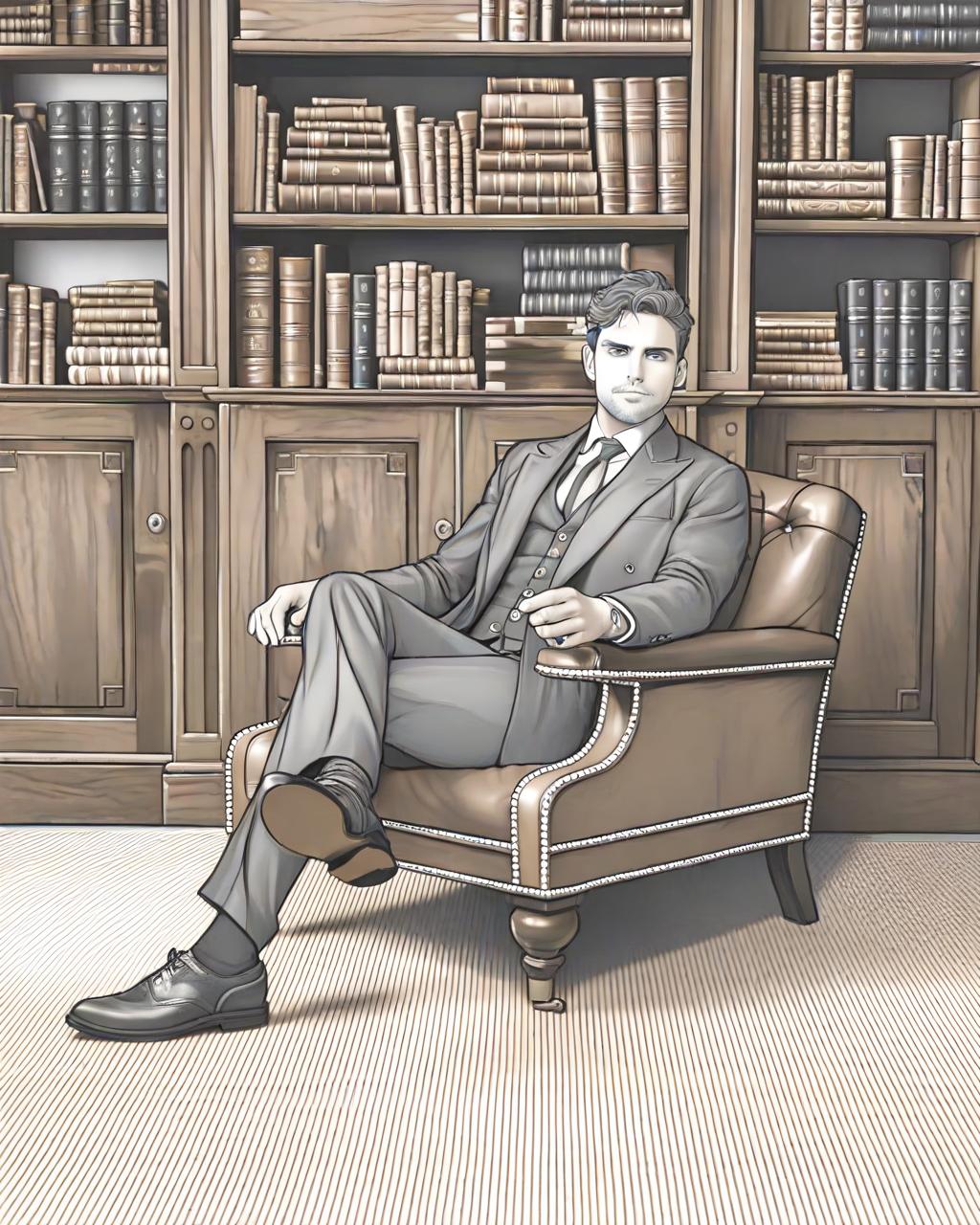} &
\myimg{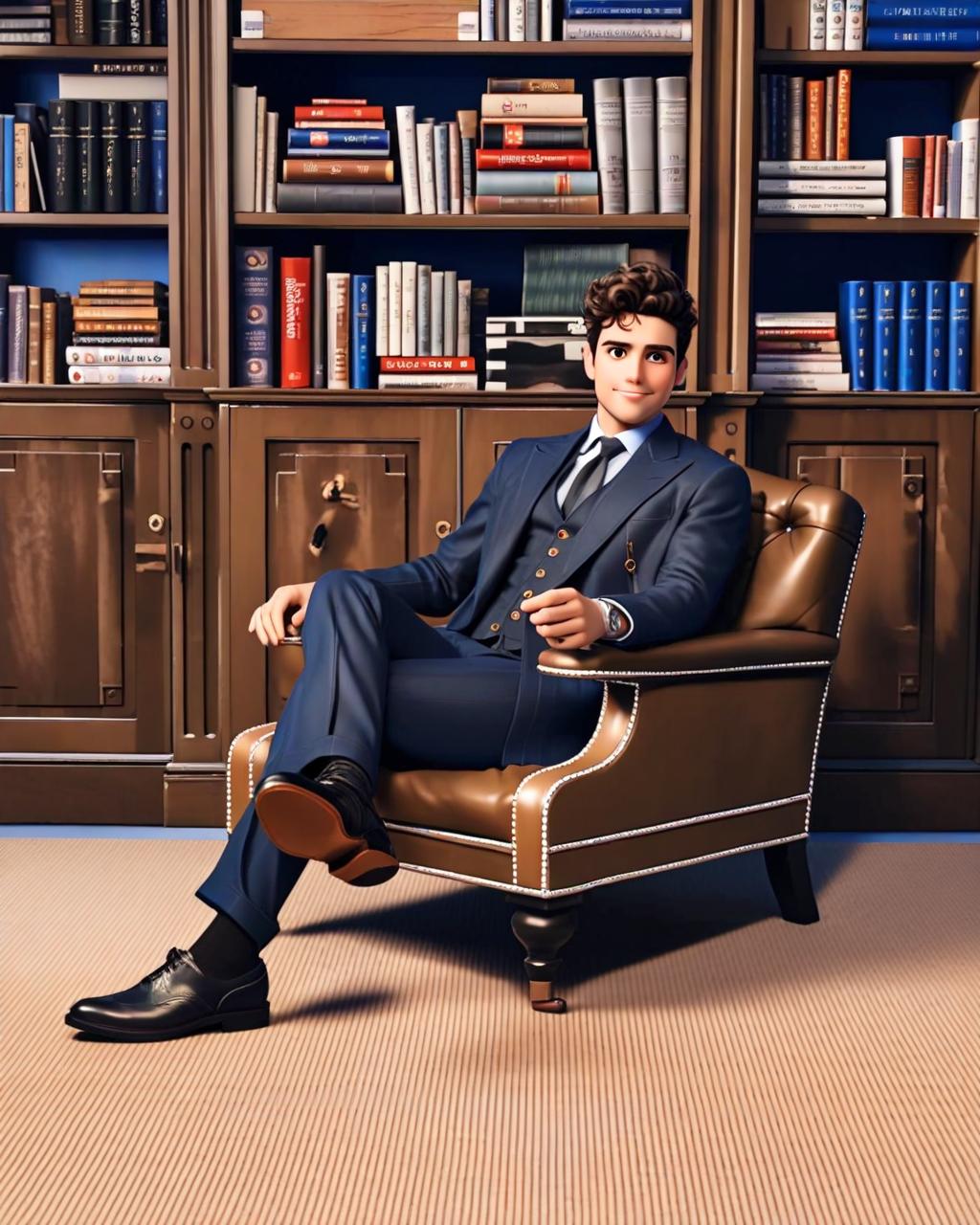} &
\myimg{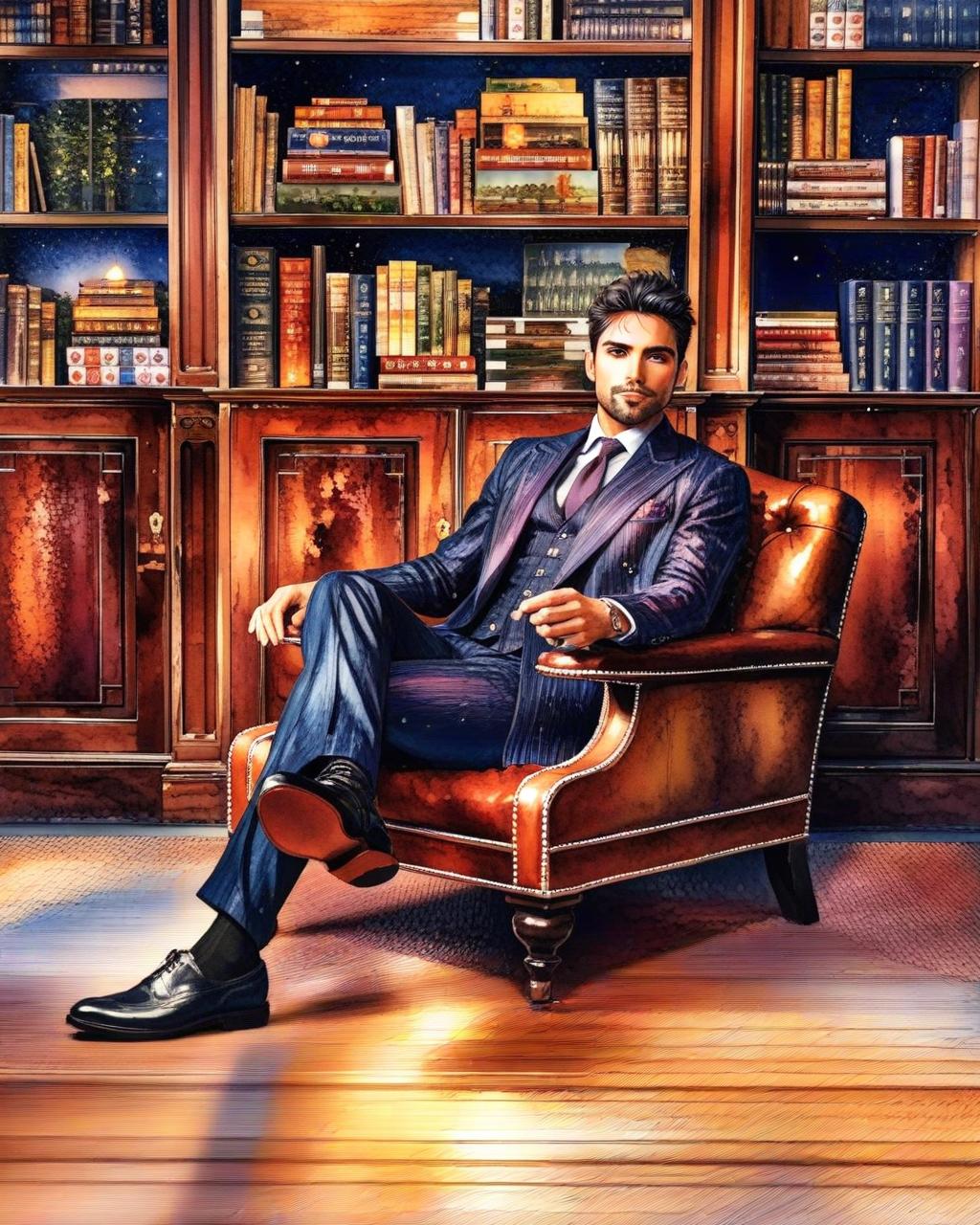} &
\myimg{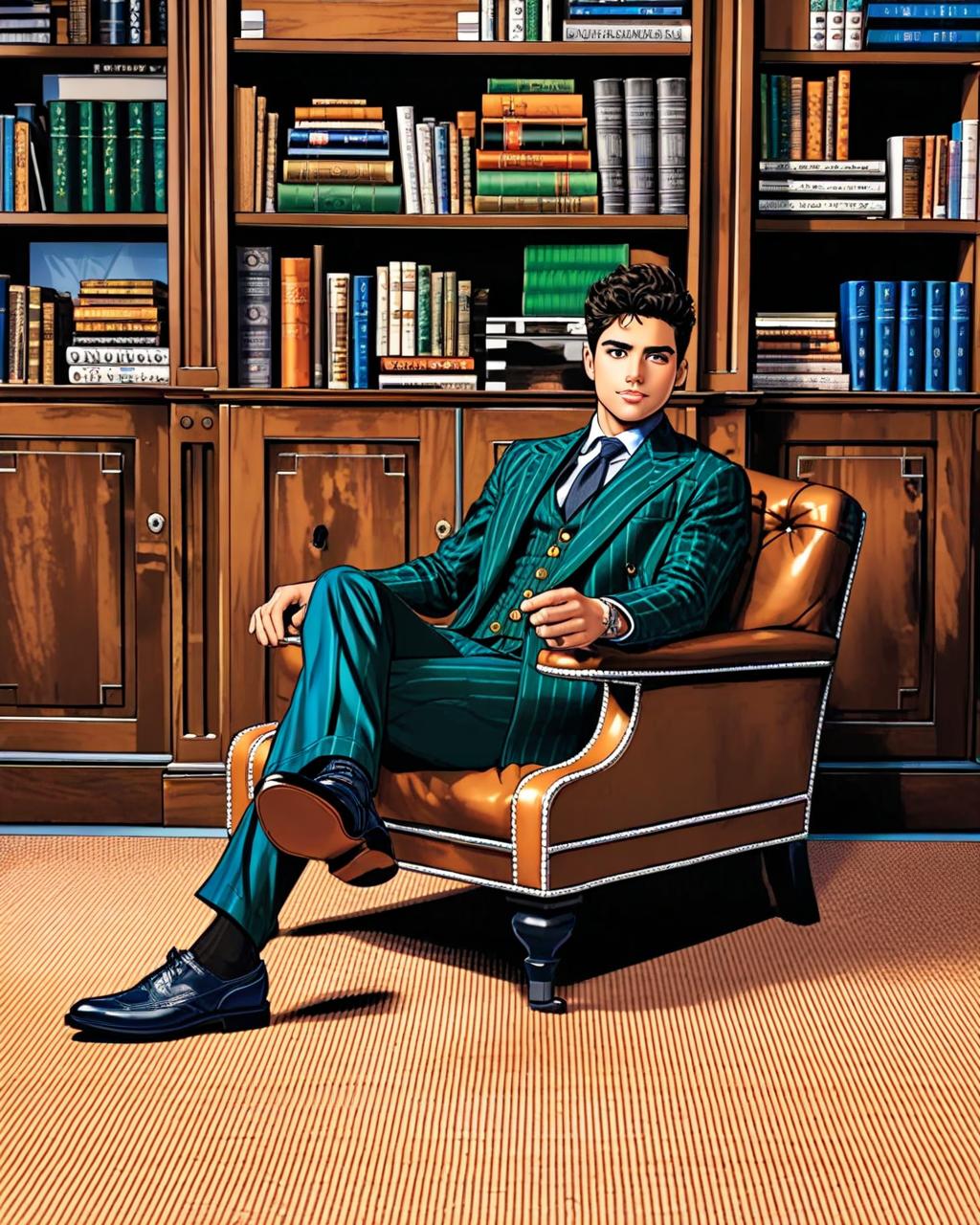} \\

\bottomrule
\end{tabular}
\caption{This table showcases the outcomes of the
InstantStyle-Plus pipeline\cite{wang2024instantstyleplus} with and without our proposed
Mosaic Restored Content Image technique. Our approach
demonstrates notable improvements in preserving individual identities, particularly in images with multiple faces or
small facial regions, while maintaining a high level of stylization quality.}
\label{tab:comparison}
\end{table*}

\subsection{Effect of Mosaic Restored Content Image on Identity Preservation}

The initial prompt engineering strategy, while effective for stylization, demonstrated limitations in preserving facial identity, particularly when targeting strong artistic styles. For instance, employing style-specific keywords (e.g., "comic art," "graphic novel art," "vibrant colors", "highly detailed") for comic-style generation yielded highly stylized outputs but often at the cost of significant facial feature distortion, thereby compromising subject identity.

To mitigate this critical issue, we introduced a "Mosaic Restored Content Image" as input to the model. As qualitatively illustrated in Table~\ref{tab:comparison} , this technique substantially improves identity preservation, especially in challenging scenarios involving multiple subjects or diminutive facial regions within the image. Compared to the baseline pipeline \cite{wang2024instantstyleplus}, our proposed method exhibits a marked enhancement in maintaining identity fidelity while concurrently achieving the desired level of stylization.

For a rigorous quantitative assessment of our method's identity preservation capabilities, we conducted experiments on the IMDB dataset. Our identity preservation evaluation strategy was informed by methodologies presented in previous work on facial attribute manipulation \cite{mohammadbagheri2024identity}, which emphasizes the importance of maintaining subject identity alongside attribute modification. We partitioned the IMDB dataset into three categories based on the face-to-image area ratio: Category 1 (for faces with an area less than 10\% of the total image), Category 2 (for faces with an area between 10\% and 20\%) and Category 3 (for faces with an area greater than 20\%). Facial regions were automatically detected, cropped, and aligned using established facial landmark detection algorithms \cite{varghese2024yolov8}. Identity preservation was quantified by computing the cosine similarity between the deep feature embeddings of the original content images and their stylized counterparts. These embeddings were extracted using a pre-trained WebFace recognition model \cite{zhu2021webface260m}.

\begin{table}[ht]
\centering

\setlength{\tabcolsep}{5pt} 
\renewcommand{\arraystretch}{1.2} 
\small 
\begin{tabular}{l|cc|cc}
\hline
& \multicolumn{2}{c|}{\textbf{Our pipeline}} & \multicolumn{2}{c}{\textbf{ISP}} \\
\textbf{Category} & \textbf{Cosine $\uparrow$} & \textbf{LPIPS $\downarrow$} & \textbf{Cosine $\uparrow$} & \textbf{LPIPS $\downarrow$} \\ \hline
\textbf{Category 1} & & & & \\
Anime  & \textbf{0.320} & 0.38 & 0.169 & \textbf{0.37} \\
Comic  & \textbf{0.380} & 0.42 & 0.143 & 0.42 \\
Sketch & \textbf{0.581} & \textbf{0.36} & 0.201 & 0.44 \\
\hline
\textbf{Category 2} & & & & \\
Anime  & \textbf{0.422} & 0.37 & 0.365 & \textbf{0.31} \\
Comic  & \textbf{0.524} & 0.42 & 0.292 & \textbf{0.37} \\
Sketch & \textbf{0.673} & \textbf{0.37} & 0.417 & 0.39 \\
\hline
\textbf{Category 3} & & & & \\
Anime  & 0.564 & 0.34 & \textbf{0.698} & \textbf{0.25} \\
Comic  & 0.637 & 0.43 & \textbf{0.646} & \textbf{0.32} \\
Sketch & \textbf{0.727} & 0.37 & 0.704 & \textbf{0.35} \\
\hline
\end{tabular}
\caption{Comparison of Cosine Similarity and LPIPS between Our pipeline and ISP. \textbf{Category 1}: faces with an area less than 10\% of the total image, \textbf{Category 2}:faces with an area between 10\% and 20\%, \textbf{Category 3}: faces with an area greater than 20\%.}
\label{tab:metrics_comparison}
\end{table}

As detailed in Table~\ref{tab:metrics_comparison}, our proposed pipeline was evaluated against the ISP baseline using Cosine Similarity for identity preservation and LPIPS for perceptual fidelity. Our approach demonstrates superior identity preservation, as evidenced by significantly higher cosine similarity scores, particularly for smaller facial regions (Categories 1 and 2), owing to our Mosaic Restored Content Image technique. Conversely, the ISP baseline generally yields lower (better) LPIPS scores, indicating higher perceptual quality, an advantage most prominent for larger faces (Category 3). These results reveal a critical trade-off: Our pipeline excels at robust identity retention, whereas the ISP pipeline often prioritizes overall perceptual similarity.

\subsection{Mosaic-Style Reference for Mitigating Identity and Background Leakage}

In the base pipeline, we observed that selecting a single style reference image containing a distinct identity and background often leads to unintended “leakage” of both facial characteristics and environmental details in the final stylized output. Specifically, the stylized image may inherit the identity of the person depicted in the style reference, along with its background elements and color distribution. To mitigate this issue, we propose to construct a mosaic-style reference composed of multiple images from the same stylistic domain (e.g., anime, 3D, or sketch). By diversifying the source of color distributions and structural cues, this mosaic-based approach dilutes the influence of any single image, thus reducing identity and background leakage. As illustrated in Figure~\ref{fig:2}, our method preserves identity in the content image more effectively while maintaining a richer and more consistent color distribution in the generated output.

\subsection{Impact of the Content Consistency Loss Hyperparameter on Stylization Results}

\begin{figure*}[ht]
    \centering
    \includegraphics[width=0.80\linewidth, height=0.35\textheight]{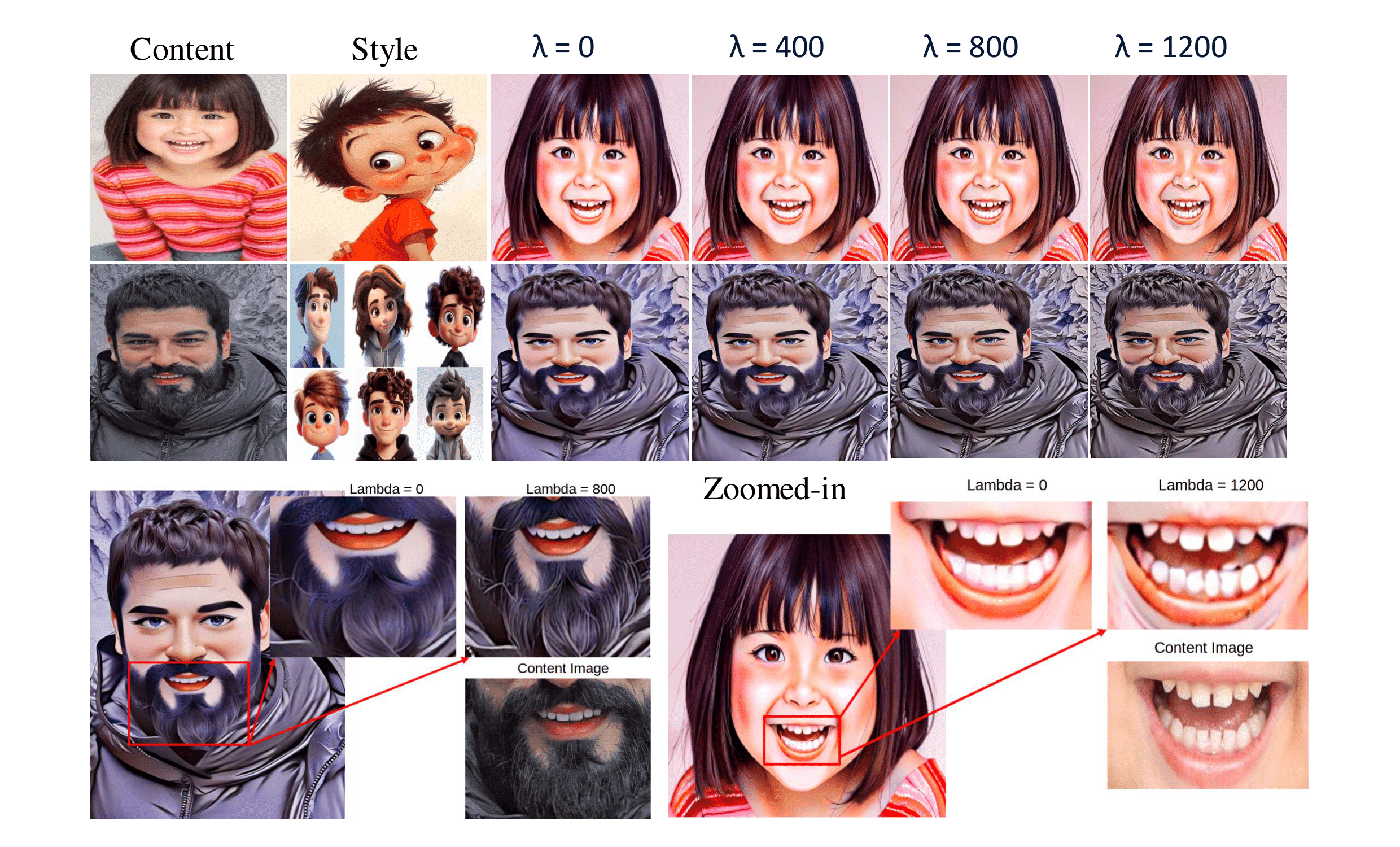}
    \caption{Effect of Varying the Content Consistency Loss on the Stylized Output, Emphasizing Attention to Fine Details.}
    \label{fig:3}
\end{figure*}

Figure~\ref{fig:3} illustrates the impact of varying the hyperparameter \( \lambda_c \) in the content consistency loss on the preservation of fine details in the stylized output (equation ~\ref{eq:refined_noise}). Our experiments reveal that increasing \( \lambda_c \) enhances the retention of subtle content features. For example, the teeth of the girl in the image become progressively clearer as \( \lambda_c \) is raised. This improvement in detail fidelity demonstrates the effectiveness of our noise-refinement strategy in maintaining structural information. However, when \( \lambda_c \) exceeds approximately 1200, the images exhibit a noticeable color saturation, indicating an overemphasis on content preservation that compromises the intended stylistic effect. These findings underscore the critical importance of optimizing \( \lambda_c \) to achieve a balanced trade-off between style transfer and content fidelity.

\section{Conclusion}

In this work, we presented a novel, training-free framework for identity-preserving stylized image generation using diffusion models. Our work specifically addresses the challenge of maintaining identity integrity in challenging scenarios, such as images with diminutive facial regions or multiple subjects. To this end, we introduced two primary technical contributions: first, the Mosaic Restored Content Image technique, which demonstrably enhances identity retention in stylized outputs; and second, a training-free content consistency loss designed to improve the preservation of fine-grained content details by focusing the model's attention on the original image during stylization.  Our findings confirm that integrating these content-aware refinements within the diffusion process leads to significant improvements in both identity preservation and overall stylistic control. We also highlighted the crucial role of text prompts in achieving desired stylization levels; specifically, strategic selection of style-related keywords was shown to effectively guide the generation towards high-quality artistic outputs.  In summary, our proposed approach provides an effective and efficient solution for generating stylized images that faithfully preserve subject identity without the need for additional model training or fine-tuning. Future research could extend this work by further developing adaptive noise refinement techniques and exploring advanced prompt engineering strategies to achieve even greater control over style and content in diffusion-based image synthesis.

{
    \small
    \bibliographystyle{ieeenat_fullname}
    \bibliography{main}

\begin{thebibliography}{49}
\providecommand{\natexlab}[1]{#1}
\providecommand{\url}[1]{\texttt{#1}}
\expandafter\ifx\csname urlstyle\endcsname\relax
  \providecommand{\doi}[1]{doi: #1}\else
  \providecommand{\doi}{doi: \begingroup \urlstyle{rm}\Url}\fi

\bibitem[Al-Mekhlafi and Liu(2024)]{al2024artistic}
Hanadi Al-Mekhlafi and Shiguang Liu.
\newblock Artistic style transfer based on attention with knowledge distillation.
\newblock In \emph{Computer Graphics Forum}, page e15127. Wiley Online Library, 2024.

\bibitem[Banar et~al.(2021)Banar, Sabatelli, Geurts, Daelemans, and Kestemont]{banar2021transfer}
Nicolay Banar, Matthia Sabatelli, Pierre Geurts, Walter Daelemans, and Mike Kestemont.
\newblock Transfer learning with style transfer between the photorealistic and artistic domain.
\newblock In \emph{IS\&T International Symposium on Electronic Imaging. Computer Vision and Image Analysis of Art 2021}, pages 041--1, 2021.

\bibitem[Chen et~al.(2025)Chen, Xiang, Hu, Ye, Yu, Cheng, and Zhang]{chen2025comprehensive}
Hang Chen, Qian Xiang, Jiaxin Hu, Meilin Ye, Chao Yu, Hao Cheng, and Lei Zhang.
\newblock Comprehensive exploration of diffusion models in image generation: a survey.
\newblock \emph{Artificial Intelligence Review}, 58\penalty0 (4):\penalty0 99, 2025.

\bibitem[Chung et~al.(2024)Chung, Hyun, and Heo]{Chung_2024_CVPR}
Jiwoo Chung, Sangeek Hyun, and Jae-Pil Heo.
\newblock Style injection in diffusion: A training-free approach for adapting large-scale diffusion models for style transfer.
\newblock In \emph{Proceedings of the IEEE/CVF Conference on Computer Vision and Pattern Recognition (CVPR)}, pages 8795--8805, 2024.

\bibitem[Efros and Leung(1999)]{efros1999texture}
Alexei Efros and Thomas Leung.
\newblock Texture synthesis by non-parametric sampling.
\newblock In \emph{Proceedings of International Conference on Computer Vision (ICCV’99)}, Kerkyra, Greece, 1999.

\bibitem[Everaert et~al.(2023)Everaert, Bocchio, Arpa, Süsstrunk, and Achanta]{everaert2023diffusion}
Martin~Nicolas Everaert, Marco Bocchio, Sami Arpa, Sabine Süsstrunk, and Radhakrishna Achanta.
\newblock Diffusion in style.
\newblock In \emph{Proceedings of the IEEE/CVF International Conference on Computer Vision}, pages 2251--2261, 2023.

\bibitem[Frenkel et~al.(2024)Frenkel, Vinker, Shamir, and Cohen-Or]{frenkel2024implicit}
Yarden Frenkel, Yael Vinker, Ariel Shamir, and Daniel Cohen-Or.
\newblock Implicit style-content separation using b-lora.
\newblock In \emph{European Conference on Computer Vision}, pages 181--198. Springer, 2024.

\bibitem[Gao et~al.(2020)Gao, Tian, and Qi]{gao2020rpd}
Xin Gao, Yong Tian, and Zhen Qi.
\newblock Rpd-gan: Learning to draw realistic paintings with generative adversarial network.
\newblock \emph{IEEE Transactions on Image Processing}, 29:\penalty0 8706--8720, 2020.

\bibitem[Gao et~al.(2022)Gao, Zhang, and Tian]{gao2022learning}
Xin Gao, Yifan Zhang, and Yong Tian.
\newblock Learning to incorporate texture saliency adaptive attention to image cartoonization.
\newblock In \emph{Proceedings of the International Conference on Machine Learning}, pages 7183--7207, 2022.

\bibitem[Garibi et~al.(2024)Garibi, Patashnik, Voynov, Averbuch-Elor, and Cohen-Or]{garibi2024renoise}
Daniel Garibi, Or Patashnik, Andrey Voynov, Hadar Averbuch-Elor, and Daniel Cohen-Or.
\newblock Renoise: Real image inversion through iterative noising.
\newblock In \emph{European Conference on Computer Vision}, pages 395--413. Springer, 2024.

\bibitem[Gatys et~al.(2016)Gatys, Ecker, and Bethge]{gatys2016image}
Leon~A. Gatys, Alexander~S. Ecker, and Matthias Bethge.
\newblock Image style transfer using convolutional neural networks.
\newblock In \emph{IEEE Conference on Computer Vision and Pattern Recognition (CVPR)}, pages 1033--1038, Piscataway, 2016. IEEE.

\bibitem[Goodfellow et~al.(2014)Goodfellow, Pouget-Abadie, Mirza, Xu, Warde-Farley, Ozair, Courville, and Bengio]{goodfellow2014generative}
Ian~J. Goodfellow, Jean Pouget-Abadie, Mehdi Mirza, Bing Xu, David Warde-Farley, Sherjil Ozair, Aaron~C. Courville, and Yoshua Bengio.
\newblock Generative adversarial networks.
\newblock \emph{Communications of the ACM}, 63:\penalty0 139--144, 2014.

\bibitem[Hertz et~al.(2023)Hertz, Voynov, Fruchter, and Cohen-Or]{hertz2023style}
A. Hertz, A. Voynov, S. Fruchter, and D. Cohen-Or.
\newblock Style aligned image generation via shared attention.
\newblock \emph{arXiv preprint arXiv:2312.02133}, 2023.

\bibitem[Hertzmann et~al.(2001)Hertzmann, Jacobs, Oliver, Curless, and Salesin]{hertzmann2001image}
Aaron Hertzmann, Brian Jacobs, Noah Oliver, Brian Curless, and David Salesin.
\newblock Image analogies.
\newblock In \emph{Computer Graphics (Proc. Siggraph 2001)}, pages 327--340, New York, 2001. ACM Press.

\bibitem[Ho et~al.(2020)Ho, Jain, and Abbeel]{ho2020denoising}
Jonathan Ho, Ajay Jain, and Pieter Abbeel.
\newblock Denoising diffusion probabilistic models.
\newblock In \emph{Proceedings of the International Conference on Neural Information Processing Systems}, pages 6840--6851, 2020.

\bibitem[Huang and Belongie(2017)]{huang2017arbitrary}
Xun Huang and Serge~J. Belongie.
\newblock Arbitrary style transfer in real-time with adaptive instance normalization.
\newblock In \emph{2017 IEEE International Conference on Computer Vision (ICCV)}, pages 1510--1519, 2017.

\bibitem[Isola et~al.(2016)Isola, Zhu, Zhou, and Efros]{isola2017image}
Phillip Isola, Jun-Yan Zhu, Tinghui Zhou, and Alexei~A. Efros.
\newblock Image-to-image translation with conditional adversarial networks.
\newblock In \emph{2017 IEEE Conference on Computer Vision and Pattern Recognition (CVPR)}, pages 5967--5976, 2016.

\bibitem[Jing et~al.(2019)Jing, Yang, Feng, Ye, Yu, and Song]{jing2019neural}
Yongcheng Jing, Yezhou Yang, Zunlei Feng, Jingwen Ye, Yizhou Yu, and Mingli Song.
\newblock Neural style transfer: A review.
\newblock \emph{IEEE transactions on visualization and computer graphics}, 26\penalty0 (11):\penalty0 3365--3385, 2019.

\bibitem[Johnson et~al.(2016)Johnson, Alahi, and Li]{johnson2016perceptual}
Justin Johnson, Anima Alahi, and Fei-Fei Li.
\newblock Perceptual losses for real-time style transfer and super-resolution.
\newblock In \emph{European Conference on Computer Vision (ECCV)}, pages 694--711, Berlin, 2016. Springer.

\bibitem[Katzir et~al.(2020)Katzir, Lischinski, and Cohen-Or]{katzir2020crossdomain}
Oren Katzir, Dani Lischinski, and Daniel Cohen-Or.
\newblock Crossdomain cascaded deep translation.
\newblock In \emph{European Conference on Computer Vision}, 2020.

\bibitem[Kwon et~al.(2023)Kwon, Jeong, and Uh]{kwon2023diffusion}
Mingi Kwon, Jaeseok Jeong, and Youngjung Uh.
\newblock Diffusion models already have a semantic latent space.
\newblock In \emph{The Eleventh International Conference on Learning Representations}, 2023.

\bibitem[Lai et~al.(2017)Lai, Huang, Ahuja, and Yang]{lai2017deep}
Wei-Sheng Lai, Jia-Bin Huang, Narendra Ahuja, and Ming-Hsuan Yang.
\newblock Deep laplacian pyramid networks for fast and accurate super-resolution.
\newblock In \emph{Proceedings of the IEEE conference on Computer Vision and pattern recognition}, pages 624--632, 2017.

\bibitem[Li et~al.(2023)Li, Li, Savarese, and Hoi]{blip}
Junnan Li, Dongxu Li, Silvio Savarese, and Steven Hoi.
\newblock Blip-2: Bootstrapping language-image pre-training with frozen image encoders and large language models.
\newblock In \emph{International conference on machine learning}, pages 19730--19742. PMLR, 2023.

\bibitem[Li et~al.(2017)Li, Fang, Yang, Wang, Lu, and Yang]{li2017universal}
Yijun Li, Chen Fang, Jimei Yang, Zhaowen Wang, Xin Lu, and Ming-Hsuan Yang.
\newblock Universal style transfer via feature transforms.
\newblock In \emph{Advances in neural information processing systems}, 2017.

\bibitem[Li et~al.(2018)Li, Liu, Li, Yang, and Kautz]{li2018closed}
Yijun Li, Ming-Yu Liu, Xueting Li, Ming-Hsuan Yang, and Jan Kautz.
\newblock A closed-form solution to photorealistic image stylization.
\newblock In \emph{Proceedings of the European conference on Computer Vision (ECCV)}, pages 453--468, 2018.

\bibitem[Men et~al.(2022)Men, Yao, Cui, Lian, and Xie]{men2022dct}
Yifang Men, Yuan Yao, Miaomiao Cui, Zhouhui Lian, and Xuansong Xie.
\newblock Dct-net: domain-calibrated translation for portrait stylization.
\newblock \emph{ACM Transactions on Graphics (TOG)}, 41\penalty0 (4):\penalty0 1--9, 2022.

\bibitem[Mohammadbagheri et~al.(2024)Mohammadbagheri, Ayar, Nickabadi, and Safabakhsh]{mohammadbagheri2024identity}
Najmeh Mohammadbagheri, Fardin Ayar, Ahmad Nickabadi, and Reza Safabakhsh.
\newblock Identity-preserving editing of multiple facial attributes by learning global edit directions and local adjustments.
\newblock \emph{Computer vision and image understanding}, 246:\penalty0 104047, 2024.

\bibitem[Mokady et~al.(2023)Mokady, Hertz, Aberman, Pritch, and Cohen-Or]{mokady2023null}
Ron Mokady, Amir Hertz, Kfir Aberman, Yael Pritch, and Daniel Cohen-Or.
\newblock Null-text inversion for editing real images using guided diffusion models.
\newblock In \emph{Proceedings of the IEEE/CVF Conference on Computer Vision and Pattern Recognition}, pages 6038--6047, 2023.

\bibitem[Park and Lee(2019)]{park2019arbitrary}
Dae~Young Park and Kwang~Hee Lee.
\newblock Arbitrary style transfer with style-attentional networks.
\newblock In \emph{Proceedings of the IEEE/CVF conference on computer vision and pattern recognition}, pages 5880--5888, 2019.

\bibitem[Park et~al.(2020)Park, Efros, Zhang, and Zhu]{park2020contrastive}
Taesung Park, Alexei~A. Efros, Richard Zhang, and Jun-Yan Zhu.
\newblock Contrastive learning for unpaired image-to-image translation.
\newblock In \emph{European Conference on Computer Vision}, 2020.

\bibitem[Rezaei et~al.(2025)Rezaei, Ayar, Javanmardi, Tsukada, and Javanmardi]{rezaei2025you}
Mohammad~Ali Rezaei, Fardin Ayar, Ehsan Javanmardi, Manabu Tsukada, and Mahdi Javanmardi.
\newblock Where do you go? pedestrian trajectory prediction using scene features.
\newblock \emph{arXiv preprint arXiv:2501.13848}, 2025.

\bibitem[Song et~al.(2020)Song, Meng, and Ermon]{song2020denoising}
Jiaming Song, Chenlin Meng, and Stefano Ermon.
\newblock Denoising diffusion implicit models.
\newblock \emph{arXiv preprint arXiv:2010.02502}, 2020.

\bibitem[Varghese and Sambath(2024)]{varghese2024yolov8}
Rejin Varghese and M Sambath.
\newblock Yolov8: A novel object detection algorithm with enhanced performance and robustness.
\newblock In \emph{2024 International Conference on Advances in Data Engineering and Intelligent Computing Systems (ADICS)}, pages 1--6. IEEE, 2024.

\bibitem[von Platen et~al.(2023)von Platen, Patil, Lozhkov, Cuenca, Lambert, Rasul, Davaadorj, and Wolf]{vonplaten2023diffusers}
Patrick von Platen, Suraj Patil, Anton Lozhkov, Pedro Cuenca, Nathan Lambert, Kashif Rasul, Mishig Davaadorj, and Thomas Wolf.
\newblock Diffusers: State-of-the-art diffusion models, 2023.

\bibitem[Wang et~al.(2024{\natexlab{a}})Wang, Spinelli, Wang, Bai, Qin, and Chen]{wang2024instantstyle}
Haofan Wang, Matteo Spinelli, Qixun Wang, Xu Bai, Zekui Qin, and Anthony Chen.
\newblock Instantstyle: Free lunch towards style-preserving in text-to-image generation.
\newblock \emph{ArXiv, abs/2404.02733}, 2024{\natexlab{a}}.

\bibitem[Wang et~al.(2024{\natexlab{b}})Wang, Xing, Huang, Ai, Wang, and Bai]{wang2024instantstyleplus}
H. Wang, P. Xing, R. Huang, H. Ai, Q. Wang, and X. Bai.
\newblock Instantstyleplus: Style transfer with content-preserving in text-to-image generation.
\newblock \emph{arXiv preprint arXiv:2407.00788}, 2024{\natexlab{b}}.

\bibitem[Wang et~al.(2021)Wang, Xie, Dong, and Shan]{wang2021real}
Xintao Wang, Liangbin Xie, Chao Dong, and Ying Shan.
\newblock Real-esrgan: Training real-world blind super-resolution with pure synthetic data.
\newblock In \emph{Proceedings of the IEEE/CVF international conference on computer vision}, pages 1905--1914, 2021.

\bibitem[Wang et~al.(2020)Wang, Zhao, Chen, Qiu, Mo, Lin, Xing, and Lu]{wang2020diversified}
Zhizhong Wang, Lei Zhao, Haibo Chen, Lihong Qiu, Qihang Mo, Sihuan Lin, Wei Xing, and Dongming Lu.
\newblock Diversified arbitrary style transfer via deep feature perturbation.
\newblock In \emph{Proceedings of the IEEE/CVF Conference on Computer Vision and Pattern Recognition}, pages 7789--7798, 2020.

\bibitem[Wang et~al.(2023{\natexlab{a}})Wang, Wang, Xie, Qi, Shan, Wang, and Luo]{wang2023styleadapter}
Z. Wang, X. Wang, L. Xie, Z. Qi, Y. Shan, W. Wang, and P. Luo.
\newblock Styleadapter: A single-pass lora-free model for stylized image generation.
\newblock \emph{arXiv preprint arXiv:2309.01770}, 2023{\natexlab{a}}.

\bibitem[Wang et~al.(2023{\natexlab{b}})Wang, Zhao, and Xing]{wang2023stylediffusion}
Zhizhong Wang, Lei Zhao, and Wei Xing.
\newblock Stylediffusion: Controllable disentangled style transfer via diffusion models.
\newblock In \emph{Proceedings of the IEEE/CVF International Conference on Computer Vision}, pages 7677--7689, 2023{\natexlab{b}}.

\bibitem[Xie et~al.(2021)Xie, Chen, Sun, and et~al.]{xie2021dg}
Yuchen Xie, Xin Chen, Lei Sun, and et al.
\newblock Dg-font: Deformable generative networks for unsupervised font generation.
\newblock In \emph{Proceedings of the IEEE Conference on Computer Vision and Pattern Recognition}, pages 5130--5140, 2021.

\bibitem[Yang et~al.(2022)Yang, Jiang, Liu, and Loy]{yang2022Pastiche}
Shuai Yang, Liming Jiang, Ziwei Liu, and Chen~Change Loy.
\newblock Pastiche master: Exemplar-based high-resolution portrait style transfer.
\newblock In \emph{CVPR}, 2022.

\bibitem[Yang et~al.(2023)Yang, Hwang, and Ye]{yang2023zero}
Serin Yang, Hyunmin Hwang, and Jong~Chul Ye.
\newblock Zero-shot contrastive loss for text-guided diffusion image style transfer.
\newblock \emph{arXiv preprint arXiv:2303.08622}, 2023.

\bibitem[Ye et~al.(2023)Ye, Zhang, Liu, Han, and Yang]{ye2023ip}
H. Ye, J. Zhang, S. Liu, X. Han, and W. Yang.
\newblock Ip-adapter: Text compatible image prompt adapter for text-to-image diffusion models.
\newblock \emph{arXiv preprint arXiv:2308.06721}, 2023.

\bibitem[Zhang et~al.(2023{\natexlab{a}})Zhang, Rao, and Agrawala]{zhang2023adding}
Lvmin Zhang, Anyi Rao, and Maneesh Agrawala.
\newblock Adding conditional control to text-to-image diffusion models.
\newblock In \emph{Proceedings of the IEEE/CVF International Conference on Computer Vision}, pages 3836--3847, 2023{\natexlab{a}}.

\bibitem[Zhang et~al.(2022)Zhang, Tang, Dong, Huang, Ma, Lee, and Xu]{zhang2022domain}
Yuxin Zhang, Fan Tang, Weiming Dong, Haibin Huang, Chongyang Ma, Tong-Yee Lee, and Changsheng Xu.
\newblock Domain enhanced arbitrary image style transfer via contrastive learning.
\newblock In \emph{ACM SIGGRAPH 2022 Conference Proceedings}, pages 1--8, 2022.

\bibitem[Zhang et~al.(2023{\natexlab{b}})Zhang, Huang, Tang, Huang, Ma, Dong, and Xu]{zhang2023inversion}
Yuxin Zhang, Nisha Huang, Fan Tang, Haibin Huang, Chongyang Ma, Weiming Dong, and Changsheng Xu.
\newblock Inversion-based style transfer with diffusion models.
\newblock In \emph{Proceedings of the IEEE/CVF Conference on Computer Vision and Pattern Recognition}, pages 10146--10156, 2023{\natexlab{b}}.

\bibitem[Zhu et~al.(2017)Zhu, Park, Isola, and Efros]{zhu2017unpaired}
Jun-Yan Zhu, Taesung Park, Phillip Isola, and Alexei~A. Efros.
\newblock Unpaired image-to-image translation using cycle-consistent adversarial networks.
\newblock In \emph{2017 IEEE International Conference on Computer Vision (ICCV)}, pages 2242--2251, 2017.

\bibitem[Zhu et~al.(2021)Zhu, Huang, Deng, Ye, Huang, Chen, Zhu, Yang, Lu, Du, et~al.]{zhu2021webface260m}
Zheng Zhu, Guan Huang, Jiankang Deng, Yun Ye, Junjie Huang, Xinze Chen, Jiagang Zhu, Tian Yang, Jiwen Lu, Dalong Du, et~al.
\newblock Webface260m: A benchmark unveiling the power of million-scale deep face recognition.
\newblock In \emph{Proceedings of the IEEE/CVF Conference on Computer Vision and Pattern Recognition}, pages 10492--10502, 2021.

\end{thebibliography}
}

\end{document}